\documentclass[lettersize,onecolumn]{IEEEtran}

\usepackage{latexsym,amsmath,amssymb,bbm}
\usepackage{utfsym}

\usepackage{array,ifthen,mleftright}
\usepackage{cite}
\usepackage{booktabs,multirow,multicol}
\usepackage{makecell}
\usepackage{enumerate}
\usepackage{algorithmic,algorithm}

\usepackage{subfig}
\usepackage{stfloats}
\usepackage{textcomp,blindtext}
\usepackage{afterpage}

\usepackage{url,verbatim}

\usepackage{tikz}
\usetikzlibrary{matrix}
\usetikzlibrary{positioning}
\usetikzlibrary{patterns}

\usepackage{pgfplots}
\usepgfplotslibrary{groupplots}

\newcommand{\ifonecolumn}[2]{\ifthenelse{\boolean{CLASSOPTIONonecolumn}}{#1}{#2}}
\newcommand{\iftwocolumn}[2]{\ifthenelse{\boolean{CLASSOPTIONtwocolumn}}{#1}{#2}}

\newcommand{\visual}{}

\newcommand{\T}{\mathsf{T}}

\newcommand{\R}{\mathbb{R}}

\newcommand{\calO}{\mathcal{O}}

\newcommand{\eqdef}{\stackrel{\scriptscriptstyle\bigtriangleup}{=} }
\newcommand{\argmin}{\operatornamewithlimits{argmin}}
\newcommand{\argmax}{\operatornamewithlimits{argmax}}
\newcommand{\cond}{\hspace{0.04em}|\hspace{0.06em}}

\definecolor{gray}{rgb}{0.5, 0.5, 0.5}
\newcommand{\gray}{\color{gray}}

\newcommand{\varestim}[1]{\protect\hathat{#1}}

\makeatletter
\newcommand{\hathat}[1]{% 
\begingroup%
  \let\macc@kerna\z@%
  \let\macc@kernb\z@%
  \let\macc@nucleus\@empty%
  \hat{\mathchoice%
    {\raisebox{.3ex}{\vphantom{\ensuremath{\displaystyle #1}}}}%
    {\raisebox{.3ex}{\vphantom{\ensuremath{\textstyle #1}}}}%
    {\raisebox{.24ex}{\vphantom{\ensuremath{\scriptstyle #1}}}}%
    {\raisebox{.21ex}{\vphantom{\ensuremath{\scriptscriptstyle #1}}}}%
    \smash{\hat{#1}}}%
\endgroup%
}
\makeatother

\makeatletter
\DeclareFontFamily{U}{MnSymbolA}{}
\DeclareSymbolFont{MnSyA}{U}{MnSymbolA}{m}{n}
\DeclareFontShape{U}{MnSymbolA}{m}{n}{
	<-6> MnSymbolA5
	<6-7> MnSymbolA6
	<7-8> MnSymbolA7
	<8-9> MnSymbolA8
	<9-10> MnSymbolA9
	<10-12> MnSymbolA10
	<12-> MnSymbolA12}{}
\DeclareMathSymbol{\smallrightarrow}{\mathrel}{MnSyA}{0}
\DeclareMathSymbol{\smallleftarrow}{\mathrel}{MnSyA}{2}
\DeclareMathSymbol{\smallleftrightarrow}{\mathrel}{MnSyA}{16}
\newcommand{\smallrightarrowfill@}{\arrowfill@\relbar\relbar\smallrightarrow}
\newcommand{\smallleftarrowfill@}{\arrowfill@\smallleftarrow\relbar\relbar}
\newcommand{\smallleftrightarrowfill@}
{\arrowfill@\smallleftarrow\relbar\smallrightarrow}
\renewcommand{\overrightarrow}{\mathpalette{\overarrow@\smallrightarrowfill@}}
\renewcommand{\overleftarrow}{\mathpalette{\overarrow@\smallleftarrowfill@}}
\renewcommand{\overleftrightarrow}
{\mathpalette{\overarrow@\smallleftrightarrowfill@}}
\makeatother
\providecommand{\msgf}[2]{\protect\overrightarrow{#1}_{\mspace{-3mu}#2}} % Forward Message
\providecommand{\msgb}[2]{\protect\overleftarrow{#1}_{\mspace{-3mu}#2}} % Backward Message
\newcounter{examplecntr}
\newenvironment{example}[1][]%
{\begin{trivlist}\item[]\refstepcounter{examplecntr}%
 {\itshape Example~\theexamplecntr%
  \ifthenelse{\equal{#1}{}}{}{ (#1)}.
}}%
{\hfill$\Box$\end{trivlist}}

\newcounter{propositioncntr}
\newenvironment{proposition}[1][]%
{\begin{trivlist}\item[]\refstepcounter{propositioncntr}%
{\itshape Proposition~\thepropositioncntr%
  \ifthenelse{\equal{#1}{}}{}{ (#1)}.
}}%
{\hfill$\Box$\end{trivlist}}

\newenvironment{proofof}[1]{\begin{trivlist}\item[]{\it Proof\ifthenelse{\equal{#1}{}}{}{ #1}:}
 }{\hfill$\Box$\end{trivlist}}

\newcounter{algorithmcntr}
\newcounter{saveequation}
\newenvironment{trivalgorithm}[1][]%
{\normalsize\vspace{0.5ex}%
\refstepcounter{algorithmcntr}%
\setcounter{saveequation}{\value{equation}}%
\setcounter{equation}{0}%
\renewcommand{\theequation}{\arabic{algorithmcntr}.\arabic{equation}}%
{\bfseries Algorithm~\thealgorithmcntr%
\ifthenelse{\equal{#1}{}}{:}{: #1}% 
}}%
{\setcounter{equation}{\value{saveequation}}}

\newcommand{\pkw}[1]{\textbf{#1}}    % 

\begin{document}
%\DeclareGraphicsExtensions{.jpg,.pdf,.mps,.png}
\graphicspath{{pdf_pictures/}}

\title{L1 Regularization Paths in Linear Models by Parametric Gaussian Message Passing}

\author{Yun-Peng Li and Hans-Andrea Loeliger
\thanks{Yun-Peng Li (yunpli@isi.ee.ethz.ch) is with ETH Zurich, Switzerland}
%\thanks{Hans-Andrea Loeliger (loeliger@ethz.ch) retired from ETH Zurich in 2026.}
\thanks{Hans-Andrea Loeliger (loeliger@ethz.ch) was also with ETH Zurich and is now retired.}
}

\maketitle

\begin{abstract}
The paper proposes new algorithms
for computing the regularization path of linear-model fitting 
(i)~with L1 costs for the independent variables and L2 costs for the dependent variables, 
and
(ii)~with L2 costs for the independent variables and L1 costs for the dependent variables.
Unlike previously proposed algorithms, 
the proposed algorithms for (i) and (ii) are exact duals of each other.
The heart of the proposed algorithms is parametric Gaussian message passing
(i.e., Kalman-type forward-backward recursions)
in the pertinent factor graphs.
We first develop such algorithms for 
linear models with general (but preferably low-dimensional) state space representations.
We then specialize to specific state space representations of generic linear models;
in this case, the structure of the factor graph 
allows to refine the proposed algorithms 
such that the computational complexity drops to linear scale.
\end{abstract}

\begin{IEEEkeywords}
Factor graphs, regularization path, parametric quadratic programming, 
linear non-Gaussian models, 
homotopy techniques.
\end{IEEEkeywords}

\section{Introduction}

Consider fitting a linear model $y=Fu$ with given $F\in\R^{L\times K}$
to given data $\breve y\in\R^L$
according to 
\begin{equation} \label{eqn:qfittingL1pen}
\hat u = \argmin_{u\in\R^K} \mleft( \| Fu - \breve y \|_2^2 + s^2 \| u \|_1 \mright)
\end{equation}
(i.e., the LASSO problem \cite{Tibshirani1996})
or according to
\begin{equation} \label{eqn:L1fittingQpen}
%\hat y = F \argmin_{u\in\R^K} \mleft( s^2 \| Fu - \breve y \|_1 + \| u \|_2^2 \mright)
\hat u = \argmin_{u\in\R^K} \mleft( s^2 \| Fu - \breve y \|_1 + \| u \|_2^2 \mright)
\end{equation}
with parameter $s\in\R$.
Specifically, 
consider the problem of 
numerically computing (\ref{eqn:qfittingL1pen}) and (\ref{eqn:L1fittingQpen}),
not just for a single value of $s^2$,
but the entire function $\hat u(s^2)$%
---the regularization path---%
for all $s^2\geq 0$.
Note that, in (\ref{eqn:qfittingL1pen}), the L1 penalty applies to the independent variables 
while in  (\ref{eqn:L1fittingQpen}), the L1 penalty applies to the dependent variables. 

In this paper,
we propose a new method 
for computing the path $\hat u(s^2)$ of (\ref{eqn:qfittingL1pen}) 
and a dual method for computing $\hat u(s^2)$ for (\ref{eqn:L1fittingQpen}),
where both methods boil down to repeated parametric Gaussian message passing 
in a state space model.
The proposed methods work also if $\| u \|_1$ in (\ref{eqn:qfittingL1pen})
and $\| Fu - \breve y \|_1$ in (\ref{eqn:L1fittingQpen})
are generalized to 
$\sum_{k=1}^K \kappa_n(u_k)$ and $\sum_{\ell=1}^L \kappa_n\mleft( (Fu)_\ell - \breve y_\ell \mright)$
respectively,
where $\kappa_n$ are continuous, piecewise linear, and convex 
(which includes the Vapnik loss function and hinge loss functions),
and where $(Fu)_\ell$ denotes the $\ell$-th row of $Fu$;
with these generalizations, 
(\ref{eqn:L1fittingQpen}) includes  
the linear support vector machine (SVM) problem.

The related literature is vast. 
L1 regularization has long been heavily used and researched 
in a variety of fields including optimization, control theory, and machine learning. 
Clearly, the path $\hat u(s^2)$ 
can be computed approximately 
by computing $\hat u(s_1^2), \ldots, \hat u(s_n^2)$ 
for a sufficiently dense list $s_1^2, \ldots, s_n^2$.
However, in this paper, we are interested in efficiently computing the exact paths 
in (\ref{eqn:qfittingL1pen}) and (\ref{eqn:L1fittingQpen}),
which are parametric quadratic programming problems \cite{Ritter1981,bank1982non, murty1988linear}.
Variations of such problems have been studied, e.g., 
for sensitivity analysis in optimization,
for precomputing control laws in explicit model preditive control \cite{bemporad2021explicit}, 
and for determining an appropriate value of $s^2$ 
in machine learning
\cite{Osborne2000, EHJT2004}.

It is well known that $\hat u(s^2)$ is piecewise linear in $s^2$ 
and constant beyond some finite threshold $s^2_\text{max}$ \cite{Ritter1981, Rosset2007}.
In consequence, computing these paths amounts to computing the (finite) list of points (called knots)
$s_1^2, \ldots, s^2_\text{max}$ where the slope of the path changes.
Most existing path algorithms are path following active-set methods 
that transform the original problem into a sequence of quadratic programming (sub-)problems 
\cite{Ritter1981,bank1982non, bemporad2002explicit}.

Recent path methods 
for machine learning applications 
include 
the homotopy technique of \cite{Osborne2000}, 
least angle regression (LARS) \cite{EHJT2004}, 
fused LASSO \cite{Hoefling2010},
and generalized LASSO \cite{Tibshirani2011, Arnold2016},
all of which address (\ref{eqn:qfittingL1pen}).
Path methods for (\ref{eqn:L1fittingQpen}) have received less attention;
the related problem of computing the regularization path for support vector machines (SVM) 
was addressed in \cite{Hastie2004}.
Path methods based on sweep and inverse sweep operators were proposed in \cite{Zhou2013}.

The mentioned path algorithms usually rely on specialized linear-algebra subroutines, 
e.g., updating/downdating in Cholesky factorization for LASSO \cite{EHJT2004} 
and partitioned matrix inversion for SVM \cite{Hastie2004}, sparse QR decomposition for generalized LASSO \cite{Tibshirani2011}, 
and an external quadratic programming solver for model predictive control \cite{bemporad2021explicit}. 

In this paper, we will exploit a state space model of $F$,
which does not seem to have been considered in the prior literature.
We also build on prior work on iterative Gaussian message passing algorithms 
for state space models that can compute (\ref{eqn:qfittingL1pen}) and (\ref{eqn:L1fittingQpen})
for fixed values of $s^2$
\cite{Loeliger2016, Keusch2023,Li2024,Li2025}.
In particular, we will use the tabulated message computation rules from 
\cite{Loeliger2016}, which improve on the tables in \cite{LDHKLK2007},
and the duality of the proposed methods for (\ref{eqn:qfittingL1pen}) and (\ref{eqn:L1fittingQpen})
follows from the obvious dualities in these tables, cf.\ \cite{Li2025}.

In summary, the proposed algorithms
\begin{enumerate}
\item
apply to regularization of independent and dependent variables,
\item
reflect the duality between (\ref{eqn:qfittingL1pen}) and (\ref{eqn:L1fittingQpen}),
\item
apply to state space models and can exploit their structure,
but can also handle generic matrices $F$,
\item
apply to several different problems, 
which the prior literature has handled with specialized algorithms,
\item
address non-Gaussian problems by 
parameteric Gaussian message passing, which is a new idea in factor graphs,
\item
do not require specialized matrix operations,
\item
and their computational complexity is on a par with specialized prior methods.
\end{enumerate}

The paper is structured as follows. 
Section~\ref{sec:StatModel} introduces the state space setting.
Sections \ref{sec:InputPath} and \ref{sec:OutputPath}
describe the proposed regularized-input path algorithm
and the proposed regularized-output path algorithm, respectively,
for general state space models;
in this form, the proposed algorithms are attractive 
if $F$ admits a low-dimensional state space representation.
Sections \ref{sec:RefineFig2} and~\ref{sec:RefineFig3}
propose reductions in complexity 
for general matrices $F$.
As a sanity check, some numerical experiments are reported in Section~\ref{sec:NumericalExperiments}.

%Conclusions: Section~\ref{sec:Conclusions}
Appendix~\ref{appsec:GaussianMsgTables} reproduces the mentioned tables from \cite{Loeliger2016}.
Appendices \ref{appsec:ParametricBFFD} and~\ref{appsec:ParametricFFBDD} 
prove the pivotal Propositions \ref{prop:BFFD:affine} and~\ref{prop:FFBDD:affine}
and provide details of the proposed algorithms.

\section{Statistical State Space Model}
\label{sec:StatModel}

\subsection{Linear State Space Formulation}

We begin by reformulating (\ref{eqn:qfittingL1pen}) and (\ref{eqn:L1fittingQpen})
in terms of linear state space models. 
For ease of exposition, we consider only state space models 
with scalar inputs $u_1,\ldots,u_N\in\R$, scalar outputs $y_1,\ldots,y_N\in\R$, 
and states $x_0, x_1, \ldots, x_N\in\R^M$ that evolve according to 
\begin{equation} \label{eqn:JustLSSM:state}
x_n = A x_{n-1} + b_n u_n
\end{equation}
and
\begin{equation} \label{eqn:JustLSSM:output}
y_n = c_n^\T x_n
\end{equation}
for $n=1, 2, \ldots, N$, 
with $A\in\R^{M\times M}$ and column vectors 
$b_1,\ldots,b_N\in\R^M$ and $c_1,\ldots,c_N\in\R^M$.

As for (\ref{eqn:qfittingL1pen}), 
we will work with its generalization to
\begin{IEEEeqnarray}{r,C,l}
\mleft( \hat u, \hat x_0 \mright) 
& = &  \argmin_{u\in\R^N\!,\, x_0\in\R^M} \Big(
         (x_0 - \breve x_0)^\T Q_0 (x_0 - \breve x_0)
\iftwocolumn{\IEEEeqnarraynumspace\nonumber\\ &&}{}
         {} + s^2 \sum_{n=1}^N \kappa(u_n) 
            + \sum_{n=1}^N \mleft( y_n - \breve y_n \mright)^2
\iftwocolumn{\IEEEeqnarraynumspace\nonumber\\ &&}{}
         {} + (x_N - \breve x_N)^\T Q_N (x_N - \breve x_N) 
         \Big),
         \IEEEeqnarraynumspace
         \label{eqn:qfittingL1pen:LSSM}
\end{IEEEeqnarray}
where 
$x_1,\ldots,x_N$ and $y\eqdef (y_1,\ldots,y_N)^\T$
are determined by $x_0$ and $u\eqdef (u_1,\ldots,u_N)^\T$
via (\ref{eqn:JustLSSM:state}) and (\ref{eqn:JustLSSM:output}),
and where $Q_0$ and $Q_N$ are positive semi-definite matrices.  
The cost function $\kappa$ 
will be discussed in Section~\ref{sec:BFFD:CostFunctions}.
The term $(x_0 - \breve x_0)^\T Q_0 (x_0 - \breve x_0)$
and the minimization over $x_0$
can be omitted if a fixed initial state $x_0=\breve x_0$ is given.

\begin{example}[First-order Trend Filter]\label{ex:pwlm}
A first-order spline model is obtained with $M=2$,
\begin{equation}\label{eqn:pwlm}
A = \mleft( \begin{array}{cc}
      1 & 1 \\
      0 & 1
    \end{array} \mright),
\hspace{2em}
b_n = \mleft( \begin{array}{c}
      0 \\
      1
    \end{array} \mright),
\hspace{2em}
c_n = \mleft( 1, 0 \mright),
\end{equation}
$\kappa(u_n) = |u_n|$, and $Q_0=Q_N=0$.
\end{example}

The special case (\ref{eqn:qfittingL1pen}) (with a general matrix $F$)
is obtained with 
$\kappa(u_n)=|u_n|$,
$M=L$, $N=K$, $A=I$, 
$b_n = n$-th column of $F$, $c_n=0$, 
$x_0$ fixed to $x_0=\breve x_0=0$ (and $Q_0$ need not be specified),
$Q_N=I$, $\breve x_N = \breve y$, and noting that $x_N = Fu$ by (\ref{eqn:JustLSSM:state}).
In this special case, (\ref{eqn:qfittingL1pen:LSSM}) reduces to
\begin{equation} \label{eqn:qfittingL1pen:LSSM:matrixF}
\hat u = \argmin_{u\in\R^N} \mleft( 
         s^2 \sum_{n=1}^N \kappa(u_n) 
         + \| x_N - \breve x_N \|_2^2
         \mright).
\end{equation}
However, 
in many applications, the structure of $F$ admits a state space representation
with $M\ll \min\{ K, L \}$, cf.\ Example~\ref{ex:pwlm}.

As for (\ref{eqn:L1fittingQpen}), 
we will work with its generalization to 
\begin{IEEEeqnarray}{r,C,l}
\mleft( \hat u, \hat x_0 \mright) 
& = &  \argmin_{u\in\R^N\!,\, x_0\in\R^M} \Big(
         (x_0 - \breve x_0)^\T Q_0 (x_0 - \breve x_0) 
\iftwocolumn{\IEEEeqnarraynumspace\nonumber\\ &&}{}
         {} + \sum_{n=1}^N u_n^2
         {} + s^2 \sum_{n=1}^N \kappa\mleft( y_n - \breve y_n \mright)
\iftwocolumn{\IEEEeqnarraynumspace\nonumber\\ &&}{}
         {} + (x_N - \breve x_N)^\T Q_N (x_N - \breve x_N)
         \Big),
         \IEEEeqnarraynumspace
         \label{eqn:L1fittingQpen:LSSM}
\end{IEEEeqnarray}
where $x_1,\ldots,x_N$ and $y = (y_1,\ldots,y_N)^\T$
are determined by $x_0$ and $u= (u_1,\ldots,u_N)^\T$
via (\ref{eqn:JustLSSM:state}) and (\ref{eqn:JustLSSM:output}),
and where $Q_0$ and $Q_N$ are positive semi-definite matrices.
The term $(x_0 - \breve x_0)^\T Q_0 (x_0 - \breve x_0)$
and the minimization over $x_0$ 
can be omitted if a fixed initial state $x_0=\breve x_0$ is given.

\begin{example}[First-order Median-Kalman Smoothing]\label{ex:fomf}
A first-order median smoother is obtained with $M=2$,
$A, b_n, c_n$ as in (\ref{eqn:pwlm}), 
$\kappa\mleft( y_n - \breve y_n \mright) = | y_n - \breve y_n |$, and $Q_0=Q_N=0$.
\end{example}

The special case (\ref{eqn:L1fittingQpen}) (with a general matrix $F$)
is obtained with $M=K$, $N=L$, $A=I$, 
$b_n=0$, $c_n^\T = n$-th row of $F$, 
$\breve x_0=0$, $Q_0=I$, $Q_N=0$, and noting that $x_0$ takes the role of $u$ in (\ref{eqn:L1fittingQpen}).
In this case, (\ref{eqn:L1fittingQpen:LSSM}) reduces to
\begin{equation} \label{eqn:L1fittingQpen:LSSM:matrixF}
\hat x_0 = \argmin_{x_0\in\R^K} \mleft( 
         \| x_0 \|_2^2
         + s^2 \sum_{n=1}^N \kappa\mleft( y_n - \breve y_n \mright)
         \mright).
\end{equation}

\subsection{Statistical Model and Factor Graphs}
\label{sec:StatModelFG}

The proposed methods are naturally expressed in terms of 
Gaussian message passing in a factor graph of a statistical model.
For the general case (\ref{eqn:qfittingL1pen:LSSM}), we will use 
the joint probability density function
\begin{equation} \label{eqn:StatisticalModel}
p(u_1,\ldots,u_N, x_0,\ldots,x_N, y_1, \ldots, y_N)
= 
p(x_0) p(\breve x_N \cond x_N)
\prod_{n=1}^N p(u_n) p(\breve y_n \cond y_n),
\end{equation}
where $x_1,\ldots,x_N$ and $y$
are determined by $x_0$ and $u_1,\ldots,u_N$
according to (\ref{eqn:JustLSSM:state}) and (\ref{eqn:JustLSSM:output}),
and with
\begin{equation} \label{eqn:pX0}
p(x_0) \propto \exp\mleft( - \frac{1}{2\sigma^2} (x_0 - \breve x_0)^\T Q_0 (x_0 - \breve x_0) \mright)
\end{equation}
or with fixed initial state $x_0=\breve x_0$,
with
\begin{equation} \label{eqn:pXN}
p(\breve x_N \cond x_N) \propto \exp\mleft( - \frac{1}{2\sigma^2} (x_N - \breve x_N)^\T Q_N (x_N - \breve x_N) \mright),
\end{equation}
and either with
\begin{equation} \label{eqn:pUn:primal}
p(u_n) \propto {\exp}\big( {- \kappa(u_n)} \big)
\hspace{1em}\text{and}\hspace{1em}
p(\breve y_n \cond y_n) \propto  \exp\mleft( -\frac{(y_n - \breve y_n)^2}{2\sigma^2} \mright)
\end{equation}
for (\ref{eqn:qfittingL1pen:LSSM})
or with
\begin{equation} \label{eqn:pUn:dual}
p(u_n) \propto \exp\mleft( - \frac{u_n^2}{2\sigma^2} \mright)
\hspace{1em}\text{and}\hspace{1em}
p(\breve y_n \cond y_n) \propto  {\exp}\big( {-\kappa(y_n - \breve y_n)} \big),
\end{equation}
for (\ref{eqn:L1fittingQpen:LSSM}),
where the variances (or covariance matrices) are parameterized by
\begin{equation}
\sigma^2 \eqdef s^2/2.
\end{equation}
Note that we allow (\ref{eqn:pX0})--(\ref{eqn:pUn:dual}) to be improper (i.e., not normalizable).

With this statistical model, 
both (\ref{eqn:qfittingL1pen:LSSM}) and (\ref{eqn:L1fittingQpen:LSSM}) 
amount to (the pertinent  components of) the joint MAP estimate of 
the independent variables $X_0$ and $U=(U_1,\ldots,U_N)^\T$
and the dependent variables 
$X_1,\ldots,X_N$ and $Y=(Y_1,\ldots,Y_N)^\T$.

A~factor graph of (\ref{eqn:StatisticalModel}) 
is shown in Fig.~\ref{fig:StatisticalModel:FactorGraph}.
(We use factor graphs as in \cite{LDHKLK2007}, 
with boxes representing factors and edges representing variables.)
Factor graphs of the special cases (\ref{eqn:qfittingL1pen:LSSM:matrixF}) 
and (\ref{eqn:L1fittingQpen:LSSM:matrixF}) 
are shown in Figures \ref{fig:FactorGraph:DecompInput} and~\ref{fig:FactorGraph:DecompOutput}, 
respectively.

\begin{figure*}
\newcommand{\optional}[1]{#1}
\centering
\begin{tikzpicture}[x=1mm, y=1mm, >=latex]
\tikzset{%
smallbox/.style = {draw, rectangle, inner sep=0mm, minimum width=4mm, minimum height=4mm},
midbox/.style = {draw, rectangle, minimum width=5.5mm, minimum height=5.5mm, inner sep=1mm},
blobbox/.style = {draw, fill=black, rectangle, inner sep=0mm, minimum width=1.75mm, minimum height=1.75mm},
}
\node[smallbox, label={below:$p(x_0)$}] (pX0) {};
\draw[->] (pX0) -- node[above]{$X_0$} +(11,0);
\draw (pX0)+(14.5,0) node {$\ldots$};
\draw (pX0)+(30,0) node[midbox] (A) {$A$};
\draw[->] (A)+(-12,0) -- node[above, pos=0.4]{$X_{n-1}$} (A);

\draw (A)+(15,0) node[smallbox] (add) {$+$};
\draw[->] (A) -- node[above]{\optional{$X_n'''$}} (add);
\draw (add)+(0,11) node[midbox] (bn) {$b_n$};
\draw[->] (bn) -- (add);
\draw (bn)+(0,12) node[smallbox,label={right:$p(u_n)$}] (pUn) {};
\draw[->] (pUn) -- node[left] {$U_n$} (bn);

\draw (add)+(15,0) node[smallbox] (equ) {$=$};
\draw[->] (add) -- node[above] {\optional{$X_n''$}} (equ);
\draw (equ)+(0,-12) node[midbox] (cn) {$c_n^\T$};
\draw[->] (equ) -- node[left] {\optional{$X_n'$}} (cn);
\draw (cn)+(0,-12) node[smallbox,label={right:$p(\breve y_n \cond y_n)$}] (pYn) {};
\draw[->] (cn) -- node[left] {$Y_n$} (pYn);

\draw[->] (equ) -- node[above] {$X_n$} +(11,0);
\draw (equ)+(14.5,0) node {$\ldots$};
\draw (equ)+(30,0) node[smallbox, label={below:$p(\breve x_N \cond x_N)$}] (pXN) {};
\draw[->] (pXN)+(-12,0) -- node[above, pos=0.4]{$X_N$} (pXN);
\end{tikzpicture}
\vspace{1ex}
\caption{\label{fig:StatisticalModel:FactorGraph}%
    Factor graph of (\ref{eqn:StatisticalModel}) and (\ref{eqn:qfittingL1pen:LSSM}).
    The variables $X_n''$ and $X_n'''$
    are used in Algorithms \ref{alg:BFFD:SSM} and \ref{alg:FFBDD:SSM}.
    }
\end{figure*}

\begin{figure*}
\newcommand{\optional}[1]{{\gray #1}}
\centering
\begin{tikzpicture}[x=1mm, y=1mm, >=latex]
\tikzset{%
smallbox/.style = {draw, rectangle, inner sep=0mm, minimum width=4mm, minimum height=4mm},
midbox/.style = {draw, rectangle, minimum width=5.5mm, minimum height=5.5mm, inner sep=1mm},
blobbox/.style = {draw, fill=black, rectangle, inner sep=0mm, minimum width=1.75mm, minimum height=1.75mm},
}
\node[blobbox, label={left:0}] (x0) {};
\draw[->] (x0) -- node[above]{$X_0$} +(11,0);
\draw (x0)+(14.5,0) node {$\ldots$};

\draw (x0)+(30,0) node[smallbox] (add) {$+$};
\draw[->] (add)+(-12,0) -- node[above, pos=0.4]{$X_{n-1}$} (add);
\draw (add)+(0,11) node[midbox] (bn) {$b_n$};
\draw[->] (bn) -- (add);
\draw (bn)+(0,12) node[smallbox,label={right:$e^{-\kappa(u_n)}$}] (pUn) {};
\draw[->] (pUn) -- node[left] {$U_n$} (bn);

\draw[->] (add) -- node[above] {$X_n$} +(11,0);
\draw (add)+(14.5,0) node {$\ldots$};
\draw (add)+(30,0) node[smallbox, label={right:$\exp\mleft( - \frac{\| x_N - \breve x_N \|^2}{2\sigma^2} \mright)$}] (pXN) {};
\draw[->] (pXN)+(-12,0) -- node[above, pos=0.4]{$X_N$} (pXN);
\end{tikzpicture}
\vspace{1ex}
\caption{\label{fig:FactorGraph:DecompInput}%
    Factor graph of (\ref{eqn:qfittingL1pen:LSSM:matrixF}).}
\end{figure*}

\begin{figure*}
\newcommand{\optional}[1]{{\gray #1}}
\centering
\begin{tikzpicture}[x=1mm, y=1mm, >=latex]
\tikzset{%
smallbox/.style = {draw, rectangle, inner sep=0mm, minimum width=4mm, minimum height=4mm},
midbox/.style = {draw, rectangle, minimum width=5.5mm, minimum height=5.5mm, inner sep=1mm},
blobbox/.style = {draw, fill=black, rectangle, inner sep=0mm, minimum width=1.75mm, minimum height=1.75mm},
}
\node[smallbox, label={left:$\exp\mleft( -\frac{\| x_0 \|^2}{2\sigma^2} \mright)$}] (pX0) {};
\draw[->] (pX0) -- node[above]{$X_0$} +(11,0);
\draw (pX0)+(14.5,0) node {$\ldots$};
\draw (pX0)+(30,0) node[smallbox] (equ) {$=$};
\draw[->] (equ)+(-12,0) -- node[above, pos=0.4]{$X_{n-1}$} (equ);

\draw (equ)+(0,-12) node[midbox] (cn) {$c_n^\T$};
%\draw[->] (equ) -- node[left] {\optional{$X_n'$}} (cn);
\draw[->] (equ) -- (cn);
\draw (cn)+(0,-12) node[smallbox,label={right:$e^{-\kappa(y_n - \breve y_n)}$}] (pYn) {};
\draw[->] (cn) -- node[left] {$Y_n$} (pYn);

\draw[->] (equ) -- node[above] {$X_n$} +(11,0);
\draw (equ)+(14.5,0) node {$\ldots$};
\draw[->] (equ)++(18,0) -- node[above] {$X_N$} +(9,0);
\end{tikzpicture}
\vspace{1ex}
\caption{\label{fig:FactorGraph:DecompOutput}%
    Factor graph of (\ref{eqn:L1fittingQpen:LSSM:matrixF}).}
\end{figure*}

\section{Regularized-Input Path by Parametric Backward Filtering Forward Deciding}
\label{sec:InputPath}

\subsection{MAP Estimation in Linear Gaussian Models with Backward Filtering Forward Deciding (BFFD)}
\label{sec:BFFD}

We first consider the statistical model (\ref{eqn:StatisticalModel})--(\ref{eqn:pUn:primal})
for the case where both $p(u_n)$ and $p(\breve y_n \cond y_n)$ are Gaussian in $u_n$ and $y_n$,
respectively.
In this case, the statistical model is linear Gaussian. 
For fixed $\sigma^2$, the joint MAP estimates of $U$, $X_0,\ldots,X_N$, and $Y$
can then be computed by several forward-backward (or backward-forward) Gaussian message passing algorithms,
which can be assembled 
from the tables in \cite{LDHKLK2007} and \cite{Loeliger2016}.
For the convenience of the reader, 
these tables are reproduced in Appendix~\ref{appsec:GaussianMsgTables}.

In these tables, the Gaussian messages and posteriors 
have the form
\begin{IEEEeqnarray}{r,C,l}
\mu(z)
 & \propto & 
  \exp\mleft( -\frac{1}{2} (z - m)^\T W (z-m) \mright)
  \label{eqn:GenGaussianMean}\\
& \propto &  \exp\mleft( -\frac{1}{2} z^\T W z + z^\T \xi  \mright)
  \label{eqn:GenGaussian}
\end{IEEEeqnarray}
%(up to a scale factor) 
and are parameterized 
either by the precision matrix $W$ and $\xi \eqdef Wm$
or by the mean vector $m$ and the covariance matrix $V=W^{-1}$. 
Both $V$ and $W$ are allowed to be positive semi-definite. 
In particular, $W=0$ is permissible, in which case $V$ does not exist;
$V=0$ is permissible as well, in which case (\ref{eqn:GenGaussian}) 
shrinks to a Dirac delta centered at $m$.

Message parameters are denoted by arrows:
e.g., $\msgf{m}{X_n}$ denotes the mean vector of 
the forward message $\msgf{\mu}{X_n}$ along
the edge $X_n$ in the factor graph
and $\msgb{W}{X_n}$ denotes the precision matrix of 
the backward message $\msgb{\mu}{X_n}$
along the same edge,
where forward and backward refer to the direction of the edges 
in Figures \ref{fig:StatisticalModel:FactorGraph}--\ref{fig:FactorGraph:DecompOutput}.
%Posteriors are denoted without arrows:
%e.g., $m_{X_n}$ denotes the posterior mean (= the MAP estimate $\hat x_n$) of $X_n$.

The particular algorithm we will use here is backward filtering forward deciding (BFFD) \cite{Li2024}.
The detailed algorithm is given in Algorithm~\ref{alg:BFFD:SSM}
(and its specialization for Fig.~\ref{fig:FactorGraph:DecompInput} 
is given in Section~\ref{sec:RefineFig2}).
The backward recursion of BFFD is the standard backward information filter,
which computes the parameters $\msgb{W}{X_n''} \in\R^{M\times M}$ and $\msgb{\xi}{X_n''}\in\R^{M}$ 
of the Gaussian messages $\msgb{\mu}{X_n''}$
for $n=N,\ldots,1$;
its key steps (\ref{eqn:alg:BFFD:backw:xiXppp}) and (\ref{eqn:alg:BFFD:backw:WXppp}) 
follow from (\ref{eqn:CompositeAdditionBackwardPrecisionMean}) and (\ref{eqn:CompositeAdditionBackwardPrecision}),
respectively.
The subsequent forward recursion determines the maximizing $x_0, u_1, x_1, u_1, \ldots, u_N, x_N$ one by one
and plugs them in 
(i.e., $\msgf{V}{X_{n-1}}$ is set to zero, cf.\ Fig.~\ref{fig:BFFD:ForwardDeciding}).
Note that this algorithm is a version of dynamic programming 
and its backward recursion may be viewed as iterating the Riccati equation \cite{BellmanBook1957};
the correctness of the algorithm will be obvious
to a reader familiar with Gaussian message passing in cycle-free factor graphs, see also \cite{Li2024}.
Note also that $\msgb{V}{U_n}$ and $\msgb{m}{U_n}$ in the forward recursion do not, strictly speaking,
refer to Fig~\ref{fig:StatisticalModel:FactorGraph}, but to Fig.~\ref{fig:BFFD:ForwardDeciding}.
The possibility of a division by zero in (\ref{eqn:alg:BFFD:backw:H}) 
is addressed in Section~\ref{sec:PathBFFD:LinDep}.

Note that from the quantities computed in the backward recursion, 
only $b_{n}^\T \msgb{W}{X_{n}^{''}} \in\R^{1\times M}$, 
$b_{n}^\T \msgb{\xi}{X_{n}^{''}} \in\R$, 
and $b_{n}^\T \msgb{W}{X_{n}^{''}}b_{n} \in\R$ 
are stored for the forward recursion.
%and the complexity of (\ref{eqn:alg:BFFD:fwd:msgbVU})--(\ref{eqn:alg:BFFD:fwd:hatY}) is only $\calO(M)$.

\begin{table}[tp]
\framebox[\linewidth]{%
\normalsize%
\begin{minipage}{0.95\linewidth}
\begin{trivalgorithm}[Backward Filtering Forward Deciding (BFFD) in Fig.~\ref{fig:StatisticalModel:FactorGraph}]%
\label{alg:BFFD:SSM}\vspace{0.5ex}\\
Input: see Sections \ref{sec:StatModelFG} and \ref{sec:BFFD}\\
Output: $\hat x_0$, $\hat u_n$, $\hat y_n$, and (in this paper) also $\msgb{m}{U_n}$ and $\msgb{V}{U_n}$ for $n=1,\ldots,N$
\\[-1ex]
\hrule
\vspace{-1ex}
\begin{multicols}{2}
%\emph{Backward filtering:}
\begin{enumerate}[1.]
\item
Let $\msgb{W}{X_{N}} = \sigma^{-2} Q_{N}$ and $\msgb{\xi}{X_{N}}= \sigma^{-2}Q_{N}\breve{x}_{N}$.
\item
\pkw{for} $n=N$ \pkw{to} $1$, compute the following:
    \begin{IEEEeqnarray}{rCl}
	    \msgb{\xi}{X_{n}^{''}} & = & \msgb{\xi}{X_{n}} + c_{n}\msgb{\xi}{Y_{n}} 
			\label{eqn:alg:BFFD:backw:xiXpp}\\
	    \msgb{W}{X_{n}^{''}} & = & \msgb{W}{X_{n}} + c_{n}\msgb{W}{Y_{n}}c_{n}^{\mathsf{T}}
	    	\label{eqn:alg:BFFD:backw:WXpp}
    	%\IEEEeqnarraynumspace
	\end{IEEEeqnarray}
	\begin{IEEEeqnarray}{rCl}
		H_{n} & = & \begin{cases}
					  0,  & \text{if $\msgf{V}{U_n}=0$}\\ 
					  \big(b_{n}^{\mathsf{T}}\msgb{W}{X_{n}^{''}}b_{n}\big)^{-1},  & \text{if $\msgf{W}{U_n}=0$}\\
					  \text{by (\ref{eqn:BFFD:Hn:General})},   & \text{else} 
					  %\big(\msgf{W}{U_{n}} + b_{n}^{\mathsf{T}}\msgb{W}{X_{n}^{''}}b_{n}\big)^{-1}
			   \end{cases}
			   \label{eqn:alg:BFFD:backw:H}\\
       h_{n} & = & \begin{cases}
           b_{n}\msgf{m}{U_{n}}, &  \text{if $\msgf{V}{U_{n}}=0$}\\
           % b_{n}    \big(b_{n}^{\mathsf{T}}\msgb{W}{X_{n}^{''}}b_{n}\big)^{-1}\big(\msgf{\xi}{U_{n}}+b_{n}^{\mathsf{T}}\msgb{\xi}{X_{n}^{''}}\big), & \msgf{W}{U_{n}}=0,\\
           b_{n}H_{n}\big(\msgf{\xi}{U_{n}}+b_{n}^{\mathsf{T}}\msgb{\xi}{X_{n}^{''}}\big), & \text{else}
					\end{cases}
			    \label{eqn:alg:BFFD:backw:h}
	\end{IEEEeqnarray}
	\begin{IEEEeqnarray}{rCl}
		\msgb{\xi}{X_{n}^{'''}} & = & \msgb{\xi}{X_{n}^{''}} - \msgb{W}{X_{n}^{''}}h_{n}
			\label{eqn:alg:BFFD:backw:xiXppp}\\
		\msgb{W}{X_{n}^{'''}} & = & \msgb{W}{X_{n}^{''}} - \msgb{W}{X_{n}^{''}}b_{n}H_{n}b_{n}^{\mathsf{T}}\msgb{W}{X_{n}^{''}}
			\label{eqn:alg:BFFD:backw:WXppp}
	\end{IEEEeqnarray}
	\begin{IEEEeqnarray}{rCl}
		\msgb{\xi}{X_{n-1}} & = & A^{\mathsf{T}} \msgb{\xi}{X_{n}^{'''}}
			\label{eqn:alg:BFFD:backw:xiX}\\
		\msgb{W}{X_{n-1}} & = &  A^{\mathsf{T}}	\msgb{W}{X_{n}^{'''}}A
		\label{eqn:alg:BFFD:backw:WX}
	\end{IEEEeqnarray}
	\hspace{1.5em}Store only $b_{n}^\T \msgb{W}{X_{n}^{''}}, b_{n}^\T \msgb{\xi}{X_{n}^{''}}$ and $b_{n}^\T \msgb{W}{X_{n}^{''}}b_{n}$.%
\item[] 
\pkw{end for}
%\rule{0em}{1ex}  % manually adjust columns
\newpage

%\item[]
%\emph{Forward deciding:}
\item
\pkw{if} initial state is fixed to $\breve x_0$
	\begin{enumerate}
	\item[]
    Let $\hat x_0 = \breve x_0$.
    \end{enumerate}
\item
\pkw{else}
	\begin{enumerate}
	\item[]
	Let $\msgf{W}{X_{0}} = \sigma^{-2} Q_{0}$ and $\msgf{\xi}{X_{0}} = \sigma^{-2} Q_{0}\breve{x}_{0}$,\\
	and compute $\hat{x}_{0}$ by solving
		\begin{equation} \label{eqn:PrimalMarginalEstimateInitialState}
			\big(\msgf{W}{X_{0}}+\msgb{W}{X_{0}}\big) \hat x_0 =\msgf{\xi}{X_{0}} + \msgb{\xi}{X_{0}}
		\end{equation}
	\end{enumerate}
\item[]
\pkw{end if}
\item
\pkw{for} $n=1$ \pkw{to} $N$, compute the following:
	\begin{equation} \label{eqn:ForwardRecursionBFFD}
			\hat{x}^{'''}_{n} = A\hat{x}_{n-1}
	\end{equation}
	\begin{IEEEeqnarray}{rCl}	
		\msgb{V}{U_{n}} & = & \big(b_{n}^{\mathsf{T}}\msgb{W}{X_{n}^{''}}b_{n}\big)^{-1}
			\label{eqn:alg:BFFD:fwd:msgbVU}\\
		\msgb{m}{U_{n}} & = & \msgb{V}{U_{n}} \big(b_{n}^{\mathsf{T}}\msgb{\xi}{X_{n}^{''}} - b_{n}^{\mathsf{T}}\msgb{W}{X_{n}^{''}}\hat{x}_{n}^{'''} \big)
			\label{eqn:alg:BFFD:fwd:msgbU}
	\end{IEEEeqnarray}
	\begin{equation} \label{eqn:alg:BFFD:fwd:hatU}
	\hat u_n = \begin{cases}
				   \msgf{m}{U_n}, & \text{if $\msgf{V}{U_n}=0$} \\
				   \msgb{V}{U_n} \msgf{\xi}{U_n} + \msgb{m}{U_n},  & \text{if $\msgf{W}{U_n}=0$} \\
				   \text{by (\ref{eqn:BFFD:decideUn:General}),} 
						 & \text{else}
				\end{cases}
	\end{equation}
	\begin{IEEEeqnarray}{rCl}	
		\hat{x}_{n} & = &  \hat{x}_{n}^{'''}+ b_{n}\hat{u}_{n}
			\label{eqn:alg:BFFD:fwd:hatX} \\
		\hat{y}_{n} & = &  c_{n}^{\mathsf{T}}\hat{x}_{n}
			\label{eqn:alg:BFFD:fwd:hatY}
	\end{IEEEeqnarray}
\item[]
\pkw{end for}
\end{enumerate}
%\rule{0em}{1ex}  % manually adjust columns
\end{multicols}
\vspace{-1ex}% 
\hrule
\vspace{1ex}

Concerning (\ref{eqn:alg:BFFD:backw:H}), the general case
\begin{equation} \label{eqn:BFFD:Hn:General}
H_n =  \big(\msgf{W}{U_{n}} + b_{n}^{\mathsf{T}}\msgb{W}{X_{n}^{''}}b_{n}\big)^{-1}
\end{equation}
is not used in this paper.
\medskip

Concerning (\ref{eqn:alg:BFFD:fwd:hatU}), the general case
\begin{equation} \label{eqn:BFFD:decideUn:General}
\hat u_n = \big( \msgf{W}{U_n} + \msgb{W}{U_n} \big)^{-1} \big( \msgf{\xi}{U_n} + \msgb{\xi}{U_n} \big)
\end{equation}
is not used in this paper.
\end{trivalgorithm}

\end{minipage}%
}
\end{table}

\begin{figure*}
\newcommand{\optional}[1]{{\gray #1}}
\centering
\begin{tikzpicture}[x=1mm, y=1mm, >=latex]
\tikzset{%
smallbox/.style = {draw, rectangle, inner sep=0mm, minimum width=4mm, minimum height=4mm},
midbox/.style = {draw, rectangle, minimum width=5.5mm, minimum height=5.5mm, inner sep=1mm},
blobbox/.style = {draw, fill=black, rectangle, inner sep=0mm, minimum width=1.75mm, minimum height=1.75mm},
}
\node[blobbox, label={above:$\hat x_{n-1}$}] (hatXnm1) {};
\draw (hatXnm1)+(12.5,0) node[midbox] (A) {$A$};
\draw[->] (hatXnm1) -- (A);

\draw (A)+(15,0) node[smallbox] (add) {$+$};
\draw[->] (A) -- node[above]{$X_n'''$} (add);
\draw (add)+(0,11) node[midbox] (bn) {$b_n$};
\draw[->] (bn) -- (add);
\draw (bn)+(0,12) node[smallbox,label={right:$p(u_n)$}] (pUn) {};
\draw[->] (pUn) -- node[left] {$U_n$} (bn);

\draw (add)+(20,0) node[smallbox] (equ) {$=$};
\draw[->] (add) -- node[above,pos=0.3] {$X_n''$} (equ);
\draw (equ)+(0,-12) node[midbox] (cn) {$c_n^\T$};
\draw[->] (equ) -- (cn);
\draw (cn)+(0,-12) node[smallbox,label={right:$p(\breve y_n \cond y_n)$}] (pYn) {};
\draw[->] (cn) -- node[left] {$Y_n$} (pYn);

\draw[->] (equ) -- node[above] {$X_n$} +(11,0);
\draw (equ)+(14.5,0) node {$\ldots$};
\draw (equ)+(30,0) node[smallbox, label={below:$p(\breve x_N \cond x_N)$}] (pXN) {};
\draw[->] (pXN)+(-12,0) -- node[above, pos=0.4]{$X_N$} (pXN);

\draw (pYn)+(-7.5,-5) 
      node[draw, dashed, anchor=south west, 
           minimum width=47.5mm, minimum height=37mm, 
           label=above:$\msgb{\mu}{X_n''}$] {};
\end{tikzpicture}
\vspace{1ex}
\caption{\label{fig:BFFD:ForwardDeciding}%
Illustrating BFFD.}
\end{figure*}

\subsection{Cost Functions and Their Piecewise ``Gaussian'' Representations}
\label{sec:BFFD:CostFunctions}

The proposed algorithm can deal with cost functions $\kappa(u_n)$ that are 
continuous, piecewise linear, and convex (p.l.c),
e.g., 
\begin{equation} \label{eqn:CostPieces}
\kappa(u_n) = \mleft\{ \begin{array}{ll}
        a-u_n,  &  \text{if $u_n < a$} \\
        0,      &  \text{if $u_n = a$} \\
        u_n-a,  &  \text{if $a < u_n < b$}\\
        b-a,    &  \text{if $u_n = b$} \\
        2u_n - a -b, &  \text{if $u_n > b$.}
   \end{array}\mright.
\end{equation}
The following pertinent terminology will be needed.
A \emph{proper segmentation} of a p.l.c.\ function $\kappa$
decomposes $\kappa$ into an alternating sequence of line segments and points
such that (i)~the line segments do not include their endpoints
and (ii)~adjacent line segments have different slopes,
as illustrated by (\ref{eqn:CostPieces}).
The \emph{segments} of the cost function are the line segments and points
in a proper segmentation; e.g., the cost function (\ref{eqn:CostPieces}) 
comprises five segments.

The \emph{subdomain} of each segment refers to the set of $u_n$ to which it applies.
E.g., in (\ref{eqn:CostPieces}), the subdomain of the first segment is the open interval $(-\infty, a)$
and the subdomain of the second segment is $\{ a \}$.
In a proper segmentation, the subdomains of the segments form a partition of $\R$.

If $\kappa$ is p.l.c., then
$\exp\mleft(-\kappa(u_n)\mright)$ is piecewise degenerate Gaussian
(up to irrelevant scale factors).
For $\kappa$ as in (\ref{eqn:CostPieces}), the parameters of these Gaussians
(as introduced in Section~\ref{sec:BFFD})
are as follows:
\begin{equation}\label{eqn:CostPieces:NUP}
\begin{array}{c@{\hspace{1.7em}}cccc}
\text{subdomain} & \msgf{m}{U_n} & \msgf{V}{U_n} & \msgf{W}{U_n} & \msgf{\xi}{U_n} 
\\
\hline
u_n < a  &     &    &  0  &  1 \\
u_n = a     &  a  &  0  &   &  \\
a < u_n < b   &     &   &  0  &  -1 \\
u_n = b     &  b  &  0  &   &  \\
u_n >b  &     &    &  0  &  -2
\end{array}
\end{equation}
For line segments, $-\msgf{\xi}{U_n}$ is the slope of $\kappa(u_n)$ and $\msgf{W}{U_n}=0$;
for point segments, $\msgf{m}{U_n}$ equals the subdomain and $\msgf{V}{U_n}=0$.

\subsection{Parametric BFFD}
\label{sec:BFFD:Parametric}

The proposed algorithm will repeatedly call Algorithm~\ref{alg:BFFD:SSM} (BFFD)
with the following conditions satisfied:
\begin{enumerate}
\item
Either $\msgf{W}{X_0} = \sigma^{-2} Q_0$ and $\msgf{\xi}{X_0} = \sigma^{-2} Q_0 \breve x_0$,
or $\msgf{V}{X_0}=0$ and $\msgf{m}{X_0} = \breve x_0$. 
\item
$\msgb{W}{X_N} = \sigma^{-2} Q_N$ and $\msgb{\xi}{X_N} = \sigma^{-2} Q_N \breve x_N$.
\item
For $n=1,\ldots,N,$ 
$\msgb{W}{Y_n} = \sigma^{-2}$ and $\msgb{\xi}{Y_n} = \sigma^{-2} \breve y_n$.
\item
For $n=1,\ldots,N,$ 
either $\msgf{W}{U_n}=0$ and $\msgf{\xi}{U_n}$ does not depend on $\sigma$,
or $\msgf{V}{U_n}=0$ and $\msgf{m}{U_n}$ does not depend on $\sigma$, cf.~(\ref{eqn:CostPieces:NUP}).
\end{enumerate}

\begin{proposition} \label{prop:BFFD:affine}
When BFFD is called with the above conditions satisfied, 
then, for all $n$, 
\renewcommand{\theenumi}{\roman{enumi}}%
\begin{enumerate}
\item \label{eqn:prop:BFFD:affine:msgbXn}
$\msgb{W}{X_n}$ is a linear function of $\sigma^{-2}$ and $\msgb{\xi}{X_n}$ is an affine function of $\sigma^{-2}$;
\item
the posterior mean $m_{X_n}$ (= the estimate $\hat x_n$) is an affine function of~$\sigma^2$;
\item
$\msgb{V}{U_n}$ is a linear function of $\sigma^2$ and $\msgb{m}{U_n}$ is an affine function of $\sigma^2$;
\item
the posterior mean $m_{U_n}$ (= the estimate $\hat u_n$) is an affine function of $\sigma^2$.
\end{enumerate}
\vspace{-2ex}
\end{proposition}
Note that ``linear'' here does not include affine; 
e.g., (\ref{eqn:prop:BFFD:affine:msgbXn}) claims that, 
$\msgb{W}{X_n} = \sigma^{-2} \msgb{W}{X_n}^{(1)}$ for some matrix $\msgb{W}{X_n}^{(1)}$
while $\msgb{\xi}{X_n} = \msgb{\xi}{X_n}^{(0)} + \sigma^{-2}\msgb{\xi}{X_n}^{(1)}$
for vectors $\msgb{\xi}{X_n}^{(0)}$ and $\msgb{\xi}{X_n}^{(1)}$.
The proof of Proposition~\ref{prop:BFFD:affine} amounts to verifying and tracking 
the linear or affine representation at each step of BFFD, 
which is detailed in Appendix~\ref{appsec:ParametricBFFD};
these detailed steps are also required for carrying out the proposed algorithm.

\subsection{Proposed Algorithm (PathBFFD)}
\label{sec:PathBFFD}

The proposed algorithm is Algorithm~\ref{alg:PathBFFD}.
In this algorithm, the cost function for $U_n$ is called $\kappa_n(u_n)$.
For each index $n$,
the algorithm works with an active segment of $\kappa_n(u_n)$,
the subdomain of which is the \emph{active subdomain}.

The active segments and active subdomains
change as the algorithm progresses. 
These changes can occur only at discrete values (called \emph{knots}) of $\sigma^2$.
The algorithm determines these knots and tracks the active subdomains
for increasing $\sigma^2$, starting from $\sigma^2=0$.
The starting point ($\sigma^2=0$) 
is the minimum-norm solution of the ordinary least-squares problem.

\begin{table}[tp]
\framebox[\linewidth]{%
\normalsize%
\begin{minipage}{0.95\linewidth}
\begin{trivalgorithm}[PathBFFD]%
\label{alg:PathBFFD}\vspace{0.5ex}\\
Input: see main text.\\
Output: the knots $\sigma_0^2 = 0, \sigma_1^2, \ldots, \sigma_{\ell-1}^2, \sigma_\ell^2=\infty$
and the affine representation of $\hat u_n(\sigma^2)$ between these points.
\\[-1ex]
\hrule
\begin{algorithmic}[1]
\STATE Let $\sigma_0^2=0$.
\STATE \label{line:alg:PBFFD:firstrunBFFD}
       Run Algorithm~\ref{alg:BFFD:SSM} 
       %(or Algorithm~\ref{alg:BFFD:DecompInput:SpecUpd)
       with $\msgf{\xi}{U_n}=0$ and $\msgf{W}{U_n}=0$ for all $n$, and with arbitrary fixed $\sigma^2>0$.
\STATE For all $n$, record the resulting $\hat u_n$ as the final result $\hat u_n(\sigma_0^2)$.
\STATE \label{line:alg:PBFFD:initactivesubdomain} 
       For every $n$, set the active subdomain of $\kappa_n(u_n)$ to the subdomain that contains $\hat u_n$.
\STATE\label{line:alg:PBFFD:ell0} 
       $\ell \gets 0$.
\REPEAT \label{line:alg:PBFFD:repeat} 
   \STATE $\ell \gets \ell+1$.
   \STATE \label{line:alg:PBFFD:rerunBFFD} 
          Run Algorithm~\ref{alg:BFFD:SSM} 
          with $\msgf{\xi}{U_n}$ and $\msgf{W}{U_n}$ (or $\msgf{m}{U_n}$ and $\msgf{V}{U_n}$)
          according to the (fixed) active segments of $\kappa_n(u_n)$ (cf.\ Section~\ref{sec:BFFD:CostFunctions}),
          %and parameterized by $\sigma^2$ as detailed in Section~\ref{sec:BFFD:Parametric}.
          and with $\msgb{\xi}{X_n}$, $\msgb{W}{X_n}$,
          $\msgb{m}{U_n}$, $\msgb{V}{U_n}$, $\hat u_n$, etc.,
          as functions of $\sigma^2$
          (as detailed in Section~\ref{sec:BFFD:Parametric}).
          %We thus obtain new $\hat u_n(\sigma^2)$, $\msgb{m}{U_n}(\sigma^2)$, $\msgb{V}{U_n}(\sigma^2)$ for all $n$.
          \\[0.5ex]
   \STATE \label{line:alg:PBFFD:eoas}
          For every $n$, let 
		  \begin{equation} \label{eqn:hatun:kappa}
			\varestim{u}_n(\sigma^2)
			= \argmax_{u_n} {\exp}\big(-\kappa_n(u_n)\big) \msgb{\mu}{U_n}(u_n),
		  \end{equation}
		  where $\msgb{\mu}{U_n}(u_n)$ is Gaussian with parameters $\msgb{m}{U_n}(\sigma^2)$ and $\msgb{V}{U_n}(\sigma^2)$,
          and determine 
          \begin{equation} \label{eqn:alg:PBFFD:sigman}
            \sigma_{\ell,n}^2 = \sup \big\{ \sigma^2 : \text{$\varestim{u}_n(\sigma^2)$ is in the active subdomain of $\kappa_n(u_n)$} \big\},
          \end{equation}
          as detailed in Section~\ref{sec:PathBFFD}.
          \\[0.5ex]
   \STATE Let $\sigma_\ell^2 = \min_n\big\{ \sigma_{\ell,n}^2 :  \sigma_{\ell,n}^2 > \sigma_{\ell-1}^2 \big\}$. 
   \IF{ $\sigma_\ell^2 < \infty$ }
       \STATE \label{line:alg:PBFFD:iftext}
          For the index\footnote{%
          According to the unique path assumption of Section~\ref{sec:PathBFFD:LinDep}, there is only one such index.} 
          $n$
          such that $\sigma_{\ell,n}^2 = \sigma_\ell^2$,
          change the active segment and subdomain of $\kappa_n(u_n)$ as follows:
          if the present active segment is a line segment,
          then the next active segment is the point segment whose subdomain 
          contains $\varestim{u}_n(\sigma_\ell^2)$;
          if the present active segment is a point segment,
          then the next active segment is the line segment whose subdomain 
          contains $\varestim{u}_n(\sigma_\ell^2+\varepsilon)$
          with positive $\varepsilon \rightarrow 0$.
   \ENDIF \label{line:alg:PBFFD:endif}
\UNTIL $\sigma_\ell^2 = \infty$.
\end{algorithmic}
\end{trivalgorithm}
\vspace{1ex}

\hrule

\end{minipage}%
}
\end{table}

The quantity $\varestim{u}_n$ in (\ref{eqn:hatun:kappa}) 
is an estimate of $U_n$ that uses the whole cost function $\kappa_n(u_n)$
while $\hat u_n$ from (\ref{eqn:alg:BFFD:fwd:hatU}) 
uses only the presently active segment of $\kappa_n(u_n)$.
%Between knots, 
For $\sigma^2_{\ell-1} \leq \sigma^2 < \sigma^2_{\ell,n}$,
we have $\varestim{u}_n(\sigma^2) = \hat u_n(\sigma^2)$.

Line \ref{line:alg:PBFFD:eoas} of Algorithm~\ref{alg:PathBFFD}
is aided by Table~\ref{tbl:DecidingRules}, which shows
both $\varestim{u}_n$
and the conditions on $\msgb{m}{U_n}$ and $\msgb{V}{U_n}$ 
for (\ref{eqn:hatun:kappa}) to lie 
in the different subdomains of $\kappa_n(u_n)$.
These conditions allow to easily determine (\ref{eqn:alg:PBFFD:sigman}),
as illustrated by the following example.

\begin{table}[tp]
\normalsize
\[
\begin{array}{@{}cc@{\hspace{1.5em}}c@{\hspace{2.5em}}c@{\hspace{2em}}l@{\hspace{3em}}c@{\hspace{2em}}l@{}}
	\toprule
	\multicolumn{2}{@{}c}{\kappa_{n}(z_n)}  &  \text{subdomain}  &  \varestim{u}_n  & \quad\text{condition}  & \varestim{y}_n  & \quad\text{condition}
	\\[0.5ex]
	\midrule
	\\[-3ex]
	\multirow{3}{1em}{L1}
	& a-z_n  &  z_n < a
	&  \msgb{m}{U_n} + \msgb{V}{U_n}  & \msgb{m}{U_n} < -\msgb{V}{U_n}+a
	&  \msgf{m}{Y_n} + \msgf{V}{Y_n}  & \msgf{m}{Y_n} < -\msgf{V}{Y_n}+a
	\\
	& 0  &  z_n = a
	&  a   &  \msgb{m}{U_n} \text{ in-between} 
	&  a   &  \msgf{m}{Y_n} \text{ in-between} 
	\\
	& z_n-a  &  z_n > a
	&  \msgb{m}{U_n} - \msgb{V}{U_n}  & \msgb{m}{U_n} > \msgb{V}{U_n}+a
	&  \msgf{m}{Y_n} - \msgf{V}{Y_n}  & \msgf{m}{Y_n} > \msgf{V}{Y_n}+a
	\\[1ex]
	\midrule
	\\[-2ex]
	\multirow{3}{1em}{\rotatebox{90}{\makebox[0em][c]{\rule{0.5em}{0ex}hinge I}}}
	& a-z_n  &  z_n < a
	&   \msgb{m}{U_{n}} + \msgb{V}{U_{n}} & \msgb{m}{U_{n}} < -\msgb{V}{U_{n}}+a
	&   \msgf{m}{Y_{n}} + \msgf{V}{Y_{n}} & \msgf{m}{Y_{n}} < -\msgf{V}{Y_{n}}+a
	\\
	& 0  &  z_n = a
	&   a  &  \msgb{m}{U_{n}} \text{ in-between}
	&   a  &  \msgf{m}{Y_{n}} \text{ in-between}
	  \\
	& 0  &  z_n > a
	&      \msgb{m}{U_{n}}& \msgb{m}{U_{n}} > a
	&      \msgf{m}{Y_{n}}& \msgf{m}{Y_{n}} > a
	\\[1ex]
	\midrule
	\\[-2ex]
	\multirow{3}{1em}{\rotatebox{90}{\makebox[0em][c]{\rule{0.3em}{0ex}hinge II}}}
	& 0  &  z_n < b
	&       \msgb{m}{U_{n}} & \msgb{m}{U_{n}} < b
	&       \msgf{m}{Y_{n}} & \msgf{m}{Y_{n}} < b
	\\
	& 0  &  z_n = b
	&   b  &  \msgb{m}{U_{n}} \text{ in-between}
	&   b  &  \msgf{m}{Y_{n}} \text{ in-between}
	 \\
	& z_n - b  &  z_n > b
	&       \msgb{m}{U_{n}} - \msgb{V}{U_{n}} & \msgb{m}{U_{n}} > \msgb{V}{U_{n}}+b
	&       \msgf{m}{Y_{n}} - \msgf{V}{Y_{n}} & \msgf{m}{Y_{n}} > \msgf{V}{Y_{n}}+b
	\\[1ex]
	\midrule
	\\[-2ex]
	\multirow{3}{1em}{\rotatebox{90}{Vapnik loss\rule{1.3em}{0ex}}}
	& 2(a-z_n)  &  z_n < a
	&   \msgb{m}{U_{n}}+2\msgb{V}{U_{n}} &  \msgb{m}{U_{n}} < -2\msgb{V}{U_{n}}+a
	&   \msgf{m}{Y_{n}}+2\msgf{V}{Y_{n}} &  \msgf{m}{Y_{n}} < -2\msgf{V}{Y_{n}}+a
	  \\
	& 0  &  z_n = a
	&   a  &  \msgb{m}{U_{n}} \text{ in-between}
	&   a  &  \msgf{m}{Y_{n}} \text{ in-between}
	  \\
	& 0  &  a < z_n < b
	&       \msgb{m}{U_{n}} & a < \msgb{m}{U_{n}} < b
	&       \msgf{m}{Y_{n}} & a < \msgf{m}{Y_{n}} < b
	  \\
	& 0  &  z_n = b
	&    b  &  \msgb{m}{U_{n}} \text{ in-between}
	&    b  &  \msgf{m}{Y_{n}} \text{ in-between}
	  \\
	& 2(z_n-b)  &  z_n > b
	&       \msgb{m}{U_{n}}-2\msgb{V}{U_{n}} &  \msgb{m}{U_{n}} > 2\msgb{V}{U_{n}}+b
	&       \msgf{m}{Y_{n}}-2\msgf{V}{Y_{n}} &  \msgf{m}{Y_{n}} > 2\msgf{V}{Y_{n}}+b
	\\[1ex]
	\bottomrule
  \end{array}
%\end{math}
\]
\vspace{1ex}
\caption{\label{tbl:DecidingRules}%
Estimates $\varestim{u}_n$ (\ref{eqn:hatun:kappa}) for PathBFFD 
and estimates $\varestim{y}_n$ (\ref{eqn:PathFFBDD:hatDual}) for PathFFBDD,
as well as the pertinent subdomain of the cost function $\kappa_n(z_n)$,
where $z_n=u_n$ for input estimation and $z_n=y_n$
for output estimation.
}
\end{table}

\begin{example}%[L1 Regularization]
\label{example:PBFFD:nextSubdomain}
Let $\kappa_n(u_n) = |u_n|$. Assume that $\msgb{m}{U_n} = \sigma^2 + 2$ and $\msgb{V}{U_n} = 2\sigma^2$.
From Table~\ref{tbl:DecidingRules}, we have
\begin{equation} \label{eqn:ExSubdomainsSigma}
\begin{array}{cccc}
\toprule
\kappa_n(u_n)  &  \text{subdomain}  &  \varestim{u}_n(\sigma^2)  &  \text{condition} \\
\midrule
 -u_n  &  u_n < 0  &  3\sigma^2 + 2  &  \sigma^2 < -2/3  \\
   0   &  u_n = 0  &  0              &  -2/3 \leq \sigma^2 \geq 2  \\
  u_n  &  u_n > 0  &  2 - \sigma^2   & \sigma^2 < 2
\\
\bottomrule
\end{array}
\vspace{1ex}
\end{equation}

\begin{itemize}
\item 
Suppose the active subdomain of $\kappa_n(u_n)$ is $\{ u_n\colon u_n > 0\}$.
From (\ref{eqn:ExSubdomainsSigma}), 
$\varestim{u}_n$ is in the active subdomain
if $\sigma^2 < 2$.
Thus $\sigma_{\ell,n}^2 = 2$.
\item
Suppose the active subdomain of $\kappa_n(u_n)$ is $\{ 0 \}$.
From (\ref{eqn:ExSubdomainsSigma}),
$\varestim{u}_n$ is in the active subdomain
if $ \sigma^2 \geq 2$.
Thus $\sigma_{\ell,n}^2 = \infty$.
\item
Suppose the active subdomain of $\kappa_n(u_n)$ is $\{ u_n\colon u_n < 0\}$.
From (\ref{eqn:ExSubdomainsSigma}), $\varestim{u}_n$ cannot be in the active subdomain.
This case cannot occur. 
\end{itemize}
\vspace{-2ex}
\end{example}

\noindent
The final result of Algorithm~\ref{alg:PathBFFD}
consists of the list $\sigma_0^2 = 0, \sigma_1^2, \ldots, \sigma_{\ell-1}^2, \sigma_\ell^2=\infty$
and the affine representations 
$\hat u_n(\sigma^2)$ between these points.
(Direct linear interpolation between $\hat u_n(\sigma_{\ell-1}^2)$ and $\hat u_n(\sigma_{\ell}^2)$
works in principle, but is numerically unstable because $\mleft| \sigma_{\ell}^2 - \sigma_{\ell-1}^2 \mright|$
can be arbitrarily small.)

The essential novelty of Algorithm~\ref{alg:PathBFFD} 
is its use of parametric BFFD according to Section~\ref{sec:BFFD:Parametric}; 
the tracking of the changing active segments (with one change per iteration in Algorithm~\ref{alg:PathBFFD})
is a variation of standard ideas in active-set methods.

\subsection{Unique Path Assumption}
\label{sec:PathBFFD:LinDep}

Situations can be constructed where (\ref{eqn:alg:BFFD:backw:H})
amounts to a division by zero. 
Closer examination (not detailed here) reveals that this happens if and only if 
the regularization path is not unique. 
For $F$ as in (\ref{eqn:qfittingL1pen}), 
this condition can equivalently be formulated as follows. 
\begin{proposition}[Unique Path Condition for Regularized Input]
Let $J_\ell\subset\{1,\ldots,N\}$ (with $N=K$) be the set of indices $n$ such that, in round $\ell$ of PathBFFD,
the active subdomain of $\kappa_n$ 
is a line segment. 
Then the regularization path is unique if and only if, for every $\ell$,
the columns of $F$ with indices in $J_\ell$ 
are linearly independent.
\end{proposition}
This condition is well known in the literature and 
assumed to hold \cite{EHJT2004}.
Modifying the proposed algorithm to handle non-unique paths properly is possible, 
but complicated and irrelevant for many applications (cf.\ Section~\ref{sec:NumericalExperiments}).
Therefore, we here choose to follow the tradition and simply assume the path to be unique.

\subsection{Complexity}
\label{sec:PathBFFD:Complexity}

The space complexity of PathBFFD is $\calO(NM+M^2)$ in total.
The time complexity of PathBFFD is $\calO(NM^3)$ per knot in general.
However, it is only $\calO(NM^2 + M^3)$ per knot 
if the matrix $A$ permits matrix-(square-)matrix multiplication with $\calO(M^2)$ 
(which can often be achieved by a transformation of the state space).
If, in addition, the initial state $X_0$ is fixed, the complexity further reduces to $\calO(NM^2)$ per knot;
this applies, in particular, to Fig.~\ref{fig:FactorGraph:DecompInput}.
However, the special structure of Fig.~\ref{fig:FactorGraph:DecompInput} 
allows an additional reduction of the complexity to $\calO(NM)$,
which will be discussed in Section~\ref{sec:RefineFig2}.

\section{Regularized-Output Path by Parametric Forward Filtering Backward Dual Deciding}
\label{sec:OutputPath}

\subsection{FFBDD and Parametric FFBDD}
\label{sec:FFBDD:Parametric}

As mentioned in Section~\ref{sec:BFFD}, MAP estimation 
in Figures \ref{fig:StatisticalModel:FactorGraph}--\ref{fig:FactorGraph:DecompOutput}
for fixed $\sigma^2$
can be computed by several forward-backward (or backward-forward) Gaussian message passing algorithms,
which can be assembled 
from the tables reproduced in Appendix~\ref{appsec:GaussianMsgTables}.
In this section, 
we use forward filtering backward dual deciding (FFBDD) \cite{Li2025},
which is dual to BFFD
and effectively solves the convex dual of (\ref{eqn:L1fittingQpen:LSSM})
by dynamic programming. %\cite{Li2025}.
The detailed algorithm is given in Algorithm~\ref{alg:FFBDD:SSM}
(and its specialization to Fig.~\ref{fig:FactorGraph:DecompOutput} is given in Section~\ref{sec:RefineFig3}).
The forward recursion of FFBDD is the standard Kalman filter, which
computes the parameters $\msgf{V}{X_n''}\in\R^{M\times M}$ and $\msgf{m}{X_n''}\in\R^M$
of the Gaussian messages $\msgf{\mu}{X_n''}$
for $n=1,\ldots,N$; 
its key steps (\ref{eqn:alg:FFBDD:forw:mX}) and (\ref{eqn:alg:FFBDD:forw:VX})
follow from (\ref{eqn:CompositeEquationForwardMean}) and (\ref{eqn:CompositeEquationForwardV}), respectively.
The backward recursion of FFBDD 
uses the dual means $\tilde\xi_{X_N}, \ldots, \tilde\xi_{X_0}$ and $\tilde\xi_{Y_N},\ldots,\tilde\xi_{Y_1}$,
which are defined as in (\ref{eqn:TabSingleEdgeGMPdxiDef});
these quantities are computed one by one, 
where, in each step, $\msgb{W}{X_n} = \msgb{V}{\tilde X_n}$ is set to zero,
which amounts to fixing the dual variable $\tilde X_n$ as explained in \cite{Li2025}
(see also Fig.~\ref{fig:FFBDD:BackwardDeciding}).
The subsequent computation of 
$\hat u_n$, $\hat y_n$, and $\hat x_0$ by (\ref{eqn:alg:FFBDD:backw:hatU})--(\ref{eqn:alg:FFBDD:hatx0}) 
follows from (\ref{eqn:TabSingleEdgeGMPm}).
Note that $\msgf{V}{Y_n}$ and $\msgf{m}{Y_n}$ in the backward recursion do not 
refer to Fig~\ref{fig:StatisticalModel:FactorGraph}, but to Fig.~\ref{fig:FFBDD:BackwardDeciding}.
The possibility of division by zero in (\ref{eqn:alg:FFBDD:forw:G}) is discussed in Section~\ref{sec:PathFFBDD:LinDep}.

Note that from the quantities computed in the forward recursion, 
only 
$c_{n}^{\mathsf{T}}\msgf{V}{X_{n}^{''}} \in\R^M$, 
$c_{n}^{\mathsf{T}}\msgf{m}{X_{n}^{''}} \in\R$, 
and $c_{n}^{\mathsf{T}}\msgf{V}{X_{n}^{''}}c_{n} \in\R$
are stored for the backward recursion.

%\begin{table}[t]
\begin{table}[tp]
	\framebox[\linewidth]{%
		\normalsize%
		\begin{minipage}{0.95\linewidth}
			\begin{trivalgorithm}[Forward Filtering Backward Dual Deciding (FFBDD) in Fig.~\ref{fig:StatisticalModel:FactorGraph}]%
				\label{alg:FFBDD:SSM}\vspace{0.5ex}\\
				Input: see Section \ref{sec:StatModelFG}\\
				Output: $\hat x_0$, $\hat u_n$, $\hat y_n$,
				  and (in this paper) also $\msgf{m}{Y_n}$ and $\msgf{V}{Y_n}$ for $n=1,\ldots,N$
				\\[-1ex]
				\hrule
				\vspace{-1ex}
				\begin{multicols}{2}
					\begin{enumerate}[1.]
						\item
						Let $\msgf{V}{X_{0}}=\sigma^{2}Q_{0}^{-1}$ and $\msgf{m}{X_{0}}=\breve{x}_{0}$.
						\item
						\pkw{for} $n=1$ \pkw{to} $N$, compute the following:
						\begin{IEEEeqnarray}{rCl}
							\msgf{m}{X_{n}^{'''}} & = &A\msgf{m}{X_{n-1}},
							\label{eqn:alg:FFBDD:forw:mXppp}\\
								\msgf{V}{X_{n}^{'''}} & = &A\msgf{V}{X_{n-1}}A^{\mathsf{T}}
							\label{eqn:alg:FFBDD:forw:VXppp}
							%\IEEEeqnarraynumspace
						\end{IEEEeqnarray}
						\begin{IEEEeqnarray}{rCl}
									\msgf{m}{X_{n}^{''}} & = &\msgf{m}{X_{n}^{'''}} + b_{n}\msgf{m}{U_{n}}
							\label{eqn:alg:FFBDD:forw:mXpp}\\
							\msgf{V}{X_{n}^{''}} & = &\msgf{V}{X_{n}^{'''}} + b_{n}\msgf{V}{U_{n}}b_{n}^{\mathsf{T}}
							\label{eqn:alg:FFBDD:forw:VXpp}
							%\IEEEeqnarraynumspace
						\end{IEEEeqnarray}
						\begin{IEEEeqnarray}{rCl}
							G_{n} & = & \begin{cases}
								0,  & \text{if $\msgb{W}{Y_n}=0$}\\ 
								\big(c_{n}^{\mathsf{T}}\msgf{V}{X_{n}^{''}}c_{n}\big)^{-1},  & \text{if $\msgb{V}{Y_n}=0$}\\
								\text{by (\ref{eqn:FFBDD:Gn:General})},   & \text{else} 
								%\big(\msgf{W}{U_{n}} + b_{n}^{\mathsf{T}}\msgb{W}{X_{n}^{''}}b_{n}\big)^{-1}
							\end{cases}
							\label{eqn:alg:FFBDD:forw:G}\\
							g_{n} & = & \begin{cases}
								c_{n}\msgb{\xi}{Y_{n}}, &  \text{if $\msgb{W}{Y_{n}}=0$},\\
								% b_{n}    \big(b_{n}^{\mathsf{T}}\msgb{W}{X_{n}^{''}}b_{n}\big)^{-1}\big(\msgf{\xi}{U_{n}}+b_{n}^{\mathsf{T}}\msgb{\xi}{X_{n}^{''}}\big), & \msgf{W}{U_{n}}=0,\\
								c_{n}G_{n}\big(\msgb{m}{Y_{n}}-c_{n}^{\mathsf{T}}\msgf{m}{X_{n}^{''}}\big), & \text{else}
							\end{cases}
							\label{eqn:alg:FFBDD:forw:g}
						\end{IEEEeqnarray}
						\begin{IEEEeqnarray}{rCl}
							\msgf{m}{X_{n}} & = &\msgf{m}{X_{n}^{''}} + \msgf{V}{X_{n}^{''}}g_{n}
							\label{eqn:alg:FFBDD:forw:mX}\\
						\msgf{V}{X_{n}} & = & \msgf{V}{X_{n}^{''}} - \msgf{V}{X_{n}^{''}} c_{n}G_{n}c_{n}^{\mathsf{T}}\msgf{V}{X_{n}^{''}}
							\label{eqn:alg:FFBDD:forw:VX}
						\end{IEEEeqnarray}
						\hspace{1.5em}Store only $c_{n}^{\mathsf{T}}\msgf{V}{X_{n}^{''}}, c_{n}^{\mathsf{T}}\msgf{m}{X_{n}^{''}}$ and $c_{n}^{\mathsf{T}}\msgf{V}{X_{n}^{''}}c_{n}$.%
						\item[]\rule{0em}{2.5ex} 
						\pkw{end for}\\
						\newpage
					    %\rule{0em}{12ex}  % to end the column here

						\item \pkw{if} $Q_N=0$ 
							\begin{enumerate}
							\item[]
							Let $\tilde\xi_{X_N}=0$.
							\end{enumerate}
						\item
						\pkw{else}
							\begin{enumerate}
							\item[]
							Let $\msgb{V}{X_{N}}=\sigma^{2}Q_{N}^{-1}$ and $\msgb{m}{X_{N}}=\breve{x}_{N}$\\
							and compute $\tilde\xi_{X_N}$ by solving
							\begin{equation} \label{eqn:DualMarginalEstimateTerminalState}
								\big(\msgf{V}{X_{N}}+\msgb{V}{X_{N}}\big)\tilde{\xi}_{X_{N}}=\msgf{m}{X_{N}} -  \msgb{m}{X_{N}}
							\end{equation}
							\end{enumerate}
						\item[]
						\pkw{end if}
						\item
						\pkw{for} $n=N$ \pkw{to} $1$, compute the following:
						\begin{IEEEeqnarray}{rCl}	
							%\msgf{W}{Y_{n}} &= &\big(c_{n}^{\mathsf{T}}\msgf{V}{X_{n}^{''}}c_{n}\big)^{-1}
							\msgf{V}{Y_{n}} &= & c_{n}^{\mathsf{T}}\msgf{V}{X_{n}^{''}}c_{n}
							\label{eqn:alg:FFBDD:backw:msgfWY}\\
						%\msgf{\xi}{Y_{n}} &=  & \msgf{W}{Y_{n}}\big(c_{n}^{\mathsf{T}}\msgf{m}{X_{n}^{''}}-c_{n}^{\mathsf{T}}\msgf{V}{X_{n}^{''}}	\hat{\tilde{x}}_{n}\big) \\
						\msgf{m}{Y_{n}} &=  &  c_n^\T \msgf{m}{X_{n}^{''}} - c_n^\T \msgf{V}{X_{n}^{''}}	\tilde\xi_{X_n}
							\label{eqn:alg:FFBDD:fwd:msgfXiY}
						\end{IEEEeqnarray}
					    \begin{equation} \label{eqn:alg:FFBDD:backw:hatDualY}
							%\hat{\tilde{y}}_n 
							\tilde{\xi}_{Y_n}
							    = \begin{cases}
								-\msgb{\xi}{Y_n}, & \text{if $\msgb{W}{Y_n}=0$} \\
								 %\msgf{\xi}{Y_n}  - \msgf{W}{Y_n} \msgb{m}{Y_n},  & \text{if $\msgb{V}{Y_n}=0$} \\
								 \big(\msgf{V}{Y_n}\big)^{-1}  \Big( \msgf{m}{Y_n}  - \msgb{m}{Y_n} \Big),  & \text{if $\msgb{V}{Y_n}=0$} \\
								\text{by (\ref{eqn:FFBDD:decideDualYn:General}),} 
								& \text{else}
							\end{cases}
						\end{equation}
						\begin{IEEEeqnarray}{rCl}	
							\tilde\xi_{X_n''}  &=&  \tilde\xi_{X_n} + c_{n} \tilde\xi_{Y_n}
							\label{eqn:alg:FFBDD:backw:hatDualXpp} \\
							\tilde\xi_{X_{n-1}} & = &  A^{\mathsf{T}} \tilde\xi_{X_{n}''}
							\label{eqn:alg:FFBDD:backw:hatDualX}
						\end{IEEEeqnarray}
						\begin{IEEEeqnarray}{rCl}	
							\hat{u}_{n} & = &  \msgf{m}{U_{n}} - \msgf{V}{U_{n}} b_n^\T \tilde\xi_{X_n''}
							\label{eqn:alg:FFBDD:backw:hatU}\\
						  \hat{y}_n  
						    & = & \begin{cases}
						           \msgb{m}{Y_n},  &  \text{if $\msgb{V}{Y_n}=0$}\\
						           \msgf{m}{Y_n} - \msgf{V}{Y_n} \tilde\xi_{Y_n},  &  \text{else}
						          \end{cases}
							\label{eqn:alg:FFBDD:backw:hatY}
						\end{IEEEeqnarray}
						\item[]
						\pkw{end for}

						\item
						Compute
						\begin{equation} \label{eqn:alg:FFBDD:hatx0}
						\hat x_0 = \msgf{m}{X_0} - \msgf{V}{X_0}{} \tilde\xi_{X_0}.
						\end{equation}

					\end{enumerate}
					%\rule{0em}{0ex}  % visual adjust
				\end{multicols}
				\vspace{-1ex}% 
				\hrule
				\vspace{1ex}
				
				Concerning (\ref{eqn:alg:FFBDD:forw:G}), the general case
				\begin{equation} \label{eqn:FFBDD:Gn:General}
					G_{n}  = \big(\msgb{V}{Y_{n}}  + c_{n}^{\mathsf{T}}\msgf{V}{X_{n}^{''}}c_{n}\big)^{-1}
				\end{equation}
				is not used in this paper.
				\medskip
				
				Concerning (\ref{eqn:alg:FFBDD:backw:hatDualY}), the general case
				\begin{equation} \label{eqn:FFBDD:decideDualYn:General}
					\tilde{\xi}_{Y_n} 
					= \big( \msgf{V}{Y_n} + \msgb{V}{Y_n} \big)^{-1} \big( \msgf{m}{Y_n} - \msgb{m}{Y_n} \big)
				\end{equation}
				is not used in this paper.
			\end{trivalgorithm}
			
		\end{minipage}%
	}
\end{table}

\begin{figure*}[tp]
\centering
\begin{tikzpicture}[x=1mm, y=1mm, >=latex]
\tikzset{%
smallbox/.style = {draw, rectangle, inner sep=0mm, minimum width=4mm, minimum height=4mm},
midbox/.style = {draw, rectangle, minimum width=5.5mm, minimum height=5.5mm, inner sep=1mm},
blobbox/.style = {draw, fill=black, rectangle, inner sep=0mm, minimum width=1.75mm, minimum height=1.75mm},
xibox/.style = {draw, pattern=north east lines, inner sep=0mm, minimum width=1.75mm, minimum height=1.75mm},
}
\node[smallbox, label={below:$p(x_0)$}] (pX0) {};
\draw[->] (pX0) -- node[above]{$X_0$} +(11,0);
\draw (pX0)+(14.5,0) node {$\ldots$};
\draw (pX0)+(30,0) node[midbox] (A) {$A$};
\draw[->] (A)+(-12,0) -- node[above, pos=0.4]{$X_{n-1}$} (A);

\draw (A)+(15,0) node[smallbox] (add) {$+$};
%\draw[->] (A) -- node[above]{\optional{$X_n'''$}} (add);
\draw[->] (A) -- node[above]{$X_n'''$} (add);
\draw (add)+(0,11) node[midbox] (bn) {$b_n$};
\draw[->] (bn) -- (add);
\draw (bn)+(0,12) node[smallbox,label={left:$p(u_n)$}] (pUn) {};
\draw[->] (pUn) -- node[left] {$U_n$} (bn);

\draw (add)+(7.5,-10) 
      node[draw, dashed, anchor=south east, 
           minimum width=60mm, minimum height=40mm, 
           label=below:$\msgf{\mu}{X_n''}$] {};

\draw (add)+(20,0) node[smallbox] (equ) {$=$};
%\draw[->] (add) -- node[above] {\optional{$X_n''$}} (equ);
\draw[->] (add) -- node[above,pos=0.65] {$X_n''$} (equ);
\draw (equ)+(0,-12) node[midbox] (cn) {$c_n^\T$};
\draw[->] (equ) -- node[left] {$X_n'$} (cn);
\draw (cn)+(0,-12) node[smallbox,label={right:$p(\breve y_n \cond y_n)$}] (pYn) {};
\draw[->] (cn) -- node[left] {$Y_n$} (pYn);

\draw (equ)+(12.5,0) node[xibox, label=right:$\tilde\xi_{X_n}$] (XitildeXn) {};
\draw[->] (equ) -- node[above] {$X_n$} (XitildeXn);
\end{tikzpicture}
\newcommand{\xibox}{%
\begin{tikzpicture}[scale=0.1]
\protect\node[draw, pattern=north east lines, inner sep=0mm, minimum width=1.75mm, minimum height=1.75mm] {};
\end{tikzpicture}%
}
\vspace{1ex}
\caption{\label{fig:FFBDD:BackwardDeciding}%
Illustrating FFBDD.
The symbol ``$\,\xibox\,$'' denotes a factor (\ref{eqn:GenGaussian}) 
with $\msgb{W}{X_n}=0$.
}
\end{figure*}

The proposed algorithm will repeatedly call Algorithm~\ref{alg:FFBDD:SSM} (FFBDD)
with the following conditions satisfied:
\begin{enumerate}
\item
$\msgf{V}{X_0} = \sigma^{2} Q_0^{-1}$ and $\msgf{m}{X_0} = \breve x_0$.
\item
$\msgb{V}{X_N} = \sigma^{2} Q_N^{-1}$ and $\msgb{m}{X_N} = \breve x_N$.
\item
For $n=1,\ldots,N,$ 
$\msgf{V}{U_n} = \sigma^{2}$ and $\msgf{m}{U_n} = 0$.
\item
For $n=1,\ldots,N,$ 
either $\msgb{W}{Y_n}=0$ and $\msgb{\xi}{Y_n}$ does not depend on $\sigma$,
or $\msgb{V}{Y_n}=0$ and $\msgb{m}{Y_n}$ does not depend on $\sigma$, cf.~(\ref{eqn:CostPieces:NUP:yn}).
\end{enumerate}

\begin{proposition} \label{prop:FFBDD:affine}
When FFBDD is called with the above conditions satisfied, 
then, for all $n$, 
\renewcommand{\theenumi}{\roman{enumi}}%
\begin{enumerate}
\item %\label{eqn:prop:BFFD:affine:msgbXn}
$\msgf{V}{X_n}$ is a linear function of $\sigma^{2}$ and $\msgf{m}{X_n}$ is an affine function of $\sigma^{2}$;
\item
the dual mean $\tilde\xi_{X_n}$ is an affine function of~$\sigma^{-2}$;
\item
$\msgf{V}{Y_n}$ is a linear function of $\sigma^2$ and $\msgf{m}{Y_n}$ is an affine function of $\sigma^2$;
\item
the posterior mean $m_{Y_n}$ (= the estimate $\hat y_n$) is an affine function of $\sigma^2$.
\end{enumerate}
\vspace{-2ex}
\end{proposition}
The proof of Proposition~\ref{prop:FFBDD:affine} amounts to verifying and tracking 
the linear or affine representation at each step of FFBDD, 
which is detailed in Appendix~\ref{appsec:ParametricFFBDD};
these detailed steps are also required for carrying out the proposed algorithm.

\subsection{Proposed Algorithm (PathFFBDD)}
\label{sec:PathFFBDD}

\begin{table}[t]
\framebox[\linewidth]{%
\normalsize%
\begin{minipage}{0.95\linewidth}
%
%\begin{trivalgorithm}[Path Tracking with Parametric Forward Filtering Backward Dual Deciding (PathFFBDD)]%
\begin{trivalgorithm}[PathFFBDD]%
\label{alg:PathFFBDD}\vspace{0.5ex}\\
Input: see main text.\\
Output: the knots $\sigma_0^2 = 0, \sigma_1^2, \ldots, \sigma_{\ell-1}^2, \sigma_\ell^2=\infty$
and the affine representation of $\hat y_n(\sigma^2)$ between these points.
\\[-1ex]
\hrule
\begin{algorithmic}[1]
\STATE Let $\sigma_0^2=0$.
%\STATE Run BFFD as in Section~\ref{sec:BFFD} 
\STATE Run Algorithm~\ref{alg:FFBDD:SSM} 
       with $\msgb{\xi}{Y_n}=0$ and $\msgb{W}{Y_n}=0$ for all $n$, and with arbitrary fixed $\sigma^2>0$.
\STATE For all $n$, record the resulting $\hat y_n$ as the final result $\hat y_n(\sigma_0^2)$.
\STATE For every $n$, set the active subdomain of $\kappa_n(y_n)$ to the subdomain that contains $\hat y_n$.
%
%\STATE Set $\ell=0$.
\STATE $\ell \gets 0$.
\REPEAT
   %\STATE Set $\ell = \ell+1$.
   \STATE $\ell \gets \ell+1$.
   \STATE \label{line:PFFBDD:callParFFDDD}
          Run Algorithm~\ref{alg:FFBDD:SSM} 
          with $\msgb{\xi}{Y_n}$ and $\msgb{W}{Y_n}$ (or $\msgb{m}{Y_n}$ and $\msgb{V}{Y_n}$)
          according to the (fixed) active segments of $\kappa_n(y_n)$,
          and with $\msgf{m}{X_n}$, $\msgf{V}{X_n}$,
          $\msgf{m}{Y_n}$, $\msgf{V}{Y_n}$, $\hat y_n$, etc.,
          as functions of $\sigma^2$
          (as detailed in Section~\ref{sec:FFBDD:Parametric}).
          %We thus obtain new $\hat y_n(\sigma^2)$, $\msgf{m}{Y_n}(\sigma^2)$, $\msgf{V}{Y_n}(\sigma^2)$ for all $n$.
          \\[0.5ex]
   \STATE \label{line:alg:PFFBDD:eoas}
          For every $n$, let
		  \begin{equation} \label{eqn:PathFFBDD:hatDual}
		    \varestim{y}_n(\sigma^2) = \argmax_{y_n} \msgf{\mu}{Y_n}(y_n) \exp(-\kappa_n(y_n)),
		  \end{equation}
		  where $\msgf{\mu}{Y_n}(y_n)$ is Gaussian with parameters $\msgf{m}{Y_n}(\sigma^2)$ and $\msgf{V}{Y_n}(\sigma^2)$,
		  and determine 
          \begin{equation} \label{eqn:alg:FFBDD:sigman}
            \sigma_{\ell,n}^2 = \sup \big\{ \sigma^2 : \text{$\varestim{y}_n(\sigma^2)$ is in the active subdomain of $\kappa_n(y_n)$} \big\},
          \end{equation}
          as detailed in Section~\ref{sec:PathFFBDD}.
          \\[0.5ex]
   \STATE Let $\sigma_\ell^2 = \min_n\big\{ \sigma_{\ell,n}^2 :  \sigma_{\ell,n}^2 > \sigma_{\ell-1}^2 \big\}$. 
   \IF{ $\sigma_\ell^2 < \infty$ }
       \STATE 
          For the index\footnote{% 
          According to the unique path assumption of Section~\ref{sec:PathFFBDD:LinDep}, there is only one such index.} 
          $n$
          such that $\sigma_{\ell,n}^2 = \sigma_\ell^2$,
          change the active segment and subdomain of $\kappa_n(y_n)$ as follows:
          if the present active segment is a line segment,
          then the next active segment is the point segment whose subdomain 
          contains $\varestim{y}_n(\sigma_\ell^2)$;
          if the present active segment is a point segment,
          then the next active segment is the line segment whose subdomain 
          contains $\varestim{y}_n(\sigma_\ell^2+\varepsilon)$
          with positive $\varepsilon \rightarrow 0$.
   \ENDIF
\UNTIL $\sigma_\ell^2 = \infty$.
\end{algorithmic}
\end{trivalgorithm}
\vspace{1ex}

\hrule

\end{minipage}%
}
\end{table}

In this section, the cost function for $Y_n$ will be denoted 
$\kappa_n(y_n) \eqdef \kappa(y_n-\breve y_n)$.
The proposed algorithm is Algorithm~\ref{alg:PathFFBDD},
which is very similar to Algorithm~\ref{alg:PathBFFD};
the essential novelty of Algorithm~\ref{alg:PathFFBDD} 
is its use of parametric FFBDD. 
Note that BFFD cannot be used here 
since it does not provide $\msgf{m}{Y_n}$ and $\msgf{V}{Y_n}$
for the dependent variables $Y_n$.
Likewise, FFBDD cannot be used in Section~\ref{sec:InputPath} 
since it does not produce $\msgb{m}{U_n}$ and $\msgb{V}{U_n}$ for the independent variables $U_n$.

Concerning Line \ref{line:PFFBDD:callParFFDDD} of Algorithm~\ref{alg:PathFFBDD}, we note:
%FFBDD is called as follows:
\begin{itemize}
\item
if the active segment of $\kappa_n(y_n)$ is a line segment, 
FFBDD is called with $\msgb{W}{Y_n}=0$ and $-\msgb{\xi}{Y_n}$ is the slope of $\kappa_n(y_n)$;
\item
if the active segment of $\kappa_n(y_n)$ is a point segment,
FFBDD is called with $\msgb{V}{Y_n}=0$ and $\msgb{m}{Y_n}$ equals the pertinent subdomain of $\kappa_n(y_n)$.
\end{itemize}
E.g., for $\kappa_n$ as in (\ref{eqn:CostPieces}), we have
\begin{equation}\label{eqn:CostPieces:NUP:yn}
\begin{array}{c@{\hspace{1.7em}}cccc}
\text{subdomain} & \msgb{m}{Y_n} & \msgb{V}{Y_n} & \msgb{W}{Y_n} & \msgb{\xi}{Y_n} 
\\
\hline
y_n < a  &     &    &  0  &  1 \\
y_n = a     &  a  &  0  &   &  \\
a < y_n < b   &     &   &  0  &  -1 \\
y_n = b     &  b  &  0  &   &  \\
y_n >b  &     &    &  0  &  -2
\end{array}
\end{equation}

The quantity $\varestim{y}_n$ in (\ref{eqn:PathFFBDD:hatDual}) 
is an estimate of $Y_n$ that uses the whole cost function $\kappa_n(y_n)$
while $\hat y_n$ from (\ref{eqn:alg:FFBDD:backw:hatY}) 
uses only the presently active segment of $\kappa_n(y_n)$.
%Between knots, 
For $\sigma^2_{\ell-1} \leq \sigma^2 < \sigma^2_{\ell,n}$,
we have $\varestim{y}_n(\sigma^2) = \hat y_n(\sigma^2)$.

Line~\ref{line:alg:PFFBDD:eoas} of Algorithm~\ref{alg:PathFFBDD}
is aided by Table~\ref{tbl:DecidingRules}, which shows 
both $\varestim{y}_n$
and the conditions on $\msgf{m}{Y_n}$ and $\msgf{V}{Y_n}$
for (\ref{eqn:PathFFBDD:hatDual}) to lie
in the different subdomains of $\kappa_n(y_n)$.
These conditions allow to easily determine (\ref{eqn:alg:FFBDD:sigman}),
as illustrated by the following example.

\begin{example} \label{example:FFBDD:nextSubdomain}
Let $\kappa_n(y_n) = |y_n|$. Assume that $\msgf{m}{Y_n} = \sigma^2 + 2$ and $\msgf{V}{Y_n} = 2\sigma^2$.
From Table~\ref{tbl:DecidingRules}, we have
\begin{equation} \label{eqn:FFBDD:ExSubdomainsSigma}
\begin{array}{cccc}
\toprule
\kappa_n(y_n)  &  \text{subdomain}  &  \varestim{y}_n(\sigma^2)  &  \text{condition} \\
\midrule
 -y_n  &  y_n < 0  &  3\sigma^2 + 2  &  \sigma^2 < -2/3  \\
   0   &  y_n = 0  &  0              &  -2/3 \leq \sigma^2 \geq 2  \\
  y_n  &  y_n > 0  &  2 - \sigma^2   & \sigma^2 < 2
\\
\bottomrule
\end{array}
\vspace{1ex}
\end{equation}

\begin{itemize}
\item 
Suppose the active subdomain of $\kappa_n(y_n)$ is $\{ y_n\colon y_n > 0\}$.
From (\ref{eqn:FFBDD:ExSubdomainsSigma}), 
$\varestim{y}_n$ is in the active subdomain
if $\sigma^2 < 2$.
Thus $\sigma_{\ell,n}^2 = 2$.
\item
Suppose the active subdomain of $\kappa_n(y_n)$ is $\{ 0 \}$.
From (\ref{eqn:FFBDD:ExSubdomainsSigma}),
$\varestim{y}_n$ is in the active subdomain
if $ \sigma^2 \geq 2$.
Thus $\sigma_{\ell,n}^2 = \infty$.
\item
Suppose the active subdomain of $\kappa_n(y_n)$ is $\{ y_n\colon y_n < 0\}$.
From (\ref{eqn:FFBDD:ExSubdomainsSigma}), $\varestim{y}_n$ cannot be in the active subdomain.
This case cannot occur. 
\end{itemize}
\vspace{-2ex}
\end{example}

\subsection{Unique Path Assumption}
\label{sec:PathFFBDD:LinDep}

Division by zero in (\ref{eqn:alg:FFBDD:forw:G}) is possible.
As in Section~\ref{sec:PathBFFD:LinDep}, this happens if and only if 
the regularization path is not unique. 
For $F$ as in (\ref{eqn:L1fittingQpen}), 
this condition can equivalently be formulated as follows. 
\begin{proposition}[Unique Path Condition for Regularized Output]
% $J(\ell)$   $F_{J(\ell)}$
Let $J_\ell\subset\{1,\ldots,N\}$ (with $N=L$) be the set of indices $n$ such that, in round $\ell$ of PathFFBDD,
the active subdomain of $\kappa_n$ 
is a point segment. 
%Let $F_{J_\ell}$ be the submatrix of $F$ comprising the corresponding columns of $F$. 
Then the regularization path is unique if and only if, for every $\ell$,
the rows of $F$ with indices in $J_\ell$ 
are linearly independent.
\end{proposition}
This condition is well known in the literature and usually assumed to hold \cite{Hastie2004}.
As in Section~\ref{sec:PathBFFD:LinDep},
modifying the proposed algorithm to handle non-unique paths properly is possible, 
but complicated and irrelevant for many applications (cf.\ Section~\ref{sec:NumericalExperiments}).
Again, we here choose to follow the tradition and simply assume the path to be unique.

\subsection{Complexity}

The space complexity of PathFFBDD is $\calO(NM+M^2)$ in total.
The time complexity of PathFFBDD is $\calO(NM^3)$ per knot in general.
However, it is only $\calO(NM^2 + M^3)$ per knot 
if the matrix $A$ permits matrix-(square-)matrix multiplication with $\calO(M^2)$. 
%(which can often be achieved by a transformation of the state space).
If, in addition, $Q_N=0$, then the complexity further reduces to $\calO(NM^2)$ per knot;
this applies, in particular, to Fig.~\ref{fig:FactorGraph:DecompOutput}.
However, the special structure of Fig.~\ref{fig:FactorGraph:DecompOutput}
allows an additional reduction of the complexity to $\calO(NM)$,  
which will be discussed in Section~\ref{sec:RefineFig3}.

\section{Regularized Input: Refinements for Generic $F$ (Fig.~\protect\ref{fig:FactorGraph:DecompInput})}
\label{sec:RefineFig2}

%\begin{table}[tp]
\begin{table}[p]
	\framebox[\linewidth]{%
		\normalsize%
		\begin{minipage}{0.95\linewidth}
			\begin{trivalgorithm}[BFFD in Fig.~\ref{fig:FactorGraph:DecompInput}]%
				\label{alg:BFFD:DecompInput}\vspace{0.5ex}\\
				Input: see Sections \ref{sec:StatModelFG}\\
				Output: $\hat u_n$ and (in this paper) $\msgb{m}{U_n}$ and $\msgb{V}{U_n}$ for $n=1,\ldots,N$
				\\[-1ex]
				\hrule
				\vspace{-2ex}
				\begin{multicols}{2}
					%\emph{Backward filtering:}
					\begin{enumerate}[1.]
						\item \label{alg:BFFD:DecompInput:Step:InitBF}
						Let $\msgb{W}{X_{N}}=\sigma^{-2}I$ and $\msgb{\xi}{X_{N}}=\sigma^{-2}\breve{x}_{N}$.
						\item \label{alg:BFFD:DecompInput:Step:BF}
						\pkw{for} $n=N$ \pkw{to} $1$, compute the following:
						\begin{IEEEeqnarray}{rCl}
							H_{n} & = & \begin{cases}
								0,  & \text{if $\msgf{V}{U_n}=0$}\\ 
								\big(b_{n}^\T \msgb{W}{X_{n}}b_{n}\big)^{-1},  & \text{if $\msgf{W}{U_n}=0$}\\
								\text{by (\ref{eqn:BFFD:Hn:General})},   & \text{else} 
								%\text{by (\ref{eqn:alg:BFFD:DecompInput:backw:H:General})},   & \text{else} 
								%\big(\msgf{W}{U_{n}} + b_{n}^{\mathsf{T}}\msgb{W}{X_{n}^{''}}b_{n}\big)^{-1}
							\end{cases}
							\label{eqn:alg:BFFD:DecompInput:backw:H}\\
							h_{n} & = & \begin{cases}
								b_{n}\msgf{m}{U_{n}}, &  \msgf{V}{U_{n}}=0\\
								b_{n}H_{n}\big(\msgf{\xi}{U_{n}}+b_{n}^\T \msgb{\xi}{X_{n}}\big), & \text{else}
							\end{cases}
							\label{eqn:alg:BFFD:DecompInput:backw:h}
						\end{IEEEeqnarray}
						\begin{IEEEeqnarray}{rCl}
							\msgb{\xi}{X_{n-1}} & = & \msgb{\xi}{X_{n}} - \msgb{W}{X_{n}}h_{n}
							\label{eqn:alg:BFFD:DecompInput:backw:xiXppp}\\
							\msgb{W}{X_{n-1}} & = & \msgb{W}{X_{n}} - \msgb{W}{X_{n}}b_{n}H_{n}b_{n}^\T \msgb{W}{X_{n}}
							\label{eqn:alg:BFFD:DecompInput:backw:WXppp}
						\end{IEEEeqnarray}
						\hspace{1.5em}Store only $b_{n}^\T \msgb{W}{X_{n}}, b_{n}^\T \msgb{\xi}{X_{n}}$ and $b_{n}^\T \msgb{W}{X_{n}}b_{n}$.%
						\item[] 
						\pkw{end for}
						\newpage
						
						%\item[]
						%\emph{Forward deciding:}
						\item \label{alg:BFFD:DecompInput:Step:InitFD}
						Let $\hat{x}_{0} = 0$. 
						\item \label{alg:BFFD:DecompInput:Step:FD}
						\pkw{for} $n=1$ \pkw{to} $N$, compute the following:
						\begin{IEEEeqnarray}{rCl}	
							\msgb{V}{U_{n}} & = & \big(b_{n}^\T \msgb{W}{X_{n}}b_{n}\big)^{-1}
							\label{eqn:alg:BFFD:DecompInput:fwd:msgbVU}\\
							\msgb{m}{U_{n}} & = & \msgb{V}{U_{n}} \big(b_{n}^\T \msgb{\xi}{X_{n}} - b_{n}^\T \msgb{W}{X_{n}}\hat{x}_{n-1} \big)
							\label{eqn:alg:BFFD:DecompInput:fwd:msgbU}
						\end{IEEEeqnarray}
						\begin{equation} \label{eqn:alg:BFFD:DecompInput:fwd:hatU}
							\hat u_n = \begin{cases}
								\msgf{m}{U_n}, & \text{if $\msgf{V}{U_n}=0$} \\
								\msgb{m}{U_n} + \msgb{V}{U_n} \msgf{\xi}{U_n} ,  & \text{if $\msgf{W}{U_n}=0$} \\
								\text{by (\ref{eqn:BFFD:decideUn:General}),} & \text{else}
								%\text{by (\ref{eqn:BFFD:DecompInput:decideUn:General}),} & \text{else.}
							\end{cases}
						\end{equation}
						\begin{IEEEeqnarray}{rCl}	
							\hat{x}_{n} & = &  \hat{x}_{n-1}+ b_{n}\hat{u}_{n}
							\label{eqn:alg:BFFD:DecompInput:fwd:hatX} 
						\end{IEEEeqnarray}
						\item[]
						\pkw{end for}
					\end{enumerate}
					%\rule{0em}{3.0ex}  % visual adjust
				\end{multicols}
				\vspace{-1ex}% 
				\hrule
				\vspace{1ex}

				Concerning (\ref{eqn:alg:BFFD:DecompInput:backw:H})
				and (\ref{eqn:alg:BFFD:DecompInput:fwd:hatU}),
				the ``else'' case
				is not used in this paper.
			\end{trivalgorithm}
			
		\end{minipage}%
	}
\end{table}

\begin{table}[tp]
	\framebox[\linewidth]{%
		\normalsize%
		\begin{minipage}{0.95\linewidth}
			\begin{trivalgorithm}[Simplified BFFD Update for PathBFFD in Fig.~\ref{fig:FactorGraph:DecompInput}]%
				\label{alg:BFFD:DecompInput:SpecUpd}\vspace{0.5ex}\\
				Input: see Sections \ref{sec:StatModelFG} and \ref{sec:BFFD},
				       and precomputed $b_n^\T \msgb{W}{X_n} = \big( \msgb{W}{X_n} b_n \big)^\T$ for $n=1,\ldots,N$\\
				Output: $\hat u_n$, $\msgb{m}{U_n}$ and $\msgb{V}{U_n}$ for $n=1,\ldots,N$
				\\[-1ex]
				\hrule
				\vspace{-2ex}
				\begin{multicols}{2}
					%\emph{Backward filtering:}
					\begin{enumerate}[1.]
						\item \label{alg:BFFD:DecompInput:SpecUpd:Step:InitBF}
						Let $\msgb{\xi}{X_{N}}=\sigma^{-2}\breve{x}_{N}$.
						\item \label{alg:BFFD:DecompInput:SpecUpd:Step:BF}
						\pkw{for} $n=N$ \pkw{to} $1$, compute the following:
						\begin{IEEEeqnarray}{rCl}
							\alpha_n & = & \big(b_{n}^\T \msgb{W}{X_{n}}b_{n}\big)^{-1}
							\label{eqn:alg:BFFD:DecompInput:SpecUpd:backw:H}\\
							\beta_n & = & \begin{cases}
								\msgf{m}{U_{n}}, &  \text{if $\msgf{V}{U_{n}}=0$}\\
								\alpha_n \big(\msgf{\xi}{U_{n}}+b_{n}^\T \msgb{\xi}{X_{n}}\big), & \text{if $\msgf{W}{U_{n}}=0$}
							\end{cases}
							\IEEEeqnarraynumspace
							\label{eqn:alg:BFFD:DecompInput:SpecUpd:backw:h}
						\end{IEEEeqnarray}
						\begin{IEEEeqnarray}{rCl}
							\msgb{\xi}{X_{n-1}} & = & \msgb{\xi}{X_{n}} - \beta_n \msgb{W}{X_{n}} b_n
							\label{eqn:alg:BFFD:DecompInput:SpecUpd:backw:xiXppp}
						\end{IEEEeqnarray}
						\hspace{1.5em}Store $b_{n}^\T \msgb{\xi}{X_{n}}$ and $\alpha_n$.%
						\item[] \vspace{0.5ex} % visual
						\pkw{end for}

                        \rule{0em}{0ex}  % visual adjust
						%\item[]
						%\emph{Forward deciding:}
						\item \label{alg:BFFD:DecompInput:SpecUpd:Step:InitFD}
						Let $\hat{x}_{0} = 0$. 
						\item \label{alg:BFFD:DecompInput:SpecUpd:Step:FD}
						\pkw{for} $n=1$ \pkw{to} $N$, compute the following:
						\begin{IEEEeqnarray}{rCl}	
							\msgb{V}{U_{n}} & = & \alpha_n
							\label{eqn:alg:BFFD:DecompInput:SpecUpd:fwd:msgbVU}\\
							\msgb{m}{U_{n}} & = & \msgb{V}{U_{n}} \big(b_{n}^\T \msgb{\xi}{X_{n}} - b_{n}^\T \msgb{W}{X_{n}}\hat{x}_{n-1} \big)
							\label{eqn:alg:BFFD:DecompInput:SpecUpd:fwd:msgbU}
						\end{IEEEeqnarray}
						\begin{equation} \label{eqn:alg:BFFD:DecompInput:SpecUpd:fwd:hatU}
							\hat u_n = \begin{cases}
								\msgf{m}{U_n}, & \text{if $\msgf{V}{U_n}=0$} \\
								\msgb{m}{U_n} + \msgb{V}{U_n} \msgf{\xi}{U_n} ,  & \text{if $\msgf{W}{U_n}=0$}
								%\text{by (\ref{eqn:BFFD:DecompInput:decideUn:General}),} & \text{else.}
							\end{cases}
						\end{equation}
						\begin{IEEEeqnarray}{rCl}	
							\hat{x}_{n} & = &  \hat{x}_{n-1}+ b_{n}\hat{u}_{n}
							\label{eqn:alg:BFFD:DecompInput:SpecUpd:fwd:hatX} 
						\end{IEEEeqnarray}
						\item[]
						\pkw{end for}
					\end{enumerate}
					%\rule{0em}{0ex}  % visual adjust
				\end{multicols}
				\vspace{-2ex}   % visual adjust
			\end{trivalgorithm}
			
		\end{minipage}%
	}
\end{table}

\begin{table}[tp]
\framebox[\linewidth]{%
\normalsize%
\begin{minipage}{0.95\linewidth}
\begin{trivalgorithm}[PathBFFD for Fig.~\ref{fig:FactorGraph:DecompInput} (Regularized Least Squares)]%
\label{alg:PathBFFD4LS}\vspace{0.5ex}\\
The algorithm is Algorithm~\ref{alg:PathBFFD} with the following modifications.
\begin{enumerate}
\renewcommand{\theenumi}{\alph{enumi}}
\item In Step~\ref{line:alg:PBFFD:firstrunBFFD},
``Run Algorithm~\ref{alg:BFFD:SSM} \ldots'' 
is replaced by
``Run Algorithm~\ref{alg:BFFD:DecompInput} \ldots''.
\item 
Between Steps~\ref{line:alg:PBFFD:initactivesubdomain}  and \ref{line:alg:PBFFD:ell0},
the following steps are added:
 \begin{enumerate}
 \renewcommand{\theenumii}{\roman{enumii}}
 \item Rearrange and re-index the sequence of variables $U_1,\ldots,U_N$ such that (\ref{eqn:pathLS:sortedcond}) holds.
 \item Recompute $b_{n}^\T \msgb{W}{X_{n}}$ for all $n$  
       (as functions of $\sigma^2$)
       by the backward recursion (Steps \ref{alg:BFFD:DecompInput:SpecUpd:Step:InitBF} and \ref{alg:BFFD:DecompInput:SpecUpd:Step:BF})
       of Algorithm~\ref{alg:BFFD:DecompInput}.
 \end{enumerate}
\item 
In Step~\ref{line:alg:PBFFD:rerunBFFD}, ``Run Algorithm~\ref{alg:BFFD:SSM}\ldots''
is replaced by ``Run Algorithm~\ref{alg:BFFD:DecompInput:SpecUpd}\ldots''.
\item
Between Steps \ref{line:alg:PBFFD:iftext} and \ref{line:alg:PBFFD:endif},
the following step is added:
 \begin{enumerate}
 \renewcommand{\theenumii}{\roman{enumii}}
 \item
  Update $b_{n}^\T \msgb{W}{X_{n}}$ for all $n$ as described 
  in Sections~\ref{sec:pathBFFD:LS:point2line} and \ref{sec:pathBFFD:LS:line2point}.
  \end{enumerate}
\end{enumerate}
\smallskip
\end{trivalgorithm}

\end{minipage}%
}
\end{table}

%\afterpage{clearpage}

\begin{table}[tp]
\framebox[\linewidth]{%
\normalsize%
\begin{minipage}{0.95\linewidth}
\begin{trivalgorithm}[Reverse PathBFFD for L1-Regularized Least Squares]%
\label{alg:PathBFFD4LSV2}\vspace{0.5ex}\\
Input: see main text.\\
Output: the knots $\sigma_0^2 = \infty, \sigma_1^2, \ldots, \sigma_{\ell-1}^2, \sigma_\ell^2=0$
and the affine representation of $\hat u_n(\sigma^2)$ between these points.
\\[-1ex]
\hrule
\begin{algorithmic}[1]
\STATE Let $\sigma_0^2=\infty$.
\STATE For all $n$, let $b_{n}^\T \msgb{W}{X_{n}} = \sigma^{-2} b_{n}^\T$ 
       and $\hat u_n = \argmin_{u_n} \kappa_n(u_n)$ (must be unique).
\STATE \label{line:alg:PBFFD4LSV2:initactivesubdomain} 
       For every $n$, set the active subdomain of $\kappa_n(u_n)$ 
       to $\{ \hat u_n \}$.
\STATE\label{line:alg:PBFFD4LSV2:ell0} 
       $\ell \gets 0$.
\REPEAT \label{line:alg:PBFFD4LSV2:repeat} 
   \STATE $\ell \gets \ell+1$.
   \STATE \label{line:alg:PBFFD4LSV2:rerunBFFD} 
          Run Algorithm~\ref{alg:BFFD:DecompInput:SpecUpd} 
          with $\msgf{m}{U_n}$ and $\msgf{V}{U_n}$ (or $\msgf{\xi}{U_n}$ and $\msgf{W}{U_n}$)
          according to the (fixed) active segments of $\kappa_n(u_n)$ (cf.\ Section~\ref{sec:BFFD:CostFunctions}),
          and parameterized by $\sigma^2$ as detailed in Section~\ref{sec:BFFD:Parametric}.
          \\
          We thus obtain new $\hat u_n(\sigma^2)$, $\msgb{m}{U_n}(\sigma^2)$, $\msgb{V}{U_n}(\sigma^2)$ for all $n$.
          \\[0.5ex]
   \STATE \label{line:alg:PBFFD4LSV2:eoas}
          For every $n$, let 
		  \begin{equation} \label{eqn:PBFFD4LSV2:hatun:kappa}
			\varestim{u}_n(\sigma^2)
			= \argmax_{u_n} {\exp}\big(-\kappa_n(u_n)\big) \msgb{\mu}{U_n}(u_n),
		  \end{equation}
		  where $\msgb{\mu}{U_n}(u_n)$ is Gaussian with parameters $\msgb{m}{U_n}(\sigma^2)$ and $\msgb{V}{U_n}(\sigma^2)$,
          and determine 
          \begin{equation}
            \sigma_{\ell,n}^2 = \inf \big\{ \sigma^2 : \text{$\varestim{u}_n(\sigma^2)$ is in the active subdomain of $\kappa_n(u_n)$} \big\},
          \end{equation}
          as detailed in Section~\ref{sec:PathBFFD} (adapted to decreasing $\sigma^2$).
          \\[0.5ex]
   \STATE Let $\sigma_\ell^2 = \max_n\big\{ \sigma_{\ell,n}^2 :  \sigma_{\ell,n}^2 < \sigma_{\ell-1}^2 \big\}$. 
   \IF{ $\sigma_\ell^2 > 0$ }
       \STATE \label{line:alg:PBFFD4LSV2:iftext}
          For the index $n$ such that $\sigma_{\ell,n}^2 = \sigma_\ell^2$,
          change the active segment and subdomain of $\kappa_n(u_n)$ as follows:
          if the present active segment is a line segment,
          then the next active segment is the point segment whose subdomain 
          contains $\varestim{u}_n(\sigma_\ell^2)$;
          if the present active segment is a point segment,
          then the next active segment is the line segment whose subdomain 
          contains $\varestim{u}_n(\sigma_\ell^2-\varepsilon)$
          with positive $\varepsilon \rightarrow 0$.
       \STATE Update $b_{n}^\T \msgb{W}{X_{n}}$ for all $n$ as described 
          in Sections~\ref{sec:pathBFFD:LS:point2line} and \ref{sec:pathBFFD:LS:line2point}.
       \STATE Rearrange and re-index the sequence of variables $U_1,\ldots,U_N$ such that (\ref{eqn:pathLS:sortedcond}) holds.
   \ENDIF \label{line:alg:PBFFD4LSV2:endif}
\UNTIL $\sigma_\ell^2 = 0$.
\end{algorithmic}
\end{trivalgorithm}

\end{minipage}%
}
\end{table}

Specializing Algorithm~\ref{alg:BFFD:SSM} to Fig.~\ref{fig:FactorGraph:DecompInput} 
yields Algorithm~\ref{alg:BFFD:DecompInput}.
(Note that $X_n''$ and $X_n'''$ are not needed in Algorithm~\ref{alg:BFFD:DecompInput} 
since $X_n''=X_n$ and $X_n'''=X_{n-1}$.)
Note that the complexity of the backward recursion 
(Steps \ref{alg:BFFD:DecompInput:SpecUpd:Step:InitBF} and \ref{alg:BFFD:DecompInput:SpecUpd:Step:BF})
is $\calO(N M^2)$
while the complexity of the forward recursion 
(Steps \ref{alg:BFFD:DecompInput:SpecUpd:Step:InitFD} and \ref{alg:BFFD:DecompInput:SpecUpd:Step:FD})
is only $\calO(N M)$.

It turns out that
the complexity of PathBFFD can be reduced to $\calO(N M)$ per iteration.
The key insight is this: 
instead of computing $\msgb{W}{X_{n}}\in\R^{M\times M}$ (for all $n$) from scratch
as in Algorithm~\ref{alg:BFFD:DecompInput}, 
$b_{n}^\T \msgb{W}{X_{n}}\in\R^{1\times M}$ can be updated in each iteration 
with complexity $\calO(NM)$,
as will be explained below.
In consequence, Algorithm~\ref{alg:BFFD:DecompInput} can be simplified 
to Algorithm~\ref{alg:BFFD:DecompInput:SpecUpd}
(which has complexity $\calO(N M)$ per iteration),
and Algorithm~\ref{alg:PathBFFD} (PathBFFD) is changed into Algorithm~\ref{alg:PathBFFD4LS}.

Note that Algorithm~\ref{alg:PathBFFD4LS} still calls Algorithm~\ref{alg:BFFD:DecompInput}
(with complexity $\calO(NM^2)$) during the initialization. 
This can be avoided in the special case where $\argmin_{u_n} \kappa_n(u_n)$ is unique for all $n$,
which applies, in particular, to L1 regularized least squares.
(Note that this condition is \emph{not} implied by the unique path assumption of Section~\ref{sec:PathBFFD:LinDep}.)
In this case, we can reverse the path and start it with $\sigma^2=\infty$
as shown in Algorithm~\ref{alg:PathBFFD4LSV2},
where the first call of Algorithm~\ref{alg:BFFD:DecompInput:SpecUpd}
is with $\msgf{m}{U_n} = \argmin_{u_n} \kappa_n(u_n)$,
$\msgf{V}{U_n}=0$, and $\msgb{W}{X_n}=\sigma^{-2}I$ for all $n$.

It remains to explain the update of $b_{n}^\T \msgb{W}{X_{n}}$ in every iteration.
To this end, 
we will rearrange and re-index the variables $U_1,\ldots,U_N$
in every iteration of PathBFFD 
such that 
the following condition holds:
\begin{equation} \label{eqn:pathLS:sortedcond}
\text{\parbox{0.8\linewidth}{\centering 
For some index $r$, 
the active segments of $\kappa_1(u_1),\ldots,\kappa_r(u_r)$ are lines\\
and the active segments of $\kappa_{r+1}(u_{r+1}),\ldots,\kappa_N(u_N)$ are points.
}}
\end{equation}
The corresponding factor graph is shown in Fig.~\ref{fig:FactorGraph:DecompInput:Sorted}.
Note that keeping track of the changing indices requires maintaining 
two tables of size $\calO(N)$.

In the following, 
$k$ denotes the index $n$
for which the active segment of $\kappa_n(u_n)$ changes
in the $\ell$-th iteration.

%\subsection{Changing from a Point to a Line}
\subsection{Updating $b_{n}^\T \msgb{W}{X_{n}}$ when Changing a Point to a Line}
\label{sec:pathBFFD:LS:point2line}

If the active segment 
of $\kappa_k(u_k)$
changes from a point to a line
(which implies $k>r$), 
the procedure is as follows.
\begin{enumerate}
\item
Consider the factor graph of 
Fig.~\ref{fig:FactorGraph:DecompInput:KnownToUnknown},
which is obtained from Fig.~\ref{fig:FactorGraph:DecompInput:Sorted}
by moving $U_k$ 
to the head. 
\item
Compute $b_{n}^\T \msgb{W}{X_{n}}$ 
in Fig.~\ref{fig:FactorGraph:DecompInput:KnownToUnknown}
for all $n$ according to Proposition~\ref{prop:pathBFFD:point2line:key} below.
\item
Re-index all variables 
in Fig.~\ref{fig:FactorGraph:DecompInput:KnownToUnknown}
back to the notation in Fig.~\ref{fig:FactorGraph:DecompInput:Sorted}.
\end{enumerate}
The complexity of this procedure is $\calO(NM)$.

\begin{proposition} \label{prop:pathBFFD:point2line:key}
Let $\msgb{W}{X_n}$ in 
Figures \ref{fig:FactorGraph:DecompInput:Sorted}
and~\ref{fig:FactorGraph:DecompInput:KnownToUnknown}
be denoted by $\msgb{W}{X_n}^{\text{old}}$ and $\msgb{W}{X_n}^{\text{new}}$,
respectively.
Then
\begin{equation} \label{eqn:pathBFFD:point2line:bnWXn}
b_n^\T \msgb{W}{X_n}^{\text{new}} = b_n^\T \msgb{W}{X_n}^{\text{old}}
\end{equation}
for $n\in \{ 1, \ldots, N \} \setminus \{ k \}$
and
\begin{equation} \label{eqn:pathBFFD:point2line:key}
b_{k}^\T \msgb{W}{X_{k}}^{\text{new}}
= 
\sigma^{-2} b_{k}^\T - \sum_{n=1}^{r}
 b_{k}^\T \big( b_{n}^\T \msgb{W}{X_{n}}^{\text{old}} \big)^\T 
 \big(b_{n}^\T\msgb{W}{X_{n}}^{\text{old}}b_{n}\big)^{-1}b_{n}^\T \msgb{W}{X_{n}}^{\text{old}}.
\vspace{-2.5ex}
\end{equation}
\end{proposition}
Note that 
%the row vector 
(\ref{eqn:pathBFFD:point2line:key})
can be computed with complexity $\calO(NM)$.

\begin{proofof}{of Proposition~\ref{prop:pathBFFD:point2line:key}}
From (\ref{eqn:alg:BFFD:DecompInput:backw:H}) and (\ref{eqn:alg:BFFD:DecompInput:backw:WXppp}), 
we see first, that
\begin{equation} \label{eqn:pathBFFD:point2line:unchangedmsgbW}
\msgb{W}{X_n}^{\text{new}} = \msgb{W}{X_n}^{\text{old}} = \sigma^{-2}I
\end{equation}
for $n\in \{r, \ldots, N \} \setminus \{ k \}$,
and second, that
$\msgb{W}{X_n}^{\text{new}} = \msgb{W}{X_n}^{\text{old}}$
holds also for $n\in\{ 1,\ldots, r-1 \}$,
%for $n\in \{ 1, \ldots, N \} \setminus \{ k \}$,
which proves (\ref{eqn:pathBFFD:point2line:bnWXn}).
From 
(\ref{eqn:alg:BFFD:DecompInput:backw:WXppp}),
we further obtain
\begin{IEEEeqnarray}{r,C,l}
b_{k}^\T \msgb{W}{X_{k}}^{\text{new}} 
& = &  b_{k}^\T \bigg(\msgb{W}{X_{1}}^{\text{new}} - \msgb{W}{X_{1}}^{\text{new}}b_{1}\big(b_{1}^\T\msgb{W}{X_{1}}^{\text{new}}b_{1}\big)^{-1}b_{1}^\T \msgb{W}{X_{1}}^{\text{new}} \bigg)\\
& = &  b_{k}^\T \bigg(\msgb{W}{X_{2}}^{\text{new}} -\msgb{W}{X_{2}}^{\text{new}}b_{2}\big(b_{2}^\T \msgb{W}{X_{2}}^{\text{new}}b_{2}\big)^{-1}b_{2}^\T \msgb{W}{X_{2}}^{\text{new}} \bigg) - b_{k}^\T \msgb{W}{X_{1}}^{\text{new}}b_{1}\big(b_{1}^\T \msgb{W}{X_{1}}^{\text{new}}b_{1}\big)^{-1}b_{1}^\T \msgb{W}{X_{1}}^{\text{new}} 
       \IEEEeqnarraynumspace\\
& = & 	b_{k}^\T\msgb{W}{X_{r}}^{\text{new}} 
        -\sum_{n=1}^{r}b_{k}^\T\msgb{W}{X_{n}}^{\text{new}}b_{n}\big(b_{n}^\T\msgb{W}{X_{n}}^{\text{new}}b_{n}\big)^{-1}b_{n}^\T \msgb{W}{X_{n}}^{\text{new}},
        \IEEEeqnarraynumspace
\end{IEEEeqnarray}
and using (\ref{eqn:pathBFFD:point2line:unchangedmsgbW}) and (\ref{eqn:pathBFFD:point2line:bnWXn})
yields (\ref{eqn:pathBFFD:point2line:key}).
\end{proofof}

\begin{figure*}[tp]
\centering
\begin{tikzpicture}[x=1mm, y=1mm, >=latex]
\tikzset{%
smallbox/.style = {draw, rectangle, inner sep=0mm, minimum width=4mm, minimum height=4mm},
midbox/.style = {draw, rectangle, minimum width=5.5mm, minimum height=5.5mm, inner sep=1mm},
blobbox/.style = {draw, fill=black, rectangle, inner sep=0mm, minimum width=1.75mm, minimum height=1.75mm},
xibox/.style = {draw, pattern=north east lines, inner sep=0mm, minimum width=1.75mm, minimum height=1.75mm},
}
\node[blobbox, label={left:0}] (x0) {};
\draw (x0)+(12.5,0) node[smallbox] (addU1) {$+$};
\draw[->] (x0) -- node[above]{$X_0$} (addU1);
\draw[->] (addU1) -- node[above] {$X_1$} +(11,0);
\draw (addU1)+(0,11) node[midbox] (b1) {$b_1$};
\draw[->] (b1) -- (addU1);
%\draw (b1)+(0,12) node[xibox, label=above:$\msgf{\xi}{U_1}$] (pU1) {};
\draw (b1)+(0,12) node[xibox] (pU1) {};
\draw[->] (pU1) -- node[left] {$U_1$} (b1);
\draw (addU1)+(35,0) node[smallbox] (addUnu) {$+$};
\path (addU1) -- node[pos=0.5] {$\cdots$} (addUnu);
\draw[<-] (addUnu) -- node[above,pos=0.6]{$X_{r-1}$} +(-12.5,0);
\draw (addUnu)+(0,11) node[midbox] (bnu) {$b_r$};
\draw[->] (bnu) -- (addUnu);
\draw (bnu)+(0,12) node[xibox] (pUnu) {};
\draw[->] (pUnu) -- node[left] {$U_r$} (bnu);
\draw (addUnu)+(15,0) node[smallbox] (addUnu1) {$+$};
\draw[->] (addUnu) --node[above]{$X_r$} (addUnu1);
\draw[->] (addUnu1) -- node[above] {$X_{r+1}$} +(12.5,0);
\draw (addUnu1)+(0,11) node[midbox] (bnu1) {$b_{r+1}$};
\draw[->] (bnu1) -- (addUnu1);
\draw (bnu1)+(0,12) node[blobbox] (pUnu1) {};
\draw[->] (pUnu1) -- node[right] {$U_{r+1}$} (bnu1);
\draw (addUnu1)+(35,0) node[smallbox] (addUN) {$+$};
\path (addUnu1) -- node[pos=0.5] {$\cdots$} (addUN);
\draw[<-] (addUN) -- node[above,pos=0.6]{$X_{N-1}$} +(-12.5,0);
\draw (addUN)+(0,11) node[midbox] (bN) {$b_N$};
\draw[->] (bN) -- (addUN);
\draw (bN)+(0,12) node[blobbox] (pUN) {};
\draw[->] (pUN) -- node[right] {$U_N$} (bN);

\draw (addUN)+(15,0) node[smallbox, label={right:$\exp\mleft( - \frac{\| x_N - \breve x_N \|^2}{2\sigma^2} \mright)$}] (pXN) {};
\draw[->] (addUN) -- node[above, pos=0.45]{$X_N$} (pXN);

\end{tikzpicture}
\newcommand{\blobbox}{%
\begin{tikzpicture}[scale=0.1]
\protect\node[draw, fill=black, inner sep=0mm, minimum width=1.75mm, minimum height=1.75mm] {};
\end{tikzpicture}%
}
\newcommand{\xibox}{%
\begin{tikzpicture}[scale=0.1]
\protect\node[draw, pattern=north east lines, inner sep=0mm, minimum width=1.75mm, minimum height=1.75mm] {};
\end{tikzpicture}%
}
%
%\vspace{1ex}
\caption{\label{fig:FactorGraph:DecompInput:Sorted}%
    For Section~\ref{sec:RefineFig2}: 
    Fig.~\ref{fig:FactorGraph:DecompInput} with 
    rearranged and re-indexed variables $U_1,\ldots,U_N$
    and correspondingly re-indexed $b_1,\ldots,b_N$
    such that (\ref{eqn:pathLS:sortedcond}) holds.
    The symbols ``$\,\xibox\,$'' and ``$\,\blobbox\,$''
    denote factors (\ref{eqn:GenGaussian}) 
    with $\msgf{W}{U_n}=0$ and $\msgf{V}{U_n}=0$, respectively,
    cf.\ (\ref{eqn:CostPieces:NUP}).
    }
%\end{figure*}
\vspace{2\floatsep}

%\begin{figure*}[tp]
\centering
\begin{tikzpicture}[x=1mm, y=1mm, >=latex]
\tikzset{%
smallbox/.style = {draw, rectangle, inner sep=0mm, minimum width=4mm, minimum height=4mm},
midbox/.style = {draw, rectangle, minimum width=5.5mm, minimum height=5.5mm, inner sep=1mm},
blobbox/.style = {draw, fill=black, rectangle, inner sep=0mm, minimum width=1.75mm, minimum height=1.75mm},
xibox/.style = {draw, pattern=north east lines, inner sep=0mm, minimum width=1.75mm, minimum height=1.75mm},
}
\node[blobbox, label={left:0}] (x0) {};
\draw (x0)+(12.5,0) node[smallbox] (addUk) {$+$};
\draw[->] (x0) -- node[above]{$X_0$} (addUk);
\draw (addUk)+(0,11) node[midbox] (bk) {$b_k$};
\draw[->] (bk) -- (addUk);
\draw (bk)+(0,12) node[xibox] (pUk) {};
\draw[->] (pUk) -- node[left] {$U_k$} (bk);
\draw (addUk)+(15,0) node[smallbox] (addU1) {$+$};
\draw[->] (addUk) -- node[above] {$X_k$} (addU1);
\draw[->] (addU1) -- node[above] {$X_1$} +(11,0);
\draw (addU1)+(0,11) node[midbox] (b1) {$b_1$};
\draw[->] (b1) -- (addU1);
\draw (b1)+(0,12) node[xibox] (pU1) {};
\draw[->] (pU1) -- node[left] {$U_1$} (b1);
\draw (addU1)+(35,0) node[smallbox] (addUkm1) {$+$};
\path (addU1) -- node[pos=0.5] {$\cdots$} (addUkm1);
\draw[<-] (addUkm1) -- node[above,pos=0.6]{$X_{k-2}$} +(-12.5,0);
\draw (addUkm1)+(0,11) node[midbox] (bkm1) {$b_{k-1}$};
\draw[->] (bkm1) -- (addUkm1);
\draw (bkm1)+(0,12) node[blobbox] (pUkm1) {};
\draw[->] (pUkm1) -- node[left] {$U_{k-1}$} (bkm1);
\draw (addUkm1)+(15,0) node[smallbox] (addUkp1) {$+$};
\draw[->] (addUkm1) --node[above]{$X_{k-1}$} (addUkp1);
\draw[->] (addUkp1) -- node[above] {$X_{k+1}$} +(12.5,0);
\draw (addUkp1)+(0,11) node[midbox] (bkp1) {$b_{k+1}$};
\draw[->] (bkp1) -- (addUkp1);
\draw (bkp1)+(0,12) node[blobbox] (pUkp1) {};
\draw[->] (pUkp1) -- node[right] {$U_{k+1}$} (bkp1);
\draw (addUkp1)+(35,0) node[smallbox] (addUN) {$+$};
\path (addUkp1) -- node[pos=0.5] {$\cdots$} (addUN);
\draw[<-] (addUN) -- node[above,pos=0.6]{$X_{N-1}$} +(-12.5,0);
\draw (addUN)+(0,11) node[midbox] (bN) {$b_N$};
\draw[->] (bN) -- (addUN);
\draw (bN)+(0,12) node[blobbox] (pUN) {};
\draw[->] (pUN) -- node[right] {$U_N$} (bN);

\draw (addUN)+(15,0) node[smallbox, label={right:$\exp\mleft( - \frac{\| x_N - \breve x_N \|^2}{2\sigma^2} \mright)$}] (pXN) {};
\draw[->] (addUN) -- node[above, pos=0.45]{$X_N$} (pXN);

\end{tikzpicture}
%
%\vspace{1ex}
\caption{\label{fig:FactorGraph:DecompInput:KnownToUnknown}%
   Rearranged factor graph when the active segment of $\kappa_k(u_k)$ changes from a point to a line,
   cf.\ Section~\ref{sec:pathBFFD:LS:point2line}.
    }
\end{figure*}

\clearpage

\begin{figure*}[tp]
\centering
\newcommand{\extravspace}{\vspace{2mm}}

\subfloat[Initial factor graph]{%
\label{subfig:LS:line2point:initial}%
\begin{tikzpicture}[x=1mm, y=1mm, >=latex]
\tikzset{%
smallbox/.style = {draw, rectangle, inner sep=0mm, minimum width=4mm, minimum height=4mm},
midbox/.style = {draw, rectangle, minimum width=5.5mm, minimum height=5.5mm, inner sep=0.5mm},
blobbox/.style = {draw, fill=black, rectangle, inner sep=0mm, minimum width=1.75mm, minimum height=1.75mm},
xibox/.style = {draw, pattern=north east lines, inner sep=0mm, minimum width=1.75mm, minimum height=1.75mm},
}
\node[blobbox, label={left:0}] (x0) {};
\draw (x0)+(12.5,0) node[smallbox] (addU1) {};
\draw (x0) -- node[below] {$X_0$} (addU1);
\draw (addU1)+(0,12.5) node[xibox] (pU1) {};
\draw[->] (pU1) -- node[right] {$U_1$} (addU1);
\draw (addU1) -- node[below,pos=0.6] {$X_1$} +(10,0);
\draw (addU1)+(32.5,0) node[smallbox] (addUkm2) {};
\path (addU1) -- node[pos=0.47] {$\cdots$} (addUkm2);
\draw[<-] (addUkm2) -- node[below,pos=0.6] {$X_{k-3}$} +(-12.5,0);
\draw (addUkm2)+(0,12.5) node[xibox] (pUkm2) {};
\draw[->] (pUkm2) -- node[right] {$U_{k-2}$} (addUkm2);
\draw (addUkm2)+(15,0) node[smallbox] (addUkm1) {};
\draw[->] (addUkm2) -- node[below]{$X_{k-2}$} (addUkm1);
\draw (addUkm1)+(0,12.5) node[xibox] (pUkm1) {};
\draw[->] (pUkm1) -- node[right] {$U_{k-1}$} (addUkm1);
\draw (addUkm1)+(15,0) node[smallbox] (addUk) {};
\draw[->] (addUkm1) -- node[below]{$X_{k-1}$} (addUk);
\draw (addUk)+(0,12.5) node[xibox] (pUk) {};
\draw[->] (pUk) -- node[right] {$U_k$} (addUk);
%\draw (addUk) -- node[below,pos=0.6] {$X_k^{(0)}$} +(10,0);
\draw (addUk) -- node[below,pos=0.6] {$X_k^{(0)}\!=\!X_k$} +(15,0);
\draw (addUk)+(37.5,0) node[smallbox] (addUN) {};
\path (addUk) -- node[pos=0.55] {$\cdots$} (addUN);
\draw[<-] (addUN) -- node[below,pos=0.6] {$X_{N-1}$} +(-12.5,0);
\draw (addUN)+(0,12.5) node[blobbox] (pUN) {};
\draw[->] (pUN) -- node[right] {$U_N$} (addUN);
%
%\draw (addUN)+(15,0) node[smallbox] (xN) {};
\draw (addUN)+(15,0) node[smallbox,
       label={right:$\exp\mleft( -\frac{\| x_N - \breve x_N \|^2}{2\sigma^2} \mright)$}] 
      (xN) {};
\draw[->] (addUN) -- node[below] {$X_N$} (xN);
\end{tikzpicture}%
}
\extravspace

\subfloat[Moving $U_k$ left by one position]{%
\label{subfig:LS:line2point:moveleft1}%
\begin{tikzpicture}[x=1mm, y=1mm, >=latex]
\tikzset{%
smallbox/.style = {draw, rectangle, inner sep=0mm, minimum width=4mm, minimum height=4mm},
midbox/.style = {draw, rectangle, minimum width=5.5mm, minimum height=5.5mm, inner sep=0.5mm},
blobbox/.style = {draw, fill=black, rectangle, inner sep=0mm, minimum width=1.75mm, minimum height=1.75mm},
xibox/.style = {draw, pattern=north east lines, inner sep=0mm, minimum width=1.75mm, minimum height=1.75mm},
}
\node[blobbox, label={left:0}] (x0) {};
\draw (x0)+(12.5,0) node[smallbox] (addU1) {};
\draw (x0) -- node[below] {$X_0$} (addU1);
\draw (addU1)+(0,12.5) node[xibox] (pU1) {};
\draw[->] (pU1) -- node[right] {$U_1$} (addU1);
\draw (addU1) -- node[below,pos=0.6] {$X_1$} +(10,0);
\draw (addU1)+(32.5,0) node[smallbox] (addUkm2) {};
\path (addU1) -- node[pos=0.47] {$\cdots$} (addUkm2);
\draw[<-] (addUkm2) -- node[below,pos=0.6] {$X_{k-3}$} +(-12.5,0);
\draw (addUkm2)+(0,12.5) node[xibox] (pUkm2) {};
\draw[->] (pUkm2) -- node[right] {$U_{k-2}$} (addUkm2);
\draw (addUkm2)+(15,0) node[smallbox] (addUkm1) {};
\draw[->] (addUkm2) -- node[below]{$X_{k-2}$} (addUkm1);
\draw (addUkm1)+(0,12.5) node[xibox] (pUkm1) {};
\draw[->] (pUkm1) -- node[right] {$U_k$} (addUkm1);
\draw (addUkm1)+(15,0) node[smallbox] (addUk) {};
\draw[->] (addUkm1) -- node[below]{$X_k^{(1)}$} (addUk);
\draw (addUk)+(0,12.5) node[xibox] (pUk) {};
\draw[->] (pUk) -- node[right] {$U_{k-1}$} (addUk);
\draw (addUk) -- node[below,pos=0.6] {$X_{k-1}$} +(10,0);
\draw (addUk)+(32.5,0) node[smallbox] (addUN) {};
\path (addUk) -- node[pos=0.47] {$\cdots$} (addUN);
\draw[<-] (addUN) -- node[below,pos=0.6] {$X_{N-1}$} +(-12.5,0);
\draw (addUN)+(0,12.5) node[blobbox] (pUN) {};
\draw[->] (pUN) -- node[right] {$U_N$} (addUN);
\draw (addUN)+(15,0) node[smallbox] (xN) {};
\draw[->] (addUN) -- node[below] {$X_N$} (xN);
\end{tikzpicture}%
}
\extravspace

\subfloat[Moving $U_k$ left by two positions.]{%
\label{subfig:LS:line2point:moveleft2}%
\begin{tikzpicture}[x=1mm, y=1mm, >=latex]
\tikzset{%
smallbox/.style = {draw, rectangle, inner sep=0mm, minimum width=4mm, minimum height=4mm},
midbox/.style = {draw, rectangle, minimum width=5.5mm, minimum height=5.5mm, inner sep=0.5mm},
blobbox/.style = {draw, fill=black, rectangle, inner sep=0mm, minimum width=1.75mm, minimum height=1.75mm},
xibox/.style = {draw, pattern=north east lines, inner sep=0mm, minimum width=1.75mm, minimum height=1.75mm},
}
\node[blobbox, label={left:0}] (x0) {};
\draw (x0)+(12.5,0) node[smallbox] (addU1) {};
\draw (x0) -- node[below] {$X_0$} (addU1);
\draw (addU1)+(0,12.5) node[xibox] (pU1) {};
\draw[->] (pU1) -- node[right] {$U_1$} (addU1);
\draw (addU1) -- node[below,pos=0.6] {$X_1$} +(10,0);
\draw (addU1)+(32.5,0) node[smallbox] (addUkm2) {};
\path (addU1) -- node[pos=0.47] {$\cdots$} (addUkm2);
\draw[<-] (addUkm2) -- node[below,pos=0.6] {$X_{k-3}$} +(-12.5,0);
\draw (addUkm2)+(0,12.5) node[xibox] (pUkm2) {};
\draw[->] (pUkm2) -- node[right] {$U_k$} (addUkm2);
\draw (addUkm2)+(15,0) node[smallbox] (addUkm1) {};
\draw[->] (addUkm2) -- node[below]{$X_k^{(2)}$} (addUkm1);
\draw (addUkm1)+(0,12.5) node[xibox] (pUkm1) {};
\draw[->] (pUkm1) -- node[right] {$U_{k-2}$} (addUkm1);
\draw (addUkm1)+(15,0) node[smallbox] (addUk) {};
\draw[->] (addUkm1) -- node[below]{$X_{k-2}$} (addUk);
\draw (addUk)+(0,12.5) node[xibox] (pUk) {};
\draw[->] (pUk) -- node[right] {$U_{k-1}$} (addUk);
\draw (addUk) -- node[below,pos=0.6] {$X_{k-1}$} +(10,0);
\draw (addUk)+(32.5,0) node[smallbox] (addUN) {};
\path (addUk) -- node[pos=0.47] {$\cdots$} (addUN);
\draw[<-] (addUN) -- node[below,pos=0.6] {$X_{N-1}$} +(-12.5,0);
\draw (addUN)+(0,12.5) node[blobbox] (pUN) {};
\draw[->] (pUN) -- node[right] {$U_N$} (addUN);
\draw (addUN)+(15,0) node[smallbox] (xN) {};
\draw[->] (addUN) -- node[below] {$X_N$} (xN);
\end{tikzpicture}%
}
\extravspace

\subfloat[Moving $U_k$ (by $k-1$ positions) to the head.]{%
\label{subfig:LS:line2point:athead}%
\begin{tikzpicture}[x=1mm, y=1mm, >=latex]
\tikzset{%
smallbox/.style = {draw, rectangle, inner sep=0mm, minimum width=4mm, minimum height=4mm},
midbox/.style = {draw, rectangle, minimum width=5.5mm, minimum height=5.5mm, inner sep=0.5mm},
blobbox/.style = {draw, fill=black, rectangle, inner sep=0mm, minimum width=1.75mm, minimum height=1.75mm},
xibox/.style = {draw, pattern=north east lines, inner sep=0mm, minimum width=1.75mm, minimum height=1.75mm},
}
\node[blobbox, label={left:0}] (x0) {};
\draw (x0)+(12.5,0) node[smallbox] (addUk) {};
\draw (x0) -- node[below] {$X_0$} (addUk);
\draw (addUk)+(0,12.5) node[xibox] (pUk) {};
\draw[->] (pUk) -- node[right] {$U_k$} (addUk);
\draw (addUk)+(17.5,0) node[smallbox] (addU1) {};
\draw[->] (addUk) -- node[below]{$X_k^{(k-1)}$} (addU1);
\draw (addU1)+(0,12.5) node[xibox] (pU1) {};
\draw[->] (pU1) -- node[right] {$U_1$} (addU1);
\draw (addU1) -- node[below,pos=0.6] {$X_1$} +(10,0);
\draw (addU1)+(35,0) node[smallbox] (addUN) {};
\path (addU1) -- node[pos=0.45] {$\cdots$} (addUN);
\draw[<-] (addUN) -- node[below,pos=0.6] {$X_{N-1}$} +(-12.5,0);
\draw (addUN)+(0,12.5) node[blobbox] (pUN) {};
\draw[->] (pUN) -- node[right] {$U_N$} (addUN);
\draw (addUN)+(15,0) node[smallbox] (xN) {};
\draw[->] (addUN) -- node[below] {$X_N$} (xN);
\end{tikzpicture}%
}
\extravspace

\subfloat[Moving $U_k$ to the tail and changing the active segment of $\kappa_k(u_k)$.]{%
\label{subfig:LS:line2point:finalattail}%
\begin{tikzpicture}[x=1mm, y=1mm, >=latex]
\tikzset{%
smallbox/.style = {draw, rectangle, inner sep=0mm, minimum width=4mm, minimum height=4mm},
midbox/.style = {draw, rectangle, minimum width=5.5mm, minimum height=5.5mm, inner sep=0.5mm},
blobbox/.style = {draw, fill=black, rectangle, inner sep=0mm, minimum width=1.75mm, minimum height=1.75mm},
xibox/.style = {draw, pattern=north east lines, inner sep=0mm, minimum width=1.75mm, minimum height=1.75mm},
}
\node[blobbox, label={left:0}] (x0) {};
\draw (x0)+(12.5,0) node[smallbox] (addU1) {};
\draw (x0) -- node[below] {$X_0$} (addU1);
\draw (addU1)+(0,12.5) node[xibox] (pU1) {};
\draw[->] (pU1) -- node[right] {$U_1$} (addU1);
\draw (addU1) -- node[below,pos=0.6] {$X_1$} +(10,0);
\draw (addU1)+(32.5,0) node[smallbox] (addUN) {};
\path (addU1) -- node[pos=0.47] {$\cdots$} (addUN);
\draw[<-] (addUN) -- node[below,pos=0.6] {$X_{N-1}$} +(-12.5,0);
\draw (addUN)+(0,12.5) node[blobbox] (pUN) {};
\draw[->] (pUN) -- node[right] {$U_N$} (addUN);

\draw (addUN)+(15,0) node[smallbox] (addUk) {};
\draw[->] (addUN) -- node[below]{$X_N$} (addUk);
\draw (addUk)+(0,12.5) node[blobbox] (pUk) {};
\draw[->] (pUk) -- node[right] {$U_k$} (addUk);
\draw (addUk)+(15,0) node[smallbox,
       label={right:$\exp\mleft( -\frac{\| x_N - \breve x_N \|^2}{2\sigma^2} \mright)$}] 
      (xk) {};
\draw[->] (addUk) -- node[below] {$X_k$} (xk);
\end{tikzpicture}%
}

\caption{\label{fig:LS:line2point}%
For Section~\ref{sec:pathBFFD:LS:line2point}: 
step-by-step rearrangement when the active segment of $\kappa_k(u_k)$
changes from a line to a point,
using the notation of Fig.~\ref{fig:RegLS:SimplifiedGF}.
}
\end{figure*}

\begin{figure}
\centering
\begin{tikzpicture}[x=1mm, y=1mm, >=latex]
\tikzset{%
smallbox/.style = {draw, rectangle, inner sep=0mm, minimum width=4mm, minimum height=4mm},
midbox/.style = {draw, rectangle, minimum width=5.5mm, minimum height=5.5mm, inner sep=0.5mm},
%blobbox/.style = {draw, fill=black, rectangle, inner sep=0mm, minimum width=1.75mm, minimum height=1.75mm},
%xibox/.style = {draw, pattern=north east lines, inner sep=0mm, minimum width=1.75mm, minimum height=1.75mm},
tinybox/.style = {draw, rectangle, inner sep=0mm, minimum width=1.75mm, minimum height=1.75mm},
}
\node[smallbox] (add) {$+$};
%\draw[<-] (add) -- +(-12.5,0); 
%\draw[->] (add) -- +(12.5,0);
\draw[<-] (add) -- +(-15,0); 
\draw[->] (add) -- node[below,pos=0.6] {$X_n$} +(17.5,0);
\draw (add)+(0,10) node[midbox] (b) {$b_n$};
\draw[->] (b) -- (add);
\draw (b)+(0,15) node[tinybox] (pU) {};
\draw[->] (pU) -- node[right, pos=0.35] {$U_n$} (b);
\draw (0,-5) node[draw, rectangle, anchor=south, dashed, minimum width=12.5mm, minimum height=21mm] {};

\draw (add)+(40,0) node[smallbox] (simplebox) {};
\draw[<-] (simplebox) -- +(-12.5,0);
%\draw[->] (simplebox) -- +(12.5,0);
\draw[->] (simplebox) -- node[below] {$X_n$} +(12.5,0);
\draw (simplebox)+(0,12.5) node[tinybox] (pU2) {};
\draw[->] (pU2) -- node[right] {$U_n$} (simplebox);

\draw (add)+(20,7.5) node {$=$};
\end{tikzpicture}
\vspace{1ex}
\caption{\label{fig:RegLS:SimplifiedGF}%
The notation used in Fig.~\ref{fig:LS:line2point}
results from closing the dashed box.
The outgoing edge $X_n$ is kept together with $U_n$ under rearrangements.}
\end{figure}

\clearpage

\subsection{Updating $b_{n}^\T \msgb{W}{X_{n}}$ when Changing a Line to a Point}
\label{sec:pathBFFD:LS:line2point}

If the active segment 
of $\kappa_k(u_k)$
changes from a line to a point 
(which implies $k \leq r$), 
the procedure is as follows.

\begin{enumerate}
\item
Consider the factor graph of Fig.~\ref{subfig:LS:line2point:finalattail},
which is obtained from Fig.~\ref{subfig:LS:line2point:initial}
by moving $U_k$ to the tail of the factor graph.
(The intermediate Figures \ref{subfig:LS:line2point:moveleft1}--\ref{subfig:LS:line2point:athead}
are used in the proof of Proposition~\ref{prop:pathBFFD:line2point:key} below.)
\item
Compute $b_{n}^\T \msgb{W}{X_{n}}$ in Fig.~\ref{subfig:LS:line2point:finalattail}
for all $n$ according to Proposition~\ref{prop:pathBFFD:line2point:key}.
\item
Re-index all variables in Fig.~\ref{subfig:LS:line2point:finalattail}
back to the notation in Fig.~\ref{fig:FactorGraph:DecompInput:Sorted}.
\end{enumerate}
The total complexity of this procedure is $\calO(NM)$.

\begin{proposition} \label{prop:pathBFFD:line2point:key}
Let $\msgb{W}{X_n}$ in Figures \ref{subfig:LS:line2point:initial} and \ref{subfig:LS:line2point:finalattail}
be denoted $\msgb{W}{X_n}^{\text{old}}$ and $\msgb{W}{X_n}^{\text{new}}$, respectively.
Then
\begin{equation} \label{eqn:pathBFFD:LS:line2point:nothing2do}
b_n^\T \msgb{W}{X_n}^{\text{new}} = b_n^\T \msgb{W}{X_n}^{\text{old}}
\text{~~~for $n>k$,}
\end{equation}
\begin{equation} \label{eqn:pathBFFD:LS:line2point:obvious}
b_k^\T \msgb{W}{X_k}^{\text{new}} = \sigma^{-2} b_k^\T,
\end{equation}
and for $n<k$, we have the coupled recursions
\begin{IEEEeqnarray}{r,C,l}
b_{k-\nu}^\T \msgb{W}{X_{k-\nu}}^{\text{new}}
& = &  b_{k-\nu}^\T \msgb{W}{X_{k-\nu}}^{\text{old}} 
       + b_{k-\nu}^\T \Big( b_{k}^\T \msgb{W}{X_k^{(\nu-1)}} \Big)^\T
       \Big (b_{k}^\T \msgb{W}{X_k^{(\nu-1)}}  b_{k} \Big)^{-1}
       b_{k}^\T \msgb{W}{X_k^{(\nu-1)}}
       \label{eqn:pathBFFD:LS:line2point:WX:recursion}
\\
b_k^\T \msgb{W}{X_k^{(\nu)}}
& = &  b_k^\T \msgb{W}{X_k^{(\nu-1)}} - b_{k}^\T \Big( b_{k-\nu}^\T \msgb{W}{X_{k-\nu}}^{\text{new}} \Big)^\T
       \Big( b_{k-\nu}^\T \msgb{W}{X_{k-\nu}}^{\text{new}} b_{k-\nu} \Big)^{-1}
       b_{k-\nu}^\T \msgb{W}{X_{k-\nu}}^{\text{new}}, 
       \label{eqn:pathBFFD:LS:line2point:WXtemp:recursion}
\end{IEEEeqnarray}
for $\nu=1,\ldots,k-1$,
beginning with $b_k^\T \msgb{W}{X_k^{(0)}} = b_k^\T \msgb{W}{X_k}^{\text{old}}$.
\end{proposition}
(The matrices $\msgb{W}{X_k^{(\nu)}}$ refer to the edges $X_k^{(\nu)}$ 
in Figures \ref{subfig:LS:line2point:moveleft1}--\ref{subfig:LS:line2point:athead},
cf.\ the proof below.)
Note that, for fixed $\nu$, 
both (\ref{eqn:pathBFFD:LS:line2point:WX:recursion}) and (\ref{eqn:pathBFFD:LS:line2point:WXtemp:recursion})
can be computed with complexity $\calO(M)$.

\begin{proofof}{of Proposition~\ref{prop:pathBFFD:line2point:key}}
Eq.~(\ref{eqn:pathBFFD:LS:line2point:obvious}) is obvious from $\msgb{W}{X_k}^{\text{new}} = \sigma^{-2}I$.
From (\ref{eqn:alg:BFFD:DecompInput:backw:H}) and (\ref{eqn:alg:BFFD:DecompInput:backw:WXppp}),
we then see that
\begin{equation}
\msgb{W}{X_n}^{\text{new}} = \msgb{W}{X_n}^{\text{old}} = \sigma^{-2}I
\end{equation}
for $n \geq r$, 
and $\msgb{W}{X_n}^{\text{new}} = \msgb{W}{X_n}^{\text{old}}$ for $k<n<r$, 
which implies (\ref{eqn:pathBFFD:LS:line2point:nothing2do}).

The recursions (\ref{eqn:pathBFFD:LS:line2point:WX:recursion}) and (\ref{eqn:pathBFFD:LS:line2point:WXtemp:recursion})
are obtained by moving $U_k$ in $k-1$ steps to the head of the factor graph,
by one position per step, as shown in Figures \ref{subfig:LS:line2point:moveleft1}--\ref{subfig:LS:line2point:athead}.
Note also the variables $X_k^{(0)} = X_k$ in Fig.~\ref{subfig:LS:line2point:initial},
$X_k^{(1)}$ in Fig.~\ref{subfig:LS:line2point:moveleft1}, etc.,
and note further that 
(\ref{eqn:alg:BFFD:DecompInput:backw:H}) and (\ref{eqn:alg:BFFD:DecompInput:backw:WXppp})
imply 
$\msgb{W}{X_{k-1}}^{\text{new}} = \msgb{W}{X_k^{(0)}}$,
$\msgb{W}{X_{k-2}}^{\text{new}} = \msgb{W}{X_k^{(1)}}$,
etc.
In the first step
(from Fig.~\ref{subfig:LS:line2point:initial} to Fig.~\ref{subfig:LS:line2point:moveleft1}), 
we have both 
\begin{IEEEeqnarray}{r,C,l}
b_{k-1}^\T \msgb{W}{X_{k-1}}^{\text{new}} 
& = &  b_{k-1}^\T \msgb{W}{X_k^{(0)}}\\
& = &  b_{k-1}^\T \bigg( \msgb{W}{X_{k-1}}^{\text{old}}  
       + \msgb{W}{X_k^{(0)}} b_k
         \Big( b_k^\T \msgb{W}{X_k^{(0)}}  b_k \Big)^{-1}
         b_{k}^\T \msgb{W}{X_k^{(0)}} \bigg)
         \label{eqn:pathBFFD:LS:line2point:WX:MIL}\\
& = &  b_{k-1}^\T \msgb{W}{X_{k-1}}^{\text{old}}  
       + b_{k-1}^\T \Big( b_k^\T \msgb{W}{X_k^{(0)}} \Big)^\T
         \Big( b_k^\T \msgb{W}{X_k^{(0)}}  b_k \Big)^{-1}
         b_{k}^\T \msgb{W}{X_k^{(0)}}
         \label{eqn:pathBFFD:LS:line2point:WX:final}
\end{IEEEeqnarray}
and
\begin{IEEEeqnarray}{r,C,l}
b_{k}^\T \msgb{W}{X_{k}^{(1)}}
& = &  b_{k}^\T \bigg( \msgb{W}{X_{k-1}}^{\text{new}} - \msgb{W}{X_{k-1}}^{\text{new}} b_{k-1} 
       \Big (b_{k-1}^\T \msgb{W}{X_{k-1}}^{\text{new}}  b_{k-1} \Big)^{-1}
       b_{k-1}^\T \msgb{W}{X_{k-1}}^{\text{new}} \bigg)
       \label{eqn:pathBFFD:LS:line2point:WXtemp:MIL}\\
& = &  b_k^\T \msgb{W}{X_k^{(0)}} - b_{k}^\T \Big( b_{k-1}^\T \msgb{W}{X_{k-1}}^{\text{new}} \Big)^\T
       \Big( b_{k-1}^\T \msgb{W}{X_{k-1}}^{\text{new}} b_{k-1} \Big)^{-1}
       b_{k-1}^\T \msgb{W}{X_{k-1}}^{\text{new}}, 
       \label{eqn:pathBFFD:LS:line2point:WXtemp:final}
%\IEEEeqnarraynumspace
\end{IEEEeqnarray}
where both (\ref{eqn:pathBFFD:LS:line2point:WX:MIL}) and (\ref{eqn:pathBFFD:LS:line2point:WXtemp:MIL}) 
follow from 
%(\ref{eqn:CompositeAdditionBackwardPrecision}).
(\ref{eqn:alg:BFFD:DecompInput:backw:WXppp}).
Note that both (\ref{eqn:pathBFFD:LS:line2point:WX:final}) and (\ref{eqn:pathBFFD:LS:line2point:WXtemp:final})
can be computed with complexity $\calO(M)$.
Note also, for $n<k-1$,
$b_n^\T \msgb{W}{X_n} = b_n^\T \msgb{W}{X_n}^\text{old}$
is not affected by this step.

Likewise, in the second step 
(from Fig.~\ref{subfig:LS:line2point:moveleft1} to Fig.~\ref{subfig:LS:line2point:moveleft2}), 
we have both
\begin{IEEEeqnarray}{r,C,l}
b_{k-2}^\T \msgb{W}{X_{k-2}}^{\text{new}} 
& = &  b_{k-2}^\T \msgb{W}{X_k^{(1)}}\\
& = &  b_{k-2}^\T \msgb{W}{X_{k-2}}^{\text{old}} 
       + b_{k-2}^\T \Big( b_{k}^\T \msgb{W}{X_k^{(1)}} \Big)^\T
       \Big (b_{k}^\T \msgb{W}{X_k^{(1)}}  b_{k} \Big)^{-1}
       b_{k}^\T \msgb{W}{X_k^{(1)}}
\end{IEEEeqnarray}
and
\begin{IEEEeqnarray}{r,C,l}
b_{k}^\T \msgb{W}{X_{k}^{(2)}}
& = &  b_k^\T \msgb{W}{X_k^{(1)}} - b_{k}^\T \Big( b_{k-2}^\T \msgb{W}{X_{k-2}}^{\text{new}} \Big)^\T
       \Big( b_{k-2}^\T \msgb{W}{X_{k-2}}^{\text{new}} b_{k-2} \Big)^{-1}
       b_{k-2}^\T \msgb{W}{X_{k-2}}^{\text{new}}, 
\end{IEEEeqnarray}
and so on.
\end{proofof}

\clearpage

\begin{table}[tp]
	\framebox[\linewidth]{%
		\normalsize%
		\begin{minipage}{0.95\linewidth}
			\begin{trivalgorithm}[FFBDD in Fig.~\ref{fig:FactorGraph:DecompOutput}]%
				\label{alg:FFBDD:DecompOutput}\vspace{0.5ex}\\
				Input: see Section \ref{sec:StatModelFG}\\
				Output: $\hat x_0$ and $\hat y_n$, and (in this paper) $\msgf{m}{Y_n}$ and $\msgf{V}{Y_n}$ for $n=1,\ldots,N$
				\\[-1ex]
				\hrule
				\vspace{-2.5ex}
				\begin{multicols}{2}
					%\emph{Backward filtering:}
					\begin{enumerate}[1.]
						\item \label{alg:FFBDD:DecompOutput:Step:InitFF}
						Let $\msgf{V}{X_{0}}=\sigma^{2}I$ and $\msgf{m}{X_{0}}=0$.
						\item \label{alg:FFBDD:DecompOutput:Step:FF}
						\pkw{for} $n=1$ \pkw{to} $N$, compute the following:
						\begin{IEEEeqnarray}{rCl}
							G_{n} & = & \begin{cases}
								0,  & \text{if $\msgb{W}{Y_n}=0$}\\ 
								\big(c_{n}^{\mathsf{T}}\msgf{V}{X_{n-1}}c_{n}\big)^{-1},  & \text{if $\msgb{V}{Y_n}=0$}\\
								\text{by (\ref{eqn:FFBDD:Gn:General})},   & \text{else} 
								%\text{by (\ref{eqn:FFBDD:DecompOutput:Gn:General})},   & \text{else.} 
								%\big(\msgf{W}{U_{n}} + b_{n}^{\mathsf{T}}\msgb{W}{X_{n}^{''}}b_{n}\big)^{-1}
							\end{cases}
							\label{eqn:alg:FFBDD:DecompOutput:forw:G}\\
							g_{n} & = & \begin{cases}
								c_{n}\msgb{\xi}{Y_{n}}, &  \msgb{W}{Y_{n}}=0, \IEEEeqnarraynumspace\\
								% b_{n}    \big(b_{n}^{\mathsf{T}}\msgb{W}{X_{n}^{''}}b_{n}\big)^{-1}\big(\msgf{\xi}{U_{n}}+b_{n}^{\mathsf{T}}\msgb{\xi}{X_{n}^{''}}\big), & \msgf{W}{U_{n}}=0,\\
								c_{n}G_{n}\big(\msgb{m}{Y_{n}}-c_{n}^{\mathsf{T}}\msgf{m}{X_{n-1}}\big), & \text{else}
							\end{cases}
							\label{eqn:alg:FFBDD:DecompOutput:forw:g}
						\end{IEEEeqnarray}
						\begin{IEEEeqnarray}{rCl}
							\msgf{m}{X_{n}} & = &\msgf{m}{X_{n-1}} + \msgf{V}{X_{n-1}}g_{n}
							\label{eqn:alg:FFBDD:DecompOutput:forw:mX}\\
							\msgf{V}{X_{n}} & = & \msgf{V}{X_{n-1}} - \msgf{V}{X_{n-1}} c_{n}G_{n}c_{n}^{\mathsf{T}}\msgf{V}{X_{n-1}}
							\label{eqn:alg:FFBDD:DecompOutput:forw:VX}
						\end{IEEEeqnarray}
						\hspace{1.5em}Store only $c_{n}^{\mathsf{T}}\msgf{V}{X_{n-1}}, c_{n}^{\mathsf{T}}\msgf{m}{X_{n-1}}$ and $c_{n}^{\mathsf{T}}\msgf{V}{X_{n-1}}c_{n}$.%
						
						\item[] 
						\pkw{end for}
						\newpage

                        %\rule{0em}{4ex}  % to end the column here
						
						\item
						Let $\tilde{\xi}_{X_{N}}=0$. 
						\item
						\pkw{for} $n=N$ \pkw{to} $1$, compute the following:
						\begin{IEEEeqnarray}{rCl}	
							\msgf{V}{Y_{n}} &= &c_{n}^{\mathsf{T}}\msgf{V}{X_{n-1}}c_{n}
							\label{eqn:alg:FFBDD:DecompOutput:backw:msgfVY}\\
							\msgf{m}{Y_{n}} &=  & c_{n}^{\mathsf{T}}\msgf{m}{X_{n-1}}-c_{n}^{\mathsf{T}}\msgf{V}{X_{n-1}}	\tilde{\xi}_{X_{n}}
							\label{eqn:alg:FFBDD:DecompOutput:fwd:msgfmY}
						\end{IEEEeqnarray}
						\begin{equation} \label{eqn:alg:FFBDD:DecompOutput:backw:hatDualY}
							\tilde{\xi}_{Y_{n}} = \begin{cases}
								-\msgb{\xi}{Y_n}, & \text{if $\msgb{W}{Y_n}=0$} \\
								(\msgf{V}{Y_{n}} )^{-1} \mleft(\msgf{m}{Y_n}  - \msgb{m}{Y_n}\mright),  & \text{if $\msgb{V}{Y_n}=0$} \\
								\text{by (\ref{eqn:FFBDD:decideDualYn:General}),} 
								%\text{by (\ref{eqn:FFBDD:DecompOutput:decideDualYn:General}),} 
								& \text{else}
							\end{cases}
						\end{equation}
						\begin{IEEEeqnarray}{rCl}	
							\tilde{\xi}_{X_{n-1}}&= &\tilde{\xi}_{X_{n}}+ c_{n}\tilde{\xi}_{Y_{n}}
							\label{eqn:alg:FFBDD:DecompOutput:backw:hatDualX} 
						\end{IEEEeqnarray}
						\begin{IEEEeqnarray}{rCl}
							\hat{y}_{n}=\begin{cases}
								\msgb{m}{Y_{n}}, & \text{if $\msgb{V}{Y_n}=0$} \\
								\msgf{m}{Y_{n}} - \msgf{V}{Y_{n}}	\tilde{\xi}_{Y_{n}} ,  & \text{else} 
							\end{cases}
						\end{IEEEeqnarray}
						\item[]
						\pkw{end for}
						
						\item
						Compute
						\begin{equation} \label{eqn:alg:FFBDD:DecompOutput:hatx0}
						\hat x_0 = \msgf{m}{X_0} - \msgf{V}{X_0} \tilde\xi_{X_0}.
						\end{equation}
						
					\end{enumerate}
					%\rule{0em}{8ex}  % visual adjust
				\end{multicols}
				\vspace{-1ex}% 
				\hrule
				\vspace{1ex}
				
				Concerning (\ref{eqn:alg:FFBDD:DecompOutput:forw:G}) and 
				(\ref{eqn:alg:FFBDD:DecompOutput:backw:hatDualY}),
				the ``else'' case is not used in this paper.
			\end{trivalgorithm}
			
		\end{minipage}%
	}
\end{table}
%\vspace{\dblfloatsep}
%\vspace{\dblfloatsep}

\begin{table}[tp]
	\framebox[\linewidth]{%
		\normalsize%
		\begin{minipage}{0.95\linewidth}
			\begin{trivalgorithm}[Simplified FFBDD Update for PathFFBDD in Fig.~\ref{fig:FactorGraph:DecompOutput}]%
				\label{alg:BFFD:DecompOutput:SpecUpd}\vspace{0.5ex}\\
				Input: see Sections \ref{sec:StatModelFG} and~\ref{sec:RefineFig3},
				       and precomputed $c_n^\T \msgf{V}{X_{n-1}} = \big( \msgf{V}{X_{n-1}} c_n \big)^\T$ for $n=1,\ldots,N$\\
				Output: $\hat x_0$ and $\hat y_n$, $\msgf{m}{Y_n}$, $\msgf{V}{Y_n}$ for $n=1,\ldots,N$
				\\[-1ex]
				\hrule
				\vspace{-2ex}
				\begin{multicols}{2}
				  \begin{enumerate}[1.]
				  \item \label{alg:FFBDD:DecompOutput:SpecUpd:Step:InitFF}
				  Let $\msgf{V}{X_{0}}=\sigma^{2}I$ and $\msgf{m}{X_{0}}=0$.
				  \item \label{alg:FFBDD:DecompOutput:SpecUpd:Step:FF}
				  \pkw{for} $n=1$ \pkw{to} $N$, compute the following:
				    \begin{IEEEeqnarray}{rCl}
				      \alpha_n & = &  c_n^\T \msgf{V}{X_{n-1}} c_n
				      \\
				      \beta_n & = & \begin{cases}
								\msgb{\xi}{Y_{n}}, &  \text{if $\msgb{W}{Y_{n}}=0$}, \IEEEeqnarraynumspace\\
								\alpha_{n}^{-1} \big(\msgb{m}{Y_{n}}-c_{n}^{\mathsf{T}}\msgf{m}{X_{n-1}}\big), & \text{if $\msgb{V}{Y_{n}}=0$}
							\end{cases}
							\label{eqn:alg:FFBDD:DecompOutput:forw:h}
				    \end{IEEEeqnarray}
					\begin{IEEEeqnarray}{rCl}
					  \msgf{m}{X_n} & = & \msgf{m}{X_{n-1}} + \beta_n \msgf{V}{X_{n-1}} c_n
					  \label{eqn:alg:FFBDD:DecompOutput:SpecUpd:msgfm}
				    \end{IEEEeqnarray}
				    \hspace{1.0em}Store $c_n^\T \msgf{m}{X_{n-1}}$ and $\alpha_n$.
				  \item[] \vspace{0.5ex} % visual
				  \pkw{end for}
                  \newpage
						
						\item
						Let $\tilde{\xi}_{X_{N}}=0$. 
						\item
						\pkw{for} $n=N$ \pkw{to} $1$, compute the following:
						\begin{IEEEeqnarray}{rCl}	
							\msgf{V}{Y_{n}} &= &  \alpha_n
							\label{eqn:alg:FFBDD:DecompOutput:backw:SpecUpd:msgfVY}\\
							\msgf{m}{Y_{n}} &=  & c_{n}^{\mathsf{T}}\msgf{m}{X_{n-1}}-c_{n}^{\mathsf{T}}\msgf{V}{X_{n-1}}	\tilde{\xi}_{X_{n}}
							\label{eqn:alg:FFBDD:DecompOutput:fwd:SpecUpd:msgfmY}
						\end{IEEEeqnarray}
						\begin{equation} \label{eqn:alg:FFBDD:DecompOutput:backw:SpecUpd:hatDualY}
							\tilde{\xi}_{Y_{n}} = \begin{cases}
								-\msgb{\xi}{Y_n}, & \text{if $\msgb{W}{Y_n}=0$} \\
								(\msgf{V}{Y_{n}} )^{-1} \mleft(\msgf{m}{Y_n}  - \msgb{m}{Y_n}\mright),  & \text{if $\msgb{V}{Y_n}=0$}
								%\text{by (\ref{eqn:FFBDD:decideDualYn:General}),} 
								%& \text{else}
							\end{cases}
						\end{equation}
						\begin{IEEEeqnarray}{rCl}	
							\tilde{\xi}_{X_{n-1}}&= &\tilde{\xi}_{X_{n}}+ c_{n}\tilde{\xi}_{Y_{n}}
							\label{eqn:alg:FFBDD:DecompOutput:backw:SpecUpd:hatDualX} 
						\end{IEEEeqnarray}
						\begin{IEEEeqnarray}{rCl}
							\hat{y}_{n}=\begin{cases}
								\msgb{m}{Y_{n}}, & \text{if $\msgb{V}{Y_n}=0$} \\
								\msgf{m}{Y_{n}} - \msgf{V}{Y_{n}}	\tilde{\xi}_{Y_{n}} ,  & \text{else} 
							\end{cases}
						\end{IEEEeqnarray}
						\item[]
						\pkw{end for}
						
						\item
						Compute
						\begin{equation} \label{eqn:alg:FFBDD:DecompOutput:SpecUpd:hatx0}
						\hat x_0 = \msgf{m}{X_0} - \msgf{V}{X_0} \tilde\xi_{X_0}. 
						\end{equation}
				  %\rule{0em}{30ex}
				  \end{enumerate}
				\end{multicols}
				\vspace{-2ex}   % visual adjust
			\end{trivalgorithm}
			
		\end{minipage}%
	}
\end{table}

\begin{table}[tp]
\framebox[\linewidth]{%
\normalsize%
\begin{minipage}{0.95\linewidth}
\begin{trivalgorithm}[PathFFBDD for Fig.~\ref{fig:FactorGraph:DecompOutput}]%
\label{alg:PathFFBDD4SV}\vspace{0.5ex}\\
Input: see main text.\\
Output: the knots $\sigma_0^2 = 0, \sigma_1^2, \ldots, \sigma_{\ell-1}^2, \sigma_\ell^2=\infty$
and the affine representation of $\hat x_0(\sigma^2)$ between these points.
\\[-1ex]
\hrule
\begin{algorithmic}[1]
\STATE Let $\sigma_0^2=0$.
\STATE For all $n$, let $c_{n}^\T \msgb{V}{X_{n-1}} = \sigma^2 c_{n}^\T$. 
       %and $\hat y_n = 0$.
\STATE %\label{line:alg:PBFFD4LSV2:initactivesubdomain} 
       For every $n$, set the active subdomain of $\kappa_n(y_n)$ 
       to the subdomain that contains $\hat y_n=0$.
       %to $\{ \hat u_n \}$.
%
\STATE%\label{line:alg:PBFFD4LSV2:ell0} 
       $\ell \gets 0$.
\REPEAT %\label{line:alg:PBFFD4LSV2:repeat} 
   \STATE $\ell \gets \ell+1$.
   \STATE %\label{line:alg:PBFFD4LSV2:rerunBFFD} 
          Run Algorithm~\ref{alg:BFFD:DecompOutput:SpecUpd} 
          with $\msgb{\xi}{Y_n}$ and $\msgb{W}{Y_n}$ (or $\msgb{m}{Y_n}$ and $\msgb{V}{Y_n}$)
          according to the (fixed) active segments of $\kappa_n(y_n)$,
          and with $\msgf{m}{X_n}$, $c_n^\T \msgf{V}{X_{n-1}}$,
          $\msgf{m}{Y_n}$, $\msgf{V}{Y_n}$, $\hat y_n$, etc.,
          as functions of $\sigma^2$
          (as detailed in Section~\ref{sec:FFBDD:Parametric}).
          %\\
          %We thus obtain new $\msgf{m}{X_n}(\sigma^2)$, 
          %and new $\hat y_n(\sigma^2)$ for all $n$.
          \\[0.5ex]
   \STATE %\label{line:alg:PBFFD4LSV2:eoas}
          For every $n$, determine 
          \begin{equation} %\label{eqn:alg:FFBDD:sigman}
            \sigma_{\ell,n}^2 = \sup \big\{ \sigma^2 : \text{$\hat y_n(\sigma^2)$ is in the active subdomain of $\kappa_n(y_n)$} \big\},
          \end{equation}
          as detailed in Section~\ref{sec:PathFFBDD}.
          \\[0.5ex]
   \STATE Let $\sigma_\ell^2 = \min_n\big\{ \sigma_{\ell,n}^2 :  \sigma_{\ell,n}^2 > \sigma_{\ell-1}^2 \big\}$. 
   \IF{ $\sigma_\ell^2 < \infty$ }
       \STATE %\label{line:alg:PBFFD4LSV2:iftext}
          For the index $n$ 
          such that $\sigma_{\ell,n}^2 = \sigma_\ell^2$,
          change the active segment and subdomain of $\kappa_n(y_n)$ as follows:
          if the present active segment is a line segment,
          then the next active segment is the point segment whose subdomain contains $\hat y_n(\sigma_\ell^2)$;
          if the present active segment is a point segment,
          then the next active segment is the line segment whose subdomain contains $\hat y_n(\sigma_\ell^2+\varepsilon)$
          with positive $\varepsilon \rightarrow 0$.
       \STATE Update $c_{n}^\T \msgf{V}{X_{n-1}}$ for all $n$ as described 
          in Sections~\ref{sec:pathFFBDD:SV:line2point} and \ref{sec:pathFFBDD:SV:point2line}.
       \STATE Rearrange and re-index the sequence of variables $Y_1,\ldots,Y_N$ such that (\ref{eqn:pathSV:sortedcond}) holds.
   \ENDIF \label{line:alg:PathFFBDD4LSV:endif}
\UNTIL $\sigma_\ell^2 = \infty$.
\end{algorithmic}
\end{trivalgorithm}

\end{minipage}%
}
\end{table}

\section{Regularized Output: Refinements for Generic $F$ (Fig.~\protect\ref{fig:FactorGraph:DecompOutput})}
\label{sec:RefineFig3}

Specializing Algorithm~\ref{alg:BFFD:SSM} 
to Fig.~\ref{fig:FactorGraph:DecompOutput}
yields Algorithm~\ref{alg:FFBDD:DecompOutput}.
The complexity of the forward recursion (Steps 1 and~2) is $\calO(NM^2)$
while the complexity of the backward recursion (Steps 3 to~5) is only $\calO(NM)$.

As in Section~\ref{sec:RefineFig2}, 
it turns out that the complexity of PathFFBDD can be reduced to $\calO(N M)$ per iteration.
The key insight is that  
$c_{n}^\T \msgf{V}{X_{n-1}}\in\R^{1\times M}$
can be updated in each iteration with complexity $\calO(NM)$, as will be explained below.
In consequence, Algorithm~\ref{alg:FFBDD:DecompOutput} can be simplified 
to Algorithm~\ref{alg:BFFD:DecompOutput:SpecUpd}
(which has complexity $\calO(N M)$ per iteration),
and Algorithm~\ref{alg:PathFFBDD} (PathFFBDD) is changed into Algorithm~\ref{alg:PathFFBDD4SV}.
Note that Algorithm~\ref{alg:PathFFBDD4SV} is dual to Algorithm~\ref{alg:PathBFFD4LSV2}
rather than to Algorithm~\ref{alg:PathBFFD4LS}, but unlike Algorithm~\ref{alg:PathBFFD4LSV2},
it puts no restrictions on $\kappa_n(y_n)$.

\begin{figure*}[tp]
\centering
\begin{tikzpicture}[x=1mm, y=1mm, >=latex]
\tikzset{%
smallbox/.style = {draw, rectangle, inner sep=0mm, minimum width=4mm, minimum height=4mm},
%midbox/.style = {draw, rectangle, minimum width=5.5mm, minimum height=5.5mm, inner sep=0.5mm},
midbox/.style = {draw, rectangle, minimum width=5.5mm, minimum height=5.5mm, inner sep=1mm},
blobbox/.style = {draw, fill=black, rectangle, inner sep=0mm, minimum width=1.75mm, minimum height=1.75mm},
xibox/.style = {draw, pattern=north east lines, inner sep=0mm, minimum width=1.75mm, minimum height=1.75mm},
}
\node[smallbox, label={left:$\exp\mleft( - \frac{\| x_0 \|^2}{2\sigma^2} \mright)$}] (pX0) {};
\draw (pX0)+(15,0) node[smallbox] (equY1) {$=$};
\draw[->] (pX0) -- node[above]{$X_0$} (equY1);
\draw[->] (equY1) -- node[above] {$X_1$} +(11,0);
\draw (equY1)+(0,-11) node[midbox] (c1) {$c_1^\T$};
\draw[->] (equY1) -- (c1);
\draw (c1)+(0,-12) node[xibox] (pY1) {};
\draw[->] (c1) -- node[left] {$Y_1$} (pY1);
\draw (equY1)+(35,0) node[smallbox] (equYnu) {$=$};
\path (equY1) -- node[pos=0.5] {$\cdots$} (equYnu);
\draw[<-] (equYnu) -- node[above,pos=0.6]{$X_{r-1}$} +(-12.5,0);
\draw (equYnu)+(0,-11) node[midbox] (cnu) {$c_{r}^\T$};
\draw[->] (equYnu) -- (cnu);
\draw (cnu)+(0,-12) node[xibox] (pYnu) {};
\draw[->] (cnu) -- node[left] {$Y_{r}$} (pYnu);
\draw (equYnu)+(15,0) node[smallbox] (equYnu1) {$=$};
\draw[->] (equYnu) --node[above]{$X_{r}$} (equYnu1);
\draw[->] (equYnu1) -- node[above] {$X_{r+1}$} +(12.5,0);
\draw (equYnu1)+(0,-11) node[midbox] (cnu1) {$c_{r+1}^\T$};
\draw[->] (equYnu1) -- (cnu1);
\draw (cnu1)+(0,-12) node[blobbox] (pYnu1) {};
\draw[->] (cnu1) -- node[right] {$Y_{r+1}$} (pYnu1);
\draw (equYnu1)+(35,0) node[smallbox] (equYN) {$=$};
\path (equYnu1) -- node[pos=0.5] {$\cdots$} (equYN);
\draw[<-] (equYN) -- node[above,pos=0.6]{$X_{N-1}$} +(-12.5,0);
\draw (equYN)+(0,-11) node[midbox] (cN) {$c_N^\T$};
\draw[->] (equYN) -- (cN);
\draw (cN)+(0,-12) node[blobbox] (pYN) {};
\draw[->] (cN) -- node[right] {$Y_N$} (pYN);
\draw[->] (equYN) -- node[above]{$X_N$} +(12.5,0) {};
\end{tikzpicture}
\newcommand{\blobbox}{%
\begin{tikzpicture}[scale=0.1]
\protect\node[draw, fill=black, inner sep=0mm, minimum width=1.75mm, minimum height=1.75mm] {};
\end{tikzpicture}%
}
\newcommand{\xibox}{%
\begin{tikzpicture}[scale=0.1]
\protect\node[draw, pattern=north east lines, inner sep=0mm, minimum width=1.75mm, minimum height=1.75mm] {};
\end{tikzpicture}%
}
%
%\vspace{1ex}
\caption{\label{fig:FactorGraph:DecompOutput:Sorted}%
    For Section~\ref{sec:RefineFig3}: 
    Fig.~\ref{fig:FactorGraph:DecompOutput} with 
    rearranged and re-indexed variables $Y_1,\ldots,Y_N$
    and correspondingly re-indexed $c_1,\ldots,c_N$
    such that (\ref{eqn:pathSV:sortedcond}) holds.
    The symbols ``$\,\xibox\,$'' and ``$\,\blobbox\,$''
    denote factors (\ref{eqn:GenGaussian}) 
    with $\msgb{W}{Y_n}=0$ and $\msgb{V}{Y_n}=0$, respectively.
    %cf.\ (\ref{eqn:CostPieces:NUP}).
    }
%\end{figure*}
\vspace{2\floatsep}

%\begin{figure*}[tp]
\centering
\begin{tikzpicture}[x=1mm, y=1mm, >=latex]
\tikzset{%
smallbox/.style = {draw, rectangle, inner sep=0mm, minimum width=4mm, minimum height=4mm},
%midbox/.style = {draw, rectangle, minimum width=5.5mm, minimum height=5.5mm, inner sep=0.5mm},
midbox/.style = {draw, rectangle, minimum width=5.5mm, minimum height=5.5mm, inner sep=1mm},
blobbox/.style = {draw, fill=black, rectangle, inner sep=0mm, minimum width=1.75mm, minimum height=1.75mm},
xibox/.style = {draw, pattern=north east lines, inner sep=0mm, minimum width=1.75mm, minimum height=1.75mm},
}
\node[smallbox, label={left:$\exp\mleft( - \frac{\| x_0 \|^2}{2\sigma^2} \mright)$}] (pX0) {};
\draw (pX0)+(15,0) node[smallbox] (equY1) {$=$};
\draw[->] (pX0) -- node[above]{$X_0$} (equY1);
\draw[->] (equY1) -- node[above] {$X_1$} +(11,0);
\draw (equY1)+(0,-11) node[midbox] (c1) {$c_1^\T$};
\draw[->] (equY1) -- (c1);
\draw (c1)+(0,-12) node[xibox] (pY1) {};
\draw[->] (c1) -- node[left] {$Y_1$} (pY1);
\draw (equY1)+(35,0) node[smallbox] (equYnu) {$=$};
\path (equY1) -- node[pos=0.5] {$\cdots$} (equYnu);
\draw[<-] (equYnu) -- node[above,pos=0.6]{$X_{k-2}$} +(-12.5,0);
\draw (equYnu)+(0,-11) node[midbox] (cnu) {$c_{k-1}^\T$};
\draw[->] (equYnu) -- (cnu);
\draw (cnu)+(0,-12) node[xibox] (pYnu) {};
\draw[->] (cnu) -- node[left] {$Y_{k-1}$} (pYnu);
\draw (equYnu)+(15,0) node[smallbox] (equYnu1) {$=$};
\draw[->] (equYnu) --node[above]{$X_k$} (equYnu1);
\draw[->] (equYnu1) -- node[above] {$X_{k+1}$} +(12.5,0);
\draw (equYnu1)+(0,-11) node[midbox] (cnu1) {$c_{k+1}^\T$};
\draw[->] (equYnu1) -- (cnu1);
\draw (cnu1)+(0,-12) node[xibox] (pYnu1) {};
\draw[->] (cnu1) -- node[right] {$Y_{k+1}$} (pYnu1);
\draw (equYnu1)+(35,0) node[smallbox] (equYN) {$=$};
\path (equYnu1) -- node[pos=0.5] {$\cdots$} (equYN);
\draw[<-] (equYN) -- node[above,pos=0.6]{$X_{N-1}$} +(-12.5,0);
\draw (equYN)+(0,-11) node[midbox] (cN) {$c_N^\T$};
\draw[->] (equYN) -- (cN);
\draw (cN)+(0,-12) node[blobbox] (pYN) {};
\draw[->] (cN) -- node[right] {$Y_N$} (pYN);
%\draw[->] (equYN) -- node[above]{$X_N$} +(12.5,0) {};
%
\draw (equYN)+(15,0) node[smallbox] (equTail) {$=$};
\draw[->] (equYN) -- node[above]{$X_{k-1}$} (equTail); 
\draw[->] (equTail) -- node[above]{$X_N$} +(12.5,0) {};
\draw (equTail)+(0,-11) node[midbox] (ck) {$c_k^\T$};
\draw[->] (equTail) -- (ck);
\draw (ck)+(0,-12) node[blobbox] (pYk) {};
\draw[->] (ck) -- node[right] {$Y_k$} (pYk);
\end{tikzpicture}
%
%\vspace{1ex} 
\caption{\label{fig:FactorGraph:DecompOutput:UnknownToKnown}%
   Rearranged factor graph when the active segment of $\kappa_k(y_k)$ changes from a line to a point,
   cf.\ Section~\ref{sec:pathFFBDD:SV:line2point}.
    }
\end{figure*}

It remains to explain the update of $c_{n}^\T \msgf{V}{X_{n-1}}$ in every iteration.
To this end, 
we will rearrange and re-index the variables $Y_1,\ldots,Y_N$
in every iteration of PathFFBDD 
such that the following condition holds:
\begin{equation} \label{eqn:pathSV:sortedcond}
\text{\parbox{0.8\linewidth}{\centering 
For some index $r$,
the active segments of $\kappa_1(y_1),\ldots,\kappa_{r}(y_{r})$ are lines\\
and the active segments of $\kappa_{r+1}(y_{r+1}),\ldots,\kappa_N(y_N)$ are points.
}}
\end{equation}
The corresponding factor graph is shown in Fig.~\ref{fig:FactorGraph:DecompOutput:Sorted}.
%(In the terminology of path methods, the active set comprises $\{ Y_{r+1},\ldots,Y_N \}$ 
%and has cardinality $N-r$.)
%Note that keeping track of the changing indices requires maintaining 
%two tables of size $\calO(N)$.
%
In the following, 
$k$ denotes the index $n$
for which the active segment of $\kappa_n(y_n)$ changes
in the $\ell$-th iteration.

\subsection{Updating $c_{n}^\T \msgf{V}{X_{n-1}}$ when Changing a Line to a Point}
\label{sec:pathFFBDD:SV:line2point}

If the active segment of $\kappa_k(y_k)$ changes from a line to a point
(which implies $k\leq r$), 
the procedure is as follows.
\begin{enumerate}
\item
Consider the factor graph of 
Fig.~\ref{fig:FactorGraph:DecompOutput:UnknownToKnown},
which is obtained from Fig.~\ref{fig:FactorGraph:DecompOutput:Sorted}
by moving $Y_k$ 
to the tail. 
\item
Compute $c_{n}^\T \msgf{V}{X_{n-1}}$ 
in Fig.~\ref{fig:FactorGraph:DecompOutput:UnknownToKnown}
for all $n$ according to Proposition~\ref{prop:pathFFBDD:line2point:key} below.
\item
Re-index all variables 
in Fig.~\ref{fig:FactorGraph:DecompOutput:UnknownToKnown}
back to the notation in Fig.~\ref{fig:FactorGraph:DecompOutput:Sorted}.
\end{enumerate}
The complexity of this procedure is $\calO(NM)$.

\begin{proposition} \label{prop:pathFFBDD:line2point:key}
Let $\msgf{V}{X_{n-1}}$ in Figures \ref{fig:FactorGraph:DecompOutput:Sorted} 
and~\ref{fig:FactorGraph:DecompOutput:UnknownToKnown} be denoted by 
$\msgf{V}{X_{n-1}}^{\text{old}}$ and $\msgf{V}{X_{n-1}}^{\text{new}}$,
respectively. Then
\begin{equation} \label{eqn:pathFFBDD:line2point:bnWXn}
c_n^\T \msgf{V}{X_{n-1}}^{\text{new}} = c_n^\T \msgf{V}{X_{n-1}}^{\text{old}}
\end{equation}
for $n\in \{1,\ldots,N \} \setminus \{ k \}$ and
\begin{equation} \label{eqn:pathFFBDD:line2point:key}
c_k^\T \msgf{V}{X_{k-1}}^{\text{new}} = \sigma^2 c_k^\T - \sum_{n=r+1}^N
         c_k^\T \big( c_n^\T \msgf{V}{X_{n-1}}^{\text{old}} \big)^\T 
         \big( c_n^\T \msgf{V}{X_{n-1}}^{\text{old}} c_n \big)^{-1} 
         c_n^\T \msgf{V}{X_{n-1}}^{\text{old}}.
\end{equation}
Note that (\ref{eqn:pathFFBDD:line2point:key}) can be computed 
with complexity $O(NM)$.
\end{proposition}

\begin{proofof}{of Proposition~\ref{prop:pathFFBDD:line2point:key}}
From (\ref{eqn:alg:FFBDD:DecompOutput:forw:G}) and (\ref{eqn:alg:FFBDD:DecompOutput:forw:VX}),
we see first, that 
\begin{equation} \label{eqn:pathFFBDD:line2point:unchangedmsgfV}
%\msgf{V}{X_{n-1}}^{\text{new}} = \msgf{V}{X_{n-1}}^{\text{old}} = \sigma^2 I
\msgf{V}{X_n}^{\text{new}} = \msgf{V}{X_n}^{\text{old}} = \sigma^2 I
\end{equation}
for $n\in \{0,\ldots,r \} \setminus \{ k-1 \}$,
and second, that 
%(\ref{eqn:pathFFBDD:line2point:bnWXn}) holds for $n\in \{1,\ldots,N \} \setminus \{ k \}$.
$\msgf{V}{X_n}^{\text{new}} = \msgf{V}{X_n}^{\text{old}}$ 
holds also for $n\in \{ r+1, \ldots, N-1 \}$,
which proves (\ref{eqn:pathFFBDD:line2point:bnWXn}).
From (\ref{eqn:alg:FFBDD:DecompOutput:forw:VX}), we further obtain
\begin{IEEEeqnarray}{r,C,l}
c_k^\T \msgf{V}{X_{k-1}}^{\text{new}}
& = &  c_k^\T \bigg( \msgf{V}{X_{N-1}}^{\text{new}}
       - \msgf{V}{X_{N-1}}^{\text{new}} c_N \big( c_N^\T \msgf{V}{X_{N-1}}^{\text{new}} c_N \big)^{-1} 
         c_N^\T \msgf{V}{X_{N-1}}^{\text{new}}
       \bigg)\\
& = &  c_k^\T \bigg( \msgf{V}{X_{N-2}}^{\text{new}}
       - \msgf{V}{X_{N-2}}^{\text{new}} c_{N-1} \big( c_{N-1}^\T \msgf{V}{X_{N-2}}^{\text{new}} c_{N-1} \big)^{-1} 
         c_{N-1}^\T \msgf{V}{X_{N-2}}^{\text{new}}
       \bigg)
       - c_k^\T \msgf{V}{X_{N-1}}^{\text{new}} c_N \big( c_N^\T \msgf{V}{X_{N-1}}^{\text{new}} c_N \big)^{-1} 
         c_N^\T \msgf{V}{X_{N-1}}^{\text{new}}
       \IEEEeqnarraynumspace\\
& = &  c_k^\T \msgf{V}{X_{r}}^{\text{new}} 
        -  \sum_{n=r+1}^N
         c_k^\T \msgf{V}{X_{n-1}}^{\text{new}} c_n \big( c_n^\T \msgf{V}{X_{n-1}}^{\text{new}} c_n \big)^{-1} 
         c_n^\T \msgf{V}{X_{n-1}}^{\text{new}},
\end{IEEEeqnarray}
and using (\ref{eqn:pathFFBDD:line2point:unchangedmsgfV}) and (\ref{eqn:pathFFBDD:line2point:bnWXn}) 
yields (\ref{eqn:pathFFBDD:line2point:key}).
\end{proofof}

%\clearpage

\begin{figure}[tp]
\centering
\newcommand{\extravspace}{\vspace{2mm}}

%\subfloat[Initial factor graph]{%
\subfloat[\rule{0em}{3ex}Initial factor graph]{%
\label{subfig:SVM:point2line:initial}%
\begin{tikzpicture}[x=1mm, y=1mm, >=latex]
\tikzset{%
smallbox/.style = {draw, rectangle, inner sep=0mm, minimum width=4mm, minimum height=4mm},
midbox/.style = {draw, rectangle, minimum width=5.5mm, minimum height=5.5mm, inner sep=0.5mm},
blobbox/.style = {draw, fill=black, rectangle, inner sep=0mm, minimum width=1.75mm, minimum height=1.75mm},
xibox/.style = {draw, pattern=north east lines, inner sep=0mm, minimum width=1.75mm, minimum height=1.75mm},
}
\node[smallbox, label={left:$\exp\mleft(- \frac{\|x_0\|^2}{2\sigma^2}\mright)$}] (x0) {};
\draw (x0)+(15,0) node[smallbox] (equY1) {};
\draw (x0) -- node[above] {$X_0$} (equY1);
\draw (equY1)+(0,-12.5) node[xibox] (pY1) {};
\draw[->] (equY1) -- node[right] {$Y_1$} (pY1);
\draw (equY1) -- node[above,pos=0.6] {$X_1$} +(10,0);
\draw (equY1)+(42.5,0) node[smallbox] (equYk) {};
\path (equY1) -- node[pos=0.33] {$\cdots$} (equYk);
\draw[<-] (equYk) -- node[above,pos=0.55] {$X_{k-1}^{(0)} \!=\! X_{k-1}$} +(-22.5,0);
\draw (equYk)+(0,-12.5) node[blobbox] (pYk) {};
\draw[->] (equYk) -- node[right] {$Y_k$} (pYk);
\draw (equYk)+(15,0) node[smallbox] (equYkp1) {};
\draw[->] (equYk) -- node[above]{$X_k$} (equYkp1);
\draw (equYkp1)+(0,-12.5) node[blobbox] (pYkp1) {};
\draw[->] (equYkp1) -- node[right] {$Y_{k+1}$} (pYkp1);
\draw (equYkp1)+(15,0) node[smallbox] (equYkp2) {};
\draw[->] (equYkp1) -- node[above]{$X_{k+1}$} (equYkp2);
\draw (equYkp2)+(0,-12.5) node[blobbox] (pYkp2) {};
\draw[->] (equYkp2) -- node[right] {$Y_{k+2}$} (pYkp2);
%\draw (equYkp2) -- node[above,pos=0.6] {$X_k^{(0)}\!=\!X_k$} +(15,0);
\draw (equYkp2) -- node[above,pos=0.6] {$X_{k+2}$} +(12.5,0);
\draw (equYkp2)+(35,0) node[smallbox] (equYN) {};
\path (equYkp2) -- node[pos=0.5] {$\cdots$} (equYN);
\draw[<-] (equYN) -- node[above,pos=0.6] {$X_{N-1}$} +(-12.5,0);
\draw (equYN)+(0,-12.5) node[blobbox] (pYN) {};
\draw[->] (equYN) -- node[right] {$Y_N$} (pYN);
\draw[->] (equYN) -- node[above]{$X_N$} +(12.5,0);
\end{tikzpicture}%
}
\extravspace

\subfloat[\rule{0em}{3ex}\mbox{Moving $Y_k$ right by one position.}]{%
\label{subfig:LS:line2point:moveright1}%
\begin{tikzpicture}[x=1mm, y=1mm, >=latex]
\tikzset{%
smallbox/.style = {draw, rectangle, inner sep=0mm, minimum width=4mm, minimum height=4mm},
midbox/.style = {draw, rectangle, minimum width=5.5mm, minimum height=5.5mm, inner sep=0.5mm},
blobbox/.style = {draw, fill=black, rectangle, inner sep=0mm, minimum width=1.75mm, minimum height=1.75mm},
xibox/.style = {draw, pattern=north east lines, inner sep=0mm, minimum width=1.75mm, minimum height=1.75mm},
}
\node[smallbox, label={left:$\exp\mleft(- \frac{\|x_0\|^2}{2\sigma^2}\mright)$}] (x0) {};
\draw (x0)+(15,0) node[smallbox] (equY1) {};
\draw (x0) -- node[above] {$X_0$} (equY1);
\draw (equY1)+(0,-12.5) node[xibox] (pY1) {};
\draw[->] (equY1) -- node[right] {$Y_1$} (pY1);
\draw (equY1) -- node[above,pos=0.6] {$X_1$} +(10,0);
\draw (equY1)+(32.5,0) node[smallbox] (equYk) {};
\path (equY1) -- node[pos=0.47] {$\cdots$} (equYk);
\draw[<-] (equYk) -- node[above,pos=0.55] {$X_k$} +(-12.5,0);
\draw (equYk)+(0,-12.5) node[blobbox] (pYk) {};
\draw[->] (equYk) -- node[right] {$Y_{k+1}$} (pYk);
\draw (equYk)+(15,0) node[smallbox] (equYkp1) {};
\draw[->] (equYk) -- node[above]{$X_{k-1}^{(1)}$} (equYkp1);
\draw (equYkp1)+(0,-12.5) node[blobbox] (pYkp1) {};
\draw[->] (equYkp1) -- node[right] {$Y_k$} (pYkp1);
\draw (equYkp1)+(15,0) node[smallbox] (equYkp2) {};
\draw[->] (equYkp1) -- node[above]{$X_{k+1}$} (equYkp2);
\draw (equYkp2)+(0,-12.5) node[blobbox] (pYkp2) {};
\draw[->] (equYkp2) -- node[right] {$Y_{k+2}$} (pYkp2);
%\draw (equYkp2) -- node[above,pos=0.6] {$X_k^{(0)}\!=\!X_k$} +(15,0);
\draw (equYkp2) -- node[above,pos=0.6] {$X_{k+2}$} +(12.5,0);
\draw (equYkp2)+(35,0) node[smallbox] (equYN) {};
\path (equYkp2) -- node[pos=0.5] {$\cdots$} (equYN);
\draw[<-] (equYN) -- node[above,pos=0.6] {$X_{N-1}$} +(-12.5,0);
\draw (equYN)+(0,-12.5) node[blobbox] (pYN) {};
\draw[->] (equYN) -- node[right] {$Y_N$} (pYN);
\draw[->] (equYN) -- node[above]{$X_N$} +(12.5,0);
\end{tikzpicture}%
}
\extravspace

\subfloat[\rule{0em}{3ex}\mbox{Moving $Y_k$ right by two positions.}]{%
\label{subfig:LS:line2point:moveright2}%
\begin{tikzpicture}[x=1mm, y=1mm, >=latex]
\tikzset{%
smallbox/.style = {draw, rectangle, inner sep=0mm, minimum width=4mm, minimum height=4mm},
midbox/.style = {draw, rectangle, minimum width=5.5mm, minimum height=5.5mm, inner sep=0.5mm},
blobbox/.style = {draw, fill=black, rectangle, inner sep=0mm, minimum width=1.75mm, minimum height=1.75mm},
xibox/.style = {draw, pattern=north east lines, inner sep=0mm, minimum width=1.75mm, minimum height=1.75mm},
}
\node[smallbox, label={left:$\exp\mleft(- \frac{\|x_0\|^2}{2\sigma^2}\mright)$}] (x0) {};
\draw (x0)+(15,0) node[smallbox] (equY1) {};
\draw (x0) -- node[above] {$X_0$} (equY1);
\draw (equY1)+(0,-12.5) node[xibox] (pY1) {};
\draw[->] (equY1) -- node[right] {$Y_1$} (pY1);
\draw (equY1) -- node[above,pos=0.6] {$X_1$} +(10,0);
\draw (equY1)+(32.5,0) node[smallbox] (equYk) {};
\path (equY1) -- node[pos=0.47] {$\cdots$} (equYk);
\draw[<-] (equYk) -- node[above,pos=0.55] {$X_k$} +(-12.5,0);
\draw (equYk)+(0,-12.5) node[blobbox] (pYk) {};
\draw[->] (equYk) -- node[right] {$Y_{k+1}$} (pYk);
\draw (equYk)+(15,0) node[smallbox] (equYkp1) {};
\draw[->] (equYk) -- node[above]{$X_{k+1}$} (equYkp1);
\draw (equYkp1)+(0,-12.5) node[blobbox] (pYkp1) {};
\draw[->] (equYkp1) -- node[right] {$Y_{k+2}$} (pYkp1);
\draw (equYkp1)+(15,0) node[smallbox] (equYkp2) {};
\draw[->] (equYkp1) -- node[above]{$X_{k-1}^{(2)}$} (equYkp2);
\draw (equYkp2)+(0,-12.5) node[blobbox] (pYkp2) {};
\draw[->] (equYkp2) -- node[right] {$Y_k$} (pYkp2);
%\draw (equYkp2) -- node[above,pos=0.6] {$X_k^{(0)}\!=\!X_k$} +(15,0);
\draw (equYkp2) -- node[above,pos=0.6] {$X_{k+2}$} +(12.5,0);
\draw (equYkp2)+(35,0) node[smallbox] (equYN) {};
\path (equYkp2) -- node[pos=0.5] {$\cdots$} (equYN);
\draw[<-] (equYN) -- node[above,pos=0.6] {$X_{N-1}$} +(-12.5,0);
\draw (equYN)+(0,-12.5) node[blobbox] (pYN) {};
\draw[->] (equYN) -- node[right] {$Y_N$} (pYN);
\draw[->] (equYN) -- node[above]{$X_N$} +(12.5,0);
\end{tikzpicture}%
}
\extravspace

\subfloat[\rule{0em}{3ex}\mbox{Moving $Y_k$ (by $N-k$ positions) to the tail.}]{%
\label{subfig:LS:line2point:attail}%
\begin{tikzpicture}[x=1mm, y=1mm, >=latex]
\tikzset{%
smallbox/.style = {draw, rectangle, inner sep=0mm, minimum width=4mm, minimum height=4mm},
midbox/.style = {draw, rectangle, minimum width=5.5mm, minimum height=5.5mm, inner sep=0.5mm},
blobbox/.style = {draw, fill=black, rectangle, inner sep=0mm, minimum width=1.75mm, minimum height=1.75mm},
xibox/.style = {draw, pattern=north east lines, inner sep=0mm, minimum width=1.75mm, minimum height=1.75mm},
}
\node[smallbox, label={left:$\exp\mleft(- \frac{\|x_0\|^2}{2\sigma^2}\mright)$}] (x0) {};
\draw (x0)+(15,0) node[smallbox] (equY1) {};
\draw[->] (x0) -- node[above]{$X_0$} (equY1);
\draw (equY1)+(0,-12.5) node[xibox] (pY1) {};
\draw[->] (equY1) -- node[right] {$Y_1$} (pY1);
\draw (equY1) -- node[above]{$X_1$} +(10,0);
\draw (equY1)+(32.5,0) node[smallbox] (equYN) {};
\path (equY1) -- node[pos=0.45] {$\cdots$} (equYN);
\draw[<-] (equYN) -- node[above,pos=0.6] {$X_{N-1}$} +(-12.5,0);
\draw (equYN)+(0,-12.5) node[blobbox] (pYN) {};
\draw[->] (equYN) -- node[right] {$Y_N$} (pYN);
\draw (equYN)+(15,0) node[smallbox] (equYk) {};
\draw[->] (equYN) -- node[above,pos=0.57]{$X_{k-1}^{(N-k)}$} (equYk);
\draw (equYk)+(0,-12.5) node[blobbox] (pYk) {};
\draw[->] (equYk) -- node[right] {$Y_k$} (pYk);
\draw[->] (equYk) -- node[above]{$X_N$} +(12.5,0);
\end{tikzpicture}%
}
\extravspace

%\subfloat[\rule{0em}{3ex}\mbox{Changing the active segment of $\kappa_k(y_k)$ from a point to a line and moving it to the head.}]{%
%\subfloat[\rule{0em}{3ex}\mbox{Moving $Y_k$ to the head and changing its active segment.}]{%
\subfloat[\rule{0em}{3ex}\mbox{Moving $Y_k$ to the head and changing the active segment of $\kappa_k(y_k)$.}]{%
\label{subfig:SVM:point2line:finalathead}%
\begin{tikzpicture}[x=1mm, y=1mm, >=latex]
\tikzset{%
smallbox/.style = {draw, rectangle, inner sep=0mm, minimum width=4mm, minimum height=4mm},
midbox/.style = {draw, rectangle, minimum width=5.5mm, minimum height=5.5mm, inner sep=0.5mm},
blobbox/.style = {draw, fill=black, rectangle, inner sep=0mm, minimum width=1.75mm, minimum height=1.75mm},
xibox/.style = {draw, pattern=north east lines, inner sep=0mm, minimum width=1.75mm, minimum height=1.75mm},
}
\node[smallbox, label={left:$\exp\mleft(- \frac{\|x_0\|^2}{2\sigma^2}\mright)$}] (xkm1) {};
\draw (xkm1)+(17.5,0) node[smallbox] (equYk) {};
\draw[->] (xkm1) -- node[above]{$X_{k-1}$} (equYk);
\draw (equYk)+(0,-12.5) node[xibox] (pYk) {};
\draw[->] (equYk) -- node[right] {$Y_k$} (pYk);
\draw (equYk)+(15,0) node[smallbox] (equY1) {};
\draw (equYk) -- node[above] {$X_0$} (equY1);
\draw (equY1)+(0,-12.5) node[xibox] (pY1) {};
\draw[->] (equY1) -- node[right] {$Y_1$} (pY1);
\draw (equY1) -- node[above,pos=0.6] {$X_1$} +(10,0);
\draw (equY1)+(32.5,0) node[smallbox] (equYN) {};
\path (equY1) -- node[pos=0.45] {$\cdots$} (equYN);
\draw[<-] (equYN) -- node[above,pos=0.6] {$X_{N-1}$} +(-12.5,0);
\draw (equYN)+(0,-12.5) node[blobbox] (pYN) {};
\draw[->] (equYN) -- node[right] {$Y_N$} (pYN);
\draw[->] (equYN) -- node[above]{$X_N$} +(12.5,0);
\end{tikzpicture}%
}

\caption{\label{fig:SVM:point2line}%
For Section~\ref{sec:pathFFBDD:SV:point2line}: 
step-by-step rearrangement when the active segment of $\kappa_k(y_k)$
changes from a point to a line,
using the notation of Fig.~\ref{fig:SVM:SimplifiedGF}.
}
\end{figure}
%\vspace{\floatsep}

\begin{figure}[tp]
\centering
\begin{tikzpicture}[x=1mm, y=1mm, >=latex]
\tikzset{%
smallbox/.style = {draw, rectangle, inner sep=0mm, minimum width=4mm, minimum height=4mm},
midbox/.style = {draw, rectangle, minimum width=5.5mm, minimum height=5.5mm, inner sep=1mm},
%blobbox/.style = {draw, fill=black, rectangle, inner sep=0mm, minimum width=1.75mm, minimum height=1.75mm},
%xibox/.style = {draw, pattern=north east lines, inner sep=0mm, minimum width=1.75mm, minimum height=1.75mm},
tinybox/.style = {draw, rectangle, inner sep=0mm, minimum width=1.75mm, minimum height=1.75mm},
}
\node[smallbox] (equ) {$=$};
\draw[<-] (equ) -- node[above,pos=0.65] {$X_{n-1}$} (-17.5,0); 
\draw[->] (equ) -- (15,0);
\draw (equ)+(0,-10) node[midbox] (b) {$c_n^\T$};
\draw[->] (equ) -- (b);
\draw (b)+(0,-15) node[tinybox] (pY) {};
\draw[->] (b) -- node[right, pos=0.67] {$Y_n$} (pY);
\draw (0,5) node[draw, rectangle, anchor=north, dashed, minimum width=12.5mm, minimum height=21mm] {};

\draw (equ)+(40,0) node[smallbox] (simplebox) {};
\draw[<-] (simplebox) -- node[above,pos=0.6]{$X_{n-1}$} +(-12.5,0);
\draw[->] (simplebox) -- +(12.5,0);
\draw (simplebox)+(0,-12.5) node[tinybox] (pY2) {};
\draw[->] (simplebox) -- node[right] {$Y_n$} (pY2);

\draw (equ)+(20,-7.5) node {$=$};
\end{tikzpicture}
\vspace{1ex}
\caption{\label{fig:SVM:SimplifiedGF}%
The notation used in Fig.~\ref{fig:SVM:point2line}
results from closing the dashed box.
The incoming edge $X_{n-1}$ is kept together with $Y_n$ under rearrangements.
}
\end{figure}

\subsection{Updating $c_{n}^\T \msgf{V}{X_{n-1}}$ when Changing a Point to a Line}
\label{sec:pathFFBDD:SV:point2line}

If the active segment of $\kappa_k(y_k)$ changes from a point to a line (which implies $k>r$),
the procedure is as follows.
\begin{enumerate}
\item
Consider the factor graph of Fig.~\ref{subfig:SVM:point2line:finalathead},
which is obtained from Fig.~\ref{subfig:SVM:point2line:initial}
by moving $Y_k$ to the head of the factor graph.
(The intermediate Figures \ref{subfig:LS:line2point:moveright1}--\ref{subfig:LS:line2point:attail}
are used in the proof of Proposition~\ref{prop:pathFFBDD:point2line:key} below.)
\item
Compute $c_n^\T \msgf{V}{X_{n-1}}$ in Fig.~\ref{subfig:SVM:point2line:finalathead}
for all $n$ according to Proposition~\ref{prop:pathFFBDD:point2line:key}.
\item
Re-index all variables in Fig.~\ref{subfig:SVM:point2line:finalathead}
back to the notation in Fig.~\ref{fig:FactorGraph:DecompOutput:Sorted}.
\end{enumerate}
The total complexity of this procedure is $\calO(NM)$.

\begin{proposition} \label{prop:pathFFBDD:point2line:key}
Let $\msgf{V}{X_n}$
in Figures \ref{subfig:SVM:point2line:initial} and~\ref{subfig:SVM:point2line:finalathead}
be denoted $\msgf{V}{X_n}^{\text{old}}$ and $\msgf{V}{X_n}^{\text{new}}$,
respectively. Then
\begin{equation} \label{eqn:pathFFBDD:SV:point2line:nothing2do}
c_n^\T \msgf{V}{X_{n-1}}^{\text{new}} = c_n^\T \msgf{V}{X_{n-1}}^{\text{old}}
\text{~~~for $n<k$,}
\end{equation}
%for $n<k$,
\begin{equation} \label{eqn:pathFFBDD:SV:point2line:obvious}
c_k^\T \msgf{V}{X_{k-1}}^{\text{new}} = \sigma^2 c_k^\T,
\end{equation}
and for $n>k$, 
we have the coupled recursions
\begin{IEEEeqnarray}{r,C,l}
c_{k+\nu}^\T \msgf{V}{X_{k+\nu-1}}^{\text{new}}
& = &  c_{k+\nu}^\T \msgf{V}{X_{k+\nu-1}}^{\text{old}}
       + c_{k+\nu}^\T  \Big( c_k^\T \msgf{V}{X_{k-1}^{(\nu-1)}} \Big)^\T
          \Big( c_k^\T \msgf{V}{X_{k-1}^{(\nu-1)}} c_k \Big)^{-1}
           c_k^\T \msgf{V}{X_{k-1}^{(\nu-1)}}
         \label{eqn:pathFFBDD:SV:point2line:VX:recursion}
%\end{IEEEeqnarray}
%and
\\
%\begin{IEEEeqnarray}{r,C,l}
c_k^\T \msgf{V}{X_{k-1}^{(\nu)}}
& = &  c_k^\T \msgf{V}{X_{k-1}^{(\nu-1)}}
       -  c_k^\T \Big( c_{k+\nu}^\T \msgf{V}{X_{k+\nu-1}}^{\text{new}} \Big)^\T
         \Big(c_{k+\nu}^\T \msgf{V}{X_{k+\nu-1}}^{\text{new}} c_{k+\nu} \Big)^{-1} 
         c_{k+\nu}^\T \msgf{V}{X_{k+\nu-1}}^{\text{new}}
         \label{eqn:pathFFBDD:SV:point2line:VXtemp:recursion}
\end{IEEEeqnarray}
for $\nu=1,2,\ldots,N-k,$
%where $\msgf{V}{X_{k-1}^{(\nu)}}$ is an interpretable quantity (as explained below) and
beginning with 
$c_k^\T \msgf{V}{X_{k-1}^{(0)}} = c_k^\T \msgf{V}{X_{k-1}}^{\text{old}}$.
\end{proposition}
(The matrices $\msgf{V}{X_{k-1}^{(\nu)}}$ refer to the edges $X_{k-1}^{(\nu)}$
in Figures \ref{subfig:LS:line2point:moveright1}--\ref{subfig:LS:line2point:attail},
cf.\ the proof below.)
Note that, for fixed $\nu$, 
both (\ref{eqn:pathFFBDD:SV:point2line:VX:recursion}) and (\ref{eqn:pathFFBDD:SV:point2line:VXtemp:recursion})
can be computed with complexity $\calO(M)$.

\begin{proofof}{of Proposition~\ref{prop:pathFFBDD:point2line:key}}
Eq.~(\ref{eqn:pathFFBDD:SV:point2line:obvious}) is obvious from 
$\msgf{V}{X_{k-1}}^{\text{new}} = \sigma^2 I$.
From (\ref{eqn:alg:FFBDD:DecompOutput:forw:G}) and (\ref{eqn:alg:FFBDD:DecompOutput:forw:VX}),
we then see that
\begin{equation} %\label{eqn:pathFFBDD:point2line:unchangedmsgfV}
\msgf{V}{X_n}^{\text{new}} = \msgf{V}{X_n}^{\text{old}} = \sigma^2 I
\end{equation}
for $n\in\{ 0, \ldots, r \}$, 
and $\msgf{V}{X_n}^{\text{new}} = \msgf{V}{X_n}^{\text{old}}$ for $n\in\{r+1,\ldots,k-2\}$,
which implies (\ref{eqn:pathFFBDD:SV:point2line:nothing2do}).

The recursions (\ref{eqn:pathFFBDD:SV:point2line:VX:recursion}) 
and (\ref{eqn:pathFFBDD:SV:point2line:VXtemp:recursion})
are obtained by moving $Y_k$ in $N-k$ steps to the tail of the factor graph, by one position per step,
as shown in Figures \ref{subfig:LS:line2point:moveright1}--\ref{subfig:LS:line2point:attail}.
Note also the variables $X_{k-1}^{(0)} = X_{k-1}$ in Fig.~\ref{subfig:SVM:point2line:initial},
$X_{k-1}^{(1)}$ in Fig.~\ref{subfig:LS:line2point:moveright1},
etc.,
and note further that 
(\ref{eqn:alg:FFBDD:DecompOutput:forw:G}) and (\ref{eqn:alg:FFBDD:DecompOutput:forw:VX})
imply 
%(from Fig.~\ref{fig:LS:line2point} and (\ref{eqn:alg:BFFD:DecompInput:backw:H}) and (\ref{eqn:alg:BFFD:DecompInput:backw:WXppp})) 
$\msgf{V}{X_{k}}^{\text{new}}   = \msgf{V}{X_{k-1}^{(0)}}$,
$\msgf{V}{X_{k+1}}^{\text{new}} = \msgf{V}{X_{k-1}^{(1)}}$,
etc.
In the first step (from Fig.~\ref{subfig:SVM:point2line:initial} to Fig.~\ref{subfig:LS:line2point:moveright1}), 
we have both
\begin{IEEEeqnarray}{r,C,l}
c_{k+1}^\T \msgf{V}{X_k}^{\text{new}}
& = &  c_{k+1}^\T \msgf{V}{X_{k-1}^{(0)}} \\
& = &  c_{k+1}^\T \bigg( \msgf{V}{X_k}^{\text{old}} 
         + \msgf{V}{X_{k-1}^{(0)}} c_k  
          \Big( c_k^\T \msgf{V}{X_{k-1}^{(0)}} c_k \Big)^{-1}
           c_k^\T \msgf{V}{X_{k-1}^{(0)}} \bigg) 
         \label{eqn:pathFFBDD:SV:point2line:VX:MIL}\\
& = &  c_{k+1}^\T \msgf{V}{X_k}^{\text{old}}
       + c_{k+1}^\T  \Big( c_k^\T \msgf{V}{X_{k-1}^{(0)}} \Big)^\T
          \Big( c_k^\T \msgf{V}{X_{k-1}^{(0)}} c_k \Big)^{-1}
           c_k^\T \msgf{V}{X_{k-1}^{(0)}}
         \label{eqn:pathFFBDD:SV:point2line:VX:final}
\end{IEEEeqnarray}
and
\begin{IEEEeqnarray}{r,C,l}
c_k^\T \msgf{V}{X_{k-1}^{(1)}}
& = &  c_k^\T \bigg( \msgf{V}{X_k}^{\text{new}} 
       - \msgf{V}{X_k}^{\text{new}} c_{k+1} 
         \Big(c_{k+1}^\T \msgf{V}{X_k}^{\text{new}} c_{k+1} \Big)^{-1} 
         c_{k+1}^\T \msgf{V}{X_k}^{\text{new}} \bigg)
         \label{eqn:pathFFBDD:SV:point2line:VXtemp:MIL}\\
& = &  c_k^\T \msgf{V}{X_{k-1}^{(0)}}
       -  c_k^\T \Big( c_{k+1}^\T \msgf{V}{X_k}^{\text{new}} \Big)^\T
         \Big(c_{k+1}^\T \msgf{V}{X_k}^{\text{new}} c_{k+1} \Big)^{-1} 
         c_{k+1}^\T \msgf{V}{X_k}^{\text{new}},
         \label{eqn:pathFFBDD:SV:point2line:VXtemp:final}
\end{IEEEeqnarray}
where both (\ref{eqn:pathFFBDD:SV:point2line:VX:MIL}) and (\ref{eqn:pathFFBDD:SV:point2line:VXtemp:MIL}) 
follow from (\ref{eqn:alg:FFBDD:DecompOutput:forw:VX}).
%Note that both (\ref{eqn:pathFFBDD:SV:point2line:VX:final}) and (\ref{eqn:pathFFBDD:SV:point2line:VXtemp:final}) 
%can be computed with complexity $\calO(M)$.
Note that, for $n>k+1$, $c_{n}^\T \msgf{V}{X_{n-1}} = c_{n}^\T \msgf{V}{X_{n-1}}^{\text{old}}$
is not affected by this step.

Likewise, in the second step (from Fig.~\ref{subfig:LS:line2point:moveright1} to Fig.~\ref{subfig:LS:line2point:moveright2}), 
we have both
\begin{IEEEeqnarray}{r,C,l}
c_{k+2}^\T \msgf{V}{X_{k+1}}^{\text{new}}
& = & c_{k+2}^\T \msgf{V}{X_{k-1}^{(1)}} \\
%& = & c_{k+2}^\T \msgf{V}{X_{k+1}}^{\text{old}} 
& = &  c_{k+2}^\T \msgf{V}{X_{k+1}}^{\text{old}}
       + c_{k+2}^\T  \Big( c_k^\T \msgf{V}{X_{k-1}^{(1)}} \Big)^\T
          \Big( c_k^\T \msgf{V}{X_{k-1}^{(1)}} c_k \Big)^{-1}
           c_k^\T \msgf{V}{X_{k-1}^{(1)}}
\end{IEEEeqnarray}
and
\begin{IEEEeqnarray}{r,C,l}
c_k^\T \msgf{V}{X_{k-1}^{(2)}}
& = &  c_k^\T \bigg( \msgf{V}{X_{k+1}}^{\text{new}} 
       - \msgf{V}{X_{k+1}}^{\text{new}} c_{k+2} 
         \Big(c_{k+2}^\T \msgf{V}{X_{k+1}}^{\text{new}} c_{k+2} \Big)^{-1} 
         c_{k+2}^\T \msgf{V}{X_{k+1}}^{\text{new}} \bigg) \\
& = &  c_k^\T \msgf{V}{X_{k-1}^{(1)}}
       - c_k^\T \Big( c_{k+2}^\T \msgf{V}{X_{k+1}}^{\text{new}} \Big)^\T
         \Big(c_{k+2}^\T \msgf{V}{X_{k+1}}^{\text{new}} c_{k+2} \Big)^{-1} 
         c_{k+2}^\T \msgf{V}{X_{k+1}}^{\text{new}},
\end{IEEEeqnarray}
and so on.
\end{proofof}

\clearpage

\section{Numerical Examples}
\label{sec:NumericalExperiments}

As a sanity check, we demonstrate the proposed algorithms by some examples.

%\subsection{Filtering a Global Warming Dataset}
\subsection{Paths for a Low-Dimensional State Space Representation}

We demonstrate PathBFFD and PathFFBDD for Fig.~\ref{fig:StatisticalModel:FactorGraph}
by applications to the global warming dataset \cite{NOAA2023}, 
where the data $\breve{y}_{1},\ldots,\breve{y}_{n} \in \R$ 
are the annual temperature anomalies from 1880 to 2022.

\begin{figure*}
	\centering
	\includegraphics[width=.9\linewidth]{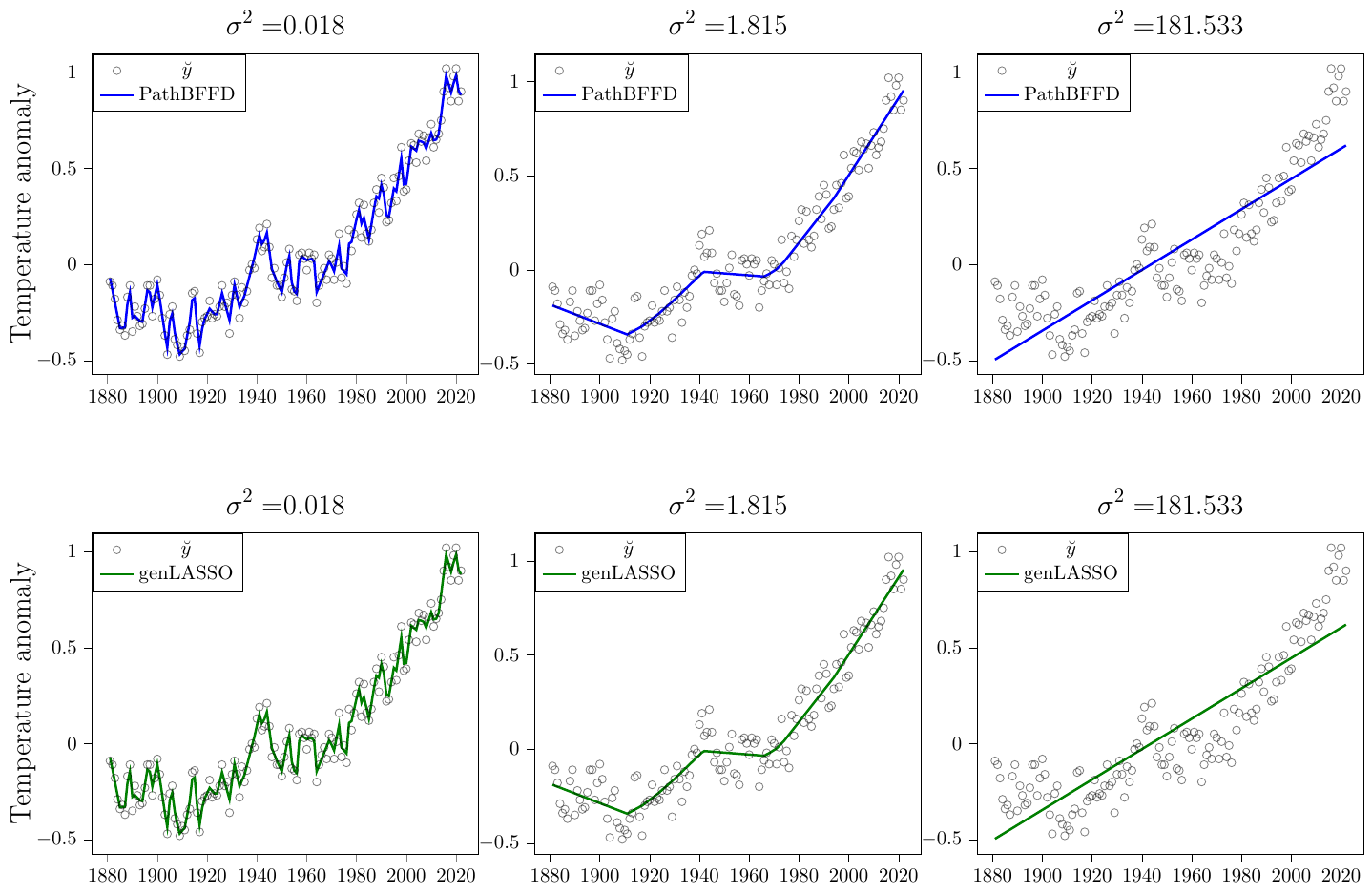} \\
	\caption{\label{fig:TrendFiltering}%
	Example~\ref{ex:pwlm:splinetrend}: 
	Snapshots of trend filtering on the global warming datasets
	computed with PathBFFD (top row) 
	and genLASSO~\cite{Arnold2016} (bottom row).
	The ordinate is $\hat{y}(\sigma^{2})$.
	%The blue shadowed region indicates the interval $\hat{y}_{n}\pm 0.25V_{Y_{n}}^{1/2}$.
	}
\end{figure*}

\begin{example}[Trend Filtering by First-order Spline Smoothing]
\label{ex:pwlm:splinetrend}
We use the model of Example~\ref{ex:pwlm},
i.e., the state space model (\ref{eqn:pwlm})
with $\kappa(u_n) = |u_n|$ and Gaussian $p(\breve y_n \cond y_n)$,
and with $Q_0=Q_N=0$.
Fig.~\ref{fig:TrendFiltering} shows 
some snapshots of the (exact) regularization paths computed by PathBFFD
and (as a sanity check) by genLASSO \cite{Arnold2016}.
%As a sanity check, Fig.~\ref{fig:TrendFiltering} also shows
%the regularization path computed with genLASSO \cite{Arnold2016}.
\end{example}

\begin{figure*}
	\centering
	\includegraphics[width=.9\linewidth]{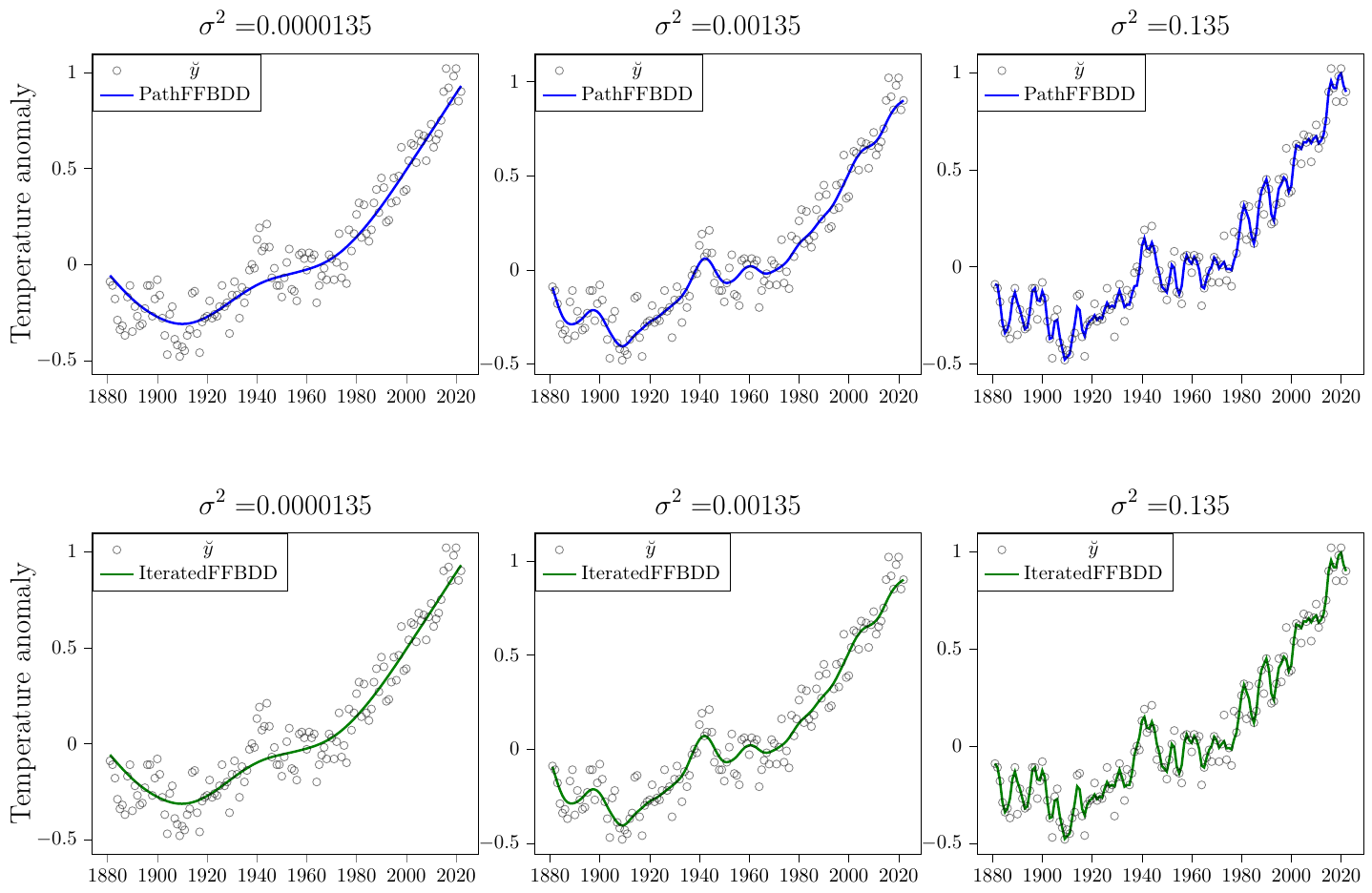} \\
	\caption{\label{fig:MedianFiltering}%
	Example~\ref{ex:pwlm:median}:
	Snapshots of median-Kalman smoothing on the global warming datasets
	computed with PathFFBDD (top row) 
	and the corresponding results from IteratedFFBDD\cite{Li2025} (bottom row).
	The ordinate is $\hat{y}(\sigma^{2})$.
	}
\end{figure*}

\begin{example}[Median-Kalman Smoothing]
\label{ex:pwlm:median}
We use the model of Example~\ref{ex:fomf},
i.e., the state space model (\ref{eqn:pwlm})
with $\kappa(y_n - \breve y_n) = |y_n - \breve y_n|$ and Gaussian $p(u_n)$.
We use $Q_N=0$ and $Q_0 = 10^{-3}$ (and $\breve x_0 = 0$) since PathFFBDD cannot handle $Q_0=0$.
Fig.~\ref{fig:MedianFiltering} shows 
some snapshots of the exact regularization path computed by PathFFBDD,
as well as the results from IteratedFFBDD\cite{Li2025}
(computed for the same discrete values of $\sigma^2$).
\end{example}

%\subsection{Regression on a Diabetes Dataset}
\subsection{Paths for a General Matrix $F$}

We demonstrate PathBFFD for (\ref{eqn:qfittingL1pen})
and PathFFBDD for (\ref{eqn:L1fittingQpen})
with a general matrix $F$ 
using the diabetes dataset \cite{EHJT2004},
where $F\in\R^{442\times10}$ is the design matrix 
and $\breve{y}=(\breve{y}_{1},\ldots,\breve{y}_{442})^{\mathsf{T}}$ 
is the response vector.\\

\begin{figure*}
	\centering
	\includegraphics[width=.95\linewidth]{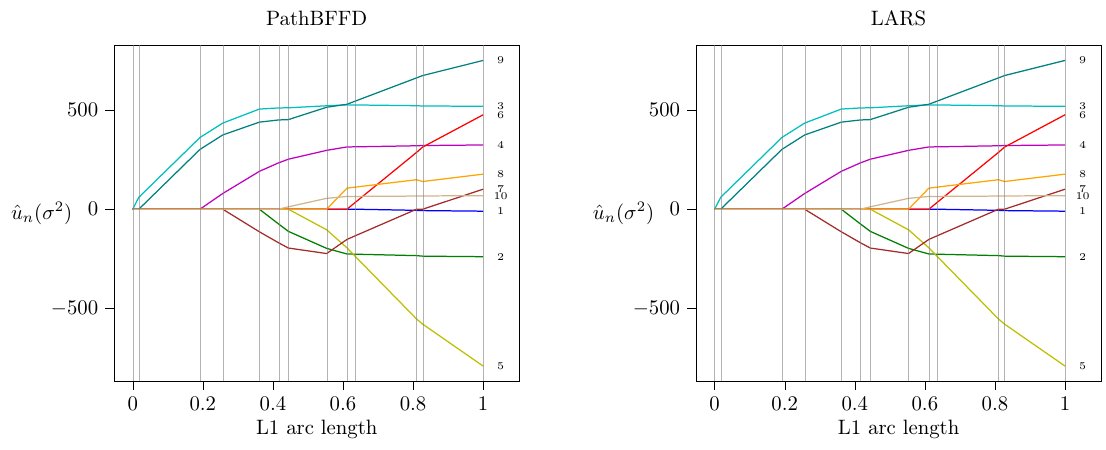} \\
	\caption{\label{fig:L1RegLS}%
	Example~\ref{ex:pwlm:LeastSquaresL1}:
	Regularization path of L1 regularized least squares for the diabetes dataset. 
	The L1 arc length is the normalized sum $\sum_{n=1}^{N}|\hat{u}_{n}(\sigma^{2})|$. 
	The labels on the right side are the indices $n$ of the coefficients $\hat{u}_{n}$.  
	The vertical lines indicate the knots. }
\end{figure*}

\begin{example}[L1 Regularized Least Squares]
\label{ex:pwlm:LeastSquaresL1}
We use the state space model of (\ref{eqn:qfittingL1pen:LSSM:matrixF}) 
and the factor graph of Fig.~\ref{fig:FactorGraph:DecompInput}
with $\kappa(u_n) = |u_n|$ and Gaussian $\breve p(\breve y_n \cond y_n)$.
Fig.~\ref{fig:L1RegLS} shows the regularization paths computed
with PathBFFD (Algorithm~\ref{alg:PathBFFD})
and, for comparison, with LARS \cite{EHJT2004}.
\end{example}

\begin{figure*}
	\centering
	%\subfloat[Small $\sigma^{2}$. Left: exact path provided by symbolic FFBDD; Right: approximate path provided by LIBLINEAR.]{
	\subfloat{
		\includegraphics[width=.95\linewidth]{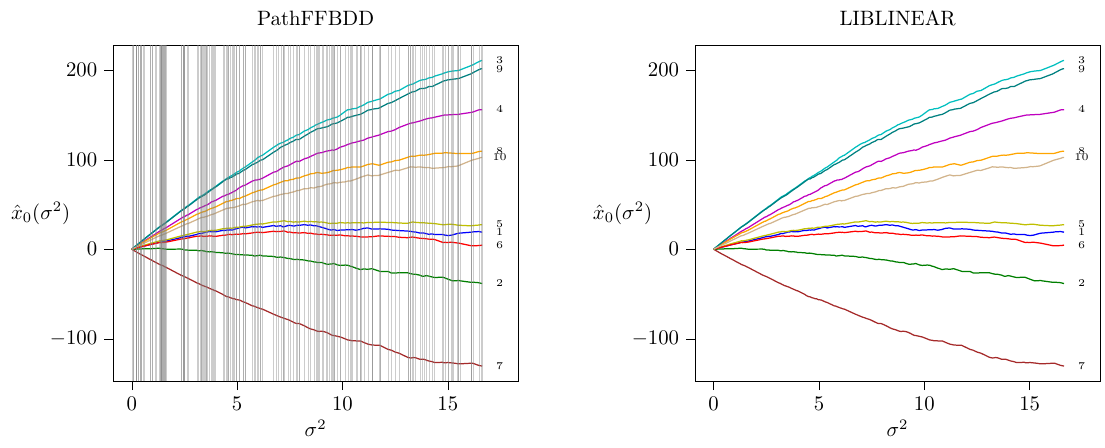}
		%\label{fig:SVRSmallSig2}
	}
	\hfill
	%\subfloat[Large $\sigma^{2}$. Left: exact path provided by symbolic FFBDD; Right: approximate path provided by LIBLINEAR.]{
	\subfloat{
		\includegraphics[width=.95\linewidth]{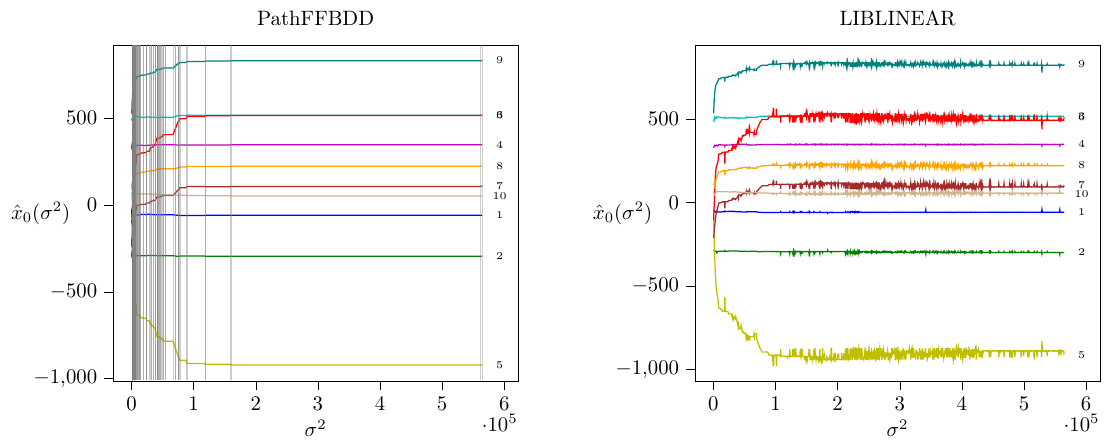}
		%\label{fig:SVRLargeSig2}
	}
	\caption{\label{fig:SVR}%
	Example \ref{ex:pwlm:SVM}:
	Regularized path of linear SVM regression for the diabetes dataset. 
	Top row: moderate values of $\sigma^2$; bottom row: large values of $\sigma^2$.
    The labels on the right side are the indices of the components of $\hat{x}_{0}$.
	The vertical lines indicate the knots.
	}
\end{figure*}

\begin{example}[Linear Support Vector Regression]
\label{ex:pwlm:SVM}
We use the state space model of (\ref{eqn:L1fittingQpen:LSSM:matrixF}) 
and the factor graph of Fig.~\ref{fig:FactorGraph:DecompOutput}
with Gaussian $p(u_n)$ and Vapnik loss 
$\kappa(y_{n}-\breve{y}_{n}) = |y_{n}-\breve{y}_{n}+20|+|y_{n}-\breve{y}_{n}-20|$.
Fig.~\ref{fig:SVR} shows the regularization paths computed
with PathFFBDD (Algorithm~\ref{alg:PathFFBDD})
and, for comparison, the approximate path of LIBLINEAR \cite{Fan2008}
(computed for discrete values of $\sigma^{2}$).
(We used LIBLINEAR since there does not seem to be good and ready prior path algorithm for this example.
The best candidate appears to be svmPath \cite{Hastie2004}, which was designed for classification
and would need to be adapted for regression.)
\end{example}

The time complexity of the algorithms used in these examples
is shown in Table~\ref{tbl:ComputationComplexity}.

\begin{table}%[!tbp]
    \begin{center}
        \begin{tabular}{@{}cccl@{}}
            \toprule
            Experiment & Algorithms & Time complexity & Dominant computations
            \\ 
            \midrule\vspace{1mm}
            \multirowcell{2}{\rule{0em}{2.5ex}Example \ref{ex:pwlm:splinetrend}:\\ Trend filter} 
            & PathBFFD  ($N=L$, $M=2$) & $\mathcal{O}(L)$ 
            & \makecell[l]{matrix-vector multiplication\\ and matrix addition} 
            \\ \cline{3-4}    
            & genLASSO \cite{Arnold2016}  & $\mathcal{O}(L)$  
            & \rule{0em}{2.5ex}sparse QR decomposition
            \\ 
            \midrule
            %\multirowcell{2}{Median-Kalman\\ smoother} 
            \makecell[c]{Example~\ref{ex:pwlm:median}:\\ Median-Kalman smoother} 
            & PathFFBDD ($N=L$, $M=2$) & $\mathcal{O}(L)$ 
            & \makecell[l]{matrix-vector multiplication\\ and matrix addition}
            \\ 
            \midrule
            \multirowcell{2}{\rule{0em}{0ex}Example \ref{ex:pwlm:LeastSquaresL1}:\\ $L_{1}$ regularized\\ least squares} 
            & PathBFFD ($N=K$, $M=L$) & $\mathcal{O}(L K)$ 
            & \makecell[l]{matrix-vector multiplication\\ and matrix addition} 
            \\ \cline{3-4} 
            & LARs \cite{EHJT2004} ($r\leq K$) & $\mathcal{O}(r^{2}+LK)$
            & \makecell[l]{\rule{0em}{2.5ex}update/downdate\\ Cholesky factorization}
            \\
            \midrule
            \multirowcell{2}{\rule{0em}{5ex}Example \ref{ex:pwlm:SVM}:\\ Linear SVM}
            & PathFFBDD ($N=L$, $M=K$)  & $\mathcal{O}(LK)$ 
            & \makecell[l]{matrix-vector multiplication\\ and matrix addition}  
            \\ \cline{3-4} 
            & \makecell[l]{svmPath \cite{Hastie2004} ($r\leq L$)} 
            & \makecell[l]{$\mathcal{O}(r^{2}+rL)$\\ or $\mathcal{O}(r^{3}+rL)$} &  
            \makecell[l]{\rule{0em}{2.5ex}update/downdate partitioned\\ matrix inversion or solving \\ linear equations}
            \\
            \bottomrule
        \end{tabular}
    \end{center}
    \vspace{1ex}
    \caption{\label{tbl:ComputationComplexity}%
    The time complexity per iteration of the path algorithms used in the numerical examples
    (disregarding the complexity of the precomputations).
    The quantity $r$ denotes the size of the active set in prior-art algorithms.
    (svmPath was not actually used, cf.\ the explanation in Example~\ref{ex:pwlm:SVM}.)}
\end{table}

\clearpage

\section{Conclusions}
\label{sec:Conclusions}

We proposed a new algorithm (PathBFFD) for computing the path of (\ref{eqn:qfittingL1pen})
and a dual algorithm (PathFFBDD) for computing 
the path of (\ref{eqn:L1fittingQpen}).
Both algorithms work not only for L1 regularization, 
but for general continuous piecewise linear convex regularizing cost functions.
%The essential novelty is the state space formulation of the problem
%and the use of parametric Gaussian message passing in the corresponding factor graphs
%using the tables in Appendix~\ref{appsec:GaussianMsgTables}.
The proposed algorithms are especially attractive if the matrix $F$ 
in (\ref{eqn:qfittingL1pen}) or (\ref{eqn:L1fittingQpen})
admits a low-dimensional state space representation.
For generic matrices $F$,
we proposed refined versions of these algorithms
with a complexity not larger than that of specialized prior algorithms.

The essential novelty of the proposed algorithms 
is the state space formulation of the problem
and the use of parametric Gaussian message passing in the corresponding factor graphs
using the tables in Appendix~\ref{appsec:GaussianMsgTables}.
Handling linear non-Gaussian models by parameteric Gaussian message passing 
appears to be a new idea. 
As the proposed algorithms as based on Kalman-type recursions, 
it is remarkable that the complexity per iteration of
(the refined versions of) the proposed algorithms for generic $F$
is only linear in the dimensions of~$F$.

\clearpage

%\appendix
\appendices
%\section*{Appendices}

\setcounter{table}{0}
\renewcommand{\thetable}{\thesection.\arabic{table}}

\setcounter{equation}{0}
\renewcommand{\theequation}{\thesection.\arabic{equation}}

\section{Tables for Gaussian Message Passing in Linear Models}
\label{appsec:GaussianMsgTables}

Tables \ref{tbl:VectorGaussianMessageRules}--\ref{tbl:SingleEdgeGMP}
are reproduced from \cite[Appendix~A]{Loeliger2016}
and  improve on the tables in \cite{LDHKLK2007}.
The parameters $V,W,m,\xi$ refer to (\ref{eqn:GenGaussianMean}) and (\ref{eqn:GenGaussian}).
The matrices $W$ and $V$ are positive semi-definite, and $V=W^{-1}$ if they are positive definite.
Quantities with an arrow (e.g., $\msgf{m}{Z_1}$) refer to messages while 
quantities without an arrow (e.g., $m_{Z_1}$) refer to posteriors.
Quantities with a tilde (e.g., the dual mean $\tilde \xi_{Z_1}$) 
are defined in Table~\ref{tbl:SingleEdgeGMP}.
\vspace{2ex}\visual

\begin{table}[bh]
	%\caption{\label{tbl:VectorGaussianMessageRules}%
	%        Gaussian message passing through elementary linear constraints.} 
	\setlength{\unitlength}{1mm}
	\renewcommand{\arraystretch}{1.2}
	\begin{center}
		\begin{tabular}{|c|l|}
			\hline
			\begin{minipage}{3cm}
				\centering
				\begin{tikzpicture} 
					\node[] (X) at (-1.0,0.0) {};
					\node [draw=black,minimum width=0.4cm,minimum height=0.4cm]  (A) at (0.0,0.0) {$=$};
					\node[] (Y) at (0.0,-1.0) {};
					\node[] (Z) at (1.0,0.0) {};
					\draw [-latex,color=black] (X.east) -- (A.west)  node[midway,above] {$Z_{1}$};
					\draw [-latex,color=black] (Y.north) -- (A.south)  node[midway,right] {$Z_{2}$};
					\draw [latex-,color=black] (Z.west) -- (A.east)  node[midway,above] {$Z_{3}$};
					\node[] (txt) at (0.1,-1.5) {$Z_{1}=Z_{2}=Z_{3}$};
				\end{tikzpicture}
			\end{minipage}
			&
			\begin{minipage}{5cm}
				\begin{IEEEeqnarray}{rCl}
					\msgf{W}{Z_{3}} &=& \msgf{W}{Z_{1}} + \msgf{W}{Z_{2}}\\
					\msgf{\xi}{Z_{3}} &=&\msgf{\xi}{Z_{1}} + \msgf{\xi}{Z_{2}}
				\end{IEEEeqnarray}
				\vspace{-2ex}
				\hrule
				\begin{IEEEeqnarray}{rCl}
					\msgb{W}{Z_{1}} &=& \msgf{W}{Z_{2}} + \msgb{W}{Z_{3}}\\
					\msgb{\xi}{Z_{1}} &=& \msgf{\xi}{Z_{2}} + \msgb{\xi}{Z_{3}}
				\end{IEEEeqnarray}
				\vspace{-2ex}
				\hrule
				\begin{IEEEeqnarray}{rCl}
					m_{Z_{1}} &=& m_{Z_{2}} = m_{Z_{3}}\label{eqn:EquationPrimalMarginal}\\
					V_{Z_{1}} &=& V_{Z_{2}} =V_{Z_{3}}
				\end{IEEEeqnarray}
				\vspace{-2ex}
				\hrule
				\begin{IEEEeqnarray}{rCl}
					\tilde{\xi}_{Z_{3}}&=&	\tilde{\xi}_{Z_{1}} + 		\tilde{\xi}_{Z_{2}}
				\end{IEEEeqnarray}
				\vspace{-2ex}
			\end{minipage}
			\\ \hline 
			\begin{minipage}{3cm}
				\centering
				\begin{tikzpicture} 
					\node[] (X) at (-1.0,0.0) {};
					\node [draw=black,minimum width=0.4cm,minimum height=0.4cm]  (A) at (0.0,0.0) {$+$};
					\node[] (Y) at (0.0,-1.0) {};
					\node[] (Z) at (1.0,0.0) {};
					\draw [-latex,color=black] (X.east) -- (A.west)  node[midway,above] {$Z_{1}$};
					\draw [-latex,color=black] (Y.north) -- (A.south)  node[midway,right] {$Z_{2}$};
					\draw [latex-,color=black] (Z.west) -- (A.east)  node[midway,above] {$Z_{3}$};
					\node[] (txt) at (0.1,-1.5) {$Z_{1}+Z_{2}=Z_{3}$};
					
				\end{tikzpicture}
			\end{minipage}
			&
			\begin{minipage}{5cm}
				\begin{IEEEeqnarray}{rCl}
					\msgf{V}{Z_{3}} &=& \msgf{V}{Z_{1}} + \msgf{V}{Z_{2}}\\
					\msgf{m}{Z_{3}} &=&\msgf{m}{Z_{1}} + \msgf{m}{Z_{2}}
				\end{IEEEeqnarray}
				\vspace{-2ex}
				\hrule
				\begin{IEEEeqnarray}{rCl}
					\msgb{V}{Z_{1}} &=& \msgf{V}{Z_{2}} + \msgb{V}{Z_{3}}\\
					\msgb{m}{Z_{1}} &=&  \msgb{m}{Z_{3}} -\msgf{m}{Z_{2}}
				\end{IEEEeqnarray}
				\vspace{-2ex}
				\hrule
				\begin{IEEEeqnarray}{rCl}
					m_{Z_{3}} &=& m_{Z_{1}}  + m_{Z_{2}}
				\end{IEEEeqnarray}
				\vspace{-2ex}
				\hrule
				\begin{IEEEeqnarray}{rCl}
					\tilde{\xi}_{Z_{1}} &=& \tilde{\xi}_{Z_{2}} = \tilde{\xi}_{Z_{3}}\label{eqn:AdditionDualMarginal}\\
					\tilde{W}_{Z_{1}} &=& \tilde{W}_{Z_{2}} =\tilde{W}_{Z_{3}}
				\end{IEEEeqnarray}
				\vspace{-2ex}
			\end{minipage}
			\\ \hline
			\begin{minipage}{3cm}
				\centering
				\begin{tikzpicture} [scale=1.]
					\node[] (X) at (-1.0,0.0) {};
					\node [draw=black,minimum width=0.4cm,minimum height=0.4cm]  (A) at (0.0,0.0) {$A$};
					\node[] (Y) at (1.0,0.0) {};
					%\node[] (Z) at (0.0,-1.0) {};
					\draw [-latex,color=black] (X.east) -- (A.west)  node[midway,above] {$Z_{1}$};
					\draw [latex-,color=black] (Y.west) -- (A.east)  node[midway,above] {$Z_{2}$};
					%\node[] (txt) at (0.0,-1.0) {$AZ_{1}=Z_{2}$};
					\node[] (txt) at (0.0,-1.0) {$Z_2 = A Z_1$};
				\end{tikzpicture}
				
			\end{minipage}
			&
			\begin{minipage}{5cm}
				\begin{IEEEeqnarray}{rCl}
					\msgf{V}{Z_{2}} &=& A\msgf{V}{Z_{1}}A^{\mathsf{T}}\\
					\msgf{m}{Z_{2}}&=& A\msgf{m}{Z_{1}}
				\end{IEEEeqnarray}
				\vspace{-2ex}
				\hrule
				\begin{IEEEeqnarray}{rCl}
					\msgb{W}{Z_{1}} &=& A^{\mathsf{T}}\msgb{W}{Z_{2}}A\\
					\msgb{\xi}{Z_{1}} &=& A^{\mathsf{T}}\msgb{\xi}{Z_{2}}
				\end{IEEEeqnarray}
				\vspace{-2ex}
				\hrule
				\begin{IEEEeqnarray}{rCl}
					m_{Z_{2}} &=& A m_{Z_{1}} \label{eqn:LinearTransformeqn:ForwardPrimalMarginal}\\
					V_{Z_{2}} &=& AV_{Z_{2}}A^{\mathsf{T}}
				\end{IEEEeqnarray}
				\vspace{-2ex}
				\hrule
				\begin{IEEEeqnarray}{rCl}
					\tilde{\xi}_{Z_{1}} &=& A^{\mathsf{T} }\tilde{\xi}_{Z_{2}} \label{eqn:LinearTransformeqn:BackwardDualMarginal}\\
					\tilde{W}_{Z_{1}} &=& A^{\mathsf{T} } 	\tilde{W}_{Z_{2}} A
				\end{IEEEeqnarray}
				\vspace{-2ex}
			\end{minipage}
			\\ \hline
		\end{tabular}
	\end{center}
	\vspace{1.5ex}
	\caption{\label{tbl:VectorGaussianMessageRules}%
		Gaussian message passing through elementary linear constraints.}
\end{table}

\begin{table}
	%\caption{\label{tbl:VectorGaussianMessageRulesCompositeBlocks}%
	%	Gaussian message passing through composite linear blocks.} 
	\setlength{\unitlength}{1mm}
	\renewcommand{\arraystretch}{1.2}
	\begin{center}
		\begin{tabular}{|c|l|}
			\hline
			\begin{minipage}{3cm}
				\centering
				\begin{tikzpicture} 
					\node[] (X) at (-1.0,0.0) {};
					\node [draw=black,minimum width=0.4cm,minimum height=0.4cm]  (A) at (0.0,0.0) {$=$};
					\node [draw=black,minimum width=0.4cm,minimum height=0.4cm]  (B) at (0.0,-1.0) {$A$};
					\node[] (Y) at (0.0,-2.0) {};
					\node[] (Z) at (1.0,0.0) {};
					\draw [-latex,color=black] (X.east) -- (A.west)  node[midway,above] {$Z_{1}$};
					\draw [latex-,color=black] (B.north) -- (A.south)  node[midway,right] {};
					\draw [latex-,color=black] (Y.north) -- (B.south)  node[midway,right] {$Z_{2}$};
					\draw [latex-,color=black] (Z.west) -- (A.east)  node[midway,above] {$Z_{3}$};
					%\node[] (txt) at (0.1,-2.5) {\text{composite equation}};
				\end{tikzpicture}

			\end{minipage}
			&
			%\begin{minipage}{5cm}
			\begin{minipage}{6cm}
				\begin{IEEEeqnarray}{rCl}
					G & = & \big(\msgb{V}{Z_{2}}+A\msgf{V}{Z_{1}}A^{\mathsf{T}}\big)^{-1} \label{eqn:CompositeEquationMatrixG}\\
					g &=& A^{\mathsf{T}}G\big(\msgb{m}{Z_{2}}-A\msgf{m}{Z_{1}}\big)
					\label{eqn:CompositeEquationVectorg}\\
					\msgf{m}{Z_{3}} &=&\msgf{m}{Z_{1}} + \msgf{V}{Z_{1}}g
					\label{eqn:CompositeEquationForwardMean}\\
					\msgf{V}{Z_{3}} &=&\msgf{V}{Z_{1}} - \msgf{V}{Z_{1}}A^{\mathsf{T}}GA \msgf{V}{Z_{1}}
					  \IEEEeqnarraynumspace
					  \label{eqn:CompositeEquationForwardV}
				\end{IEEEeqnarray}
				\vspace{-2ex}
				\hrule
				\begin{IEEEeqnarray}{rCl}
					F &=& I - 	\msgf{V}{Z_{1}} A^{\mathsf{T}}GA \\
					\tilde{\xi}_{Z_{1}} & = & F^{\mathsf{T}}\tilde{\xi}_{Z_{3}} - g\\
					\tilde{W}_{Z_{1}} & = & F^{\mathsf{T}}\tilde{W}_{Z_{3}}F +A^{\mathsf{T}}GA \label{eqn:CompositeEquationDualCovariance}
					   \IEEEeqnarraynumspace
				\end{IEEEeqnarray}
				\vspace{-2ex}
			\end{minipage}
			\\ \hline 
			\begin{minipage}{3cm}
				\centering
				\begin{tikzpicture} 
					\node[] (X) at (-1.0,0.0) {};
					\node [draw=black,minimum width=0.4cm,minimum height=0.4cm]  (A) at (0.0,0.0) {$+$};
					\node [draw=black,minimum width=0.4cm,minimum height=0.4cm]  (B) at (0.0,-1.0) {$A$};
					\node[] (Y) at (0.0,-2.0) {};
					\node[] (Z) at (1.0,0.0) {};
					\draw [-latex,color=black] (X.east) -- (A.west)  node[midway,above] {$Z_{1}$};
					\draw [-latex,color=black] (B.north) -- (A.south)  node[midway,right] {};
					\draw [-latex,color=black] (Y.north) -- (B.south)  node[midway,right] {$Z_{2}$};
					\draw [latex-,color=black] (Z.west) -- (A.east)  node[midway,above] {$Z_{3}$};
					%\node[] (txt) at (0.1,-2.5) {\text{composite addition}};
				\end{tikzpicture}
			\end{minipage}
			&
			\begin{minipage}{6cm}
				\begin{IEEEeqnarray}{rCl}
					H & = & \big(\msgf{W}{Z_{2}} + A^{\mathsf{T}}\msgb{W}{Z_{3}}A\big)^{-1}
					\label{eqn:CompositeAdditionMatrixH}\\
					h &=& AH\big(\msgf{\xi}{Z_{2}}+A^{\mathsf{T}}\msgb{\xi}{Z_{3}}\big)
					\label{eqn:CompositeAdditionVectorh}\\
					\msgb{\xi}{Z_{1}} & = & \msgb{\xi}{Z_{3}} -  \msgb{W}{Z_{3}}h
					\label{eqn:CompositeAdditionBackwardPrecisionMean}\\
					\msgb{W}{Z_{1}} &=&\msgb{W}{Z_{3}}- \msgb{W}{Z_{3}} AHA^{\mathsf{T}}\msgb{W}{Z_{3}} 
					   \label{eqn:CompositeAdditionBackwardPrecision} \IEEEeqnarraynumspace
				\end{IEEEeqnarray}
				\vspace{-2ex}
				\hrule
				\begin{IEEEeqnarray}{rCl}
					\tilde{F} &=& I - \msgb{W}{Z_{3}}AH A^{\mathsf{T}}\\
					m_{Z_{3}} & = & 	\tilde{F}^{\mathsf{T}}m_{Z_{1}} + h\\
					V_{Z_{3}}  &=&  \tilde{F}^{\mathsf{T}}V_{Z_{1}}\tilde{F} + AHA^{\mathsf{T}} 	\label{eqn:CompositeAdditionCovariance}
					   \IEEEeqnarraynumspace
				\end{IEEEeqnarray}
				\vspace{-2ex}
			\end{minipage}
			\\ \hline
		\end{tabular}
	\end{center}
	\vspace{1.5ex}
	\caption{\label{tbl:VectorGaussianMessageRulesCompositeBlocks}%
		Gaussian message passing through composite linear blocks.}
%\end{table}
\vspace{1ex}

%\begin{table}[h]
\newcommand{\dW}{\tilde W}
\begin{center}
\framebox[0.55\linewidth]{%
\begin{minipage}{0.5\linewidth}

\vspace{-1.5ex}
\begin{IEEEeqnarray}{rCl}
\tilde\xi_Z & \eqdef & \dW_Z (\msgf{m}{Z} - \msgb{m}{Z})\label{eqn:TabSingleEdgeGMPdxiDef} \\
       & = & \msgf{\xi}{Z} - \msgf{W}{Z} m_Z \label{eqn:TabSingleEdgeGMPdxi}\\
       & = &  \msgb{W}{Z} m_Z - \msgb{\xi}{Z} \label{eqn:TabSingleEdgeGMPdxi2}\\[1ex]
\dW_Z & \eqdef & (\msgf{V}{Z} + \msgb{V}{Z})^{-1} \label{eqn:TabSingleEdgeGMPdWDef}\\
  & = & \msgf{W}{Z} V_Z \msgb{W}{Z} \label{eqn:TabSingleEdgeGMPdW1}\\
  & = & \msgf{W}{Z} - \msgf{W}{Z} V_Z \msgf{W}{Z} \label{eqn:TabSingleEdgeGMPdW} \\
  & = & \msgb{W}{Z} - \msgb{W}{Z} V_Z \msgb{W}{Z} \label{eqn:TabSingleEdgeGMPdW2}
  \IEEEeqnarraynumspace\\[1ex]
  \hline\nonumber\\%[-2ex]
%\rule{0em}{3.5ex}%
m_Z & = & V_Z (\msgf{\xi}{Z} + \msgb{\xi}{Z}) \label{eqn:TabSingleEdgeGMPmDef}\\
    & = & \msgf{m}{Z} - \msgf{V}{Z} \tilde\xi_Z  \label{eqn:TabSingleEdgeGMPm}\\
    & = & \msgb{m}{Z} + \msgb{V}{Z} \tilde\xi_Z \label{eqn:TabSingleEdgeGMPm2}\\[1ex]
V_Z & = & (\msgf{W}{Z} + \msgb{W}{Z})^{-1} \label{eqn:TabSingleEdgeGMPVarDef}\\
  & = & \msgf{V}{Z} \dW_Z \msgb{V}{Z} \label{eqn:TabSingleEdgeGMPV1}\\
  & = & \msgf{V}{Z} - \msgf{V}{Z} \dW_Z \msgf{V}{Z} \label{eqn:TabSingleEdgeGMPV2} \\
  & = & \msgb{V}{Z} - \msgb{V}{Z} \dW_Z \msgb{V}{Z} \label{eqn:TabSingleEdgeGMPV3}
  \IEEEeqnarraynumspace
\end{IEEEeqnarray}
\vspace{-3.5ex}
\end{minipage}%
}%framebox
\end{center}
\vspace{1.5ex}
\caption{\label{tbl:SingleEdgeGMP}%
%Single-edge quantities ($m$, $V$) and their duals ($\tilde\xi$, $\dW$).}
Posteriors ($m$, $V$) and their duals ($\tilde\xi$, $\dW$).}
\end{table}

\clearpage

\newcommand{\itemhead}[1]{\medskip\noindent {\bf #1: }}

\section{Details of Parametric BFFD and Proof of Proposition~\ref{sec:BFFD:Parametric}}
\label{appsec:ParametricBFFD}

We give the details of, and thereby verify, the parameteric representation 
of the pertinent quantities 
for each step in Algorithms \ref{alg:BFFD:SSM} and~\ref{alg:BFFD:DecompInput:SpecUpd}.
The proof of Proposition~\ref{sec:BFFD:Parametric} 
then follows by induction, using these representations throughout the algorithm.

\itemhead{(\ref{eqn:alg:BFFD:backw:xiXpp}) and (\ref{eqn:alg:BFFD:backw:WXpp})}
If $\msgb{\xi}{X_n} = \sigma^{-2}\msgb{\xi}{X_n}^{(1)} + \msgb{\xi}{X_n}^{(0)}$ 
and $\msgb{\xi}{Y_n} = \sigma^{-2}\msgb{\xi}{Y_n}^{(1)} + \msgb{\xi}{Y_n}^{(0)}$,
then 
\begin{equation} 
\msgb{\xi}{X_n''} 
   = \sigma^{-2}\mleft( \msgb{\xi}{X_n}^{(1)} + c_n \msgb{\xi}{Y_n}^{(1)} \mright) 
     + \mleft( \msgb{\xi}{X_n}^{(0)} + c_n \msgb{\xi}{Y_n}^{(0)} \mright).
\end{equation}
Likewise, if $\msgb{W}{X_n} = \sigma^{-2}\msgb{W}{X_n}^{(1)}$ 
and $\msgb{W}{Y_n} = \sigma^{-2}\msgb{W}{Y_n}^{(1)}$,
then
\begin{equation}
\msgb{W}{X_n''} 
   = \sigma^{-2}\mleft( \msgb{W}{X_n}^{(1)} + c_n \msgb{W}{Y_n}^{(1)} c_n^\T \mright).
\end{equation}

\itemhead{(\ref{eqn:alg:BFFD:backw:H}) and (\ref{eqn:alg:BFFD:backw:h})}
If $\msgb{\xi}{X_n''} = \sigma^{-2}\msgb{\xi}{X_n''}^{(1)} + \msgb{\xi}{X_n''}^{(0)}$ and 
$\msgb{W}{X_n''} = \sigma^{-2}\msgb{W}{X_n''}^{(1)}$, then 
\begin{equation}
	H_{n}=\begin{cases}
		0, &  \text{if $\msgf{V}{U_n}=0$}\\
		\sigma^{2}	  \mleft(b_{n}^{\mathsf{T}}\msgb{W}{X_n''}^{(1)}b_{n}\mright)^{-1},  &  \text{if $\msgf{W}{U_n}=0$}\\
	\end{cases}
\end{equation} 
and 
\begin{equation}
	h_{n}=\begin{cases}
		b_{n}\msgf{m}{U_{n}}, &  \text{if $\msgf{V}{U_n}=0$}\\
		\sigma^{2} \bigg(b_{n} \mleft(b_{n}^{\mathsf{T}}\msgb{W}{X_n''}^{(1)}b_{n}\mright)^{-1}\mleft(\msgf{\xi}{U_{n}}+b_{n}^{\mathsf{T}}\msgb{\xi}{X_n''}^{(0)}\mright)\bigg) + \bigg(b_{n} \mleft(b_{n}^{\mathsf{T}}\msgb{W}{X_n''}^{(1)}b_{n}\mright)^{-1}b_{n}^{\mathsf{T}}\msgb{\xi}{X_n''}^{(1)}\bigg),  &  \text{if $\msgf{W}{U_n}=0$}\\
	\end{cases}
\end{equation} 
where $\msgf{W}{U_{n}}=0$ and $\msgf{\xi}{U_{n}}$ does not depend on $\sigma$, or else $\msgf{V}{U_{n}}=0$ and $\msgf{m}{U_{n}}$ does not depend on $\sigma$.

\itemhead{(\ref{eqn:alg:BFFD:backw:xiXppp}) and (\ref{eqn:alg:BFFD:backw:WXppp})}
If $H_{n}=\sigma^{2}H_{n}^{(1)}$, $h_{n}=\sigma^{2}h_{n}^{(1)}+h_{n}^{(0)}$, $\msgb{\xi}{X_n''} = \sigma^{-2}\msgb{\xi}{X_n''}^{(1)} + \msgb{\xi}{X_n''}^{(0)}$ and 
$\msgb{W}{X_n''} = \sigma^{-2}\msgb{W}{X_n''}^{(1)}$, then
\begin{equation}
	\msgb{\xi}{X_n'''} = \sigma^{-2}\mleft(\msgb{\xi}{X_n''}^{(1)} -\msgb{W}{X_n''}^{(1)}h_{n}^{(0)} \mright) + \mleft(\msgb{\xi}{X_n''}^{(0)} -\msgb{W}{X_n''}^{(1)}h_{n}^{(1)} \mright),
\end{equation}
and 
\begin{equation}
		\msgb{W}{X_n'''}  = \sigma^{-2}\mleft( \msgb{W}{X_n''}^{(1)} - \msgb{W}{X_n''}^{(1)}b_{n}H_{n}^{(1)}b_{n}^{\mathsf{T}}\msgb{W}{X_n''}^{(1)} \mright).
\end{equation}

\itemhead{(\ref{eqn:alg:BFFD:backw:xiX}) and (\ref{eqn:alg:BFFD:backw:WX})}
If $\msgb{\xi}{X_n'''} = \sigma^{-2}\msgb{\xi}{X_n'''}^{(1)} + \msgb{\xi}{X_n'''}^{(0)}$ and 
$\msgb{W}{X_n'''} = \sigma^{-2}\msgb{W}{X_n'''}^{(1)}$, then
\begin{equation}
	\msgb{\xi}{X_{n-1}} = \sigma^{-2}\mleft(A^{\mathsf{T}}\msgb{\xi}{X_n'''}^{(1)} \mright)+ \mleft(A^{\mathsf{T}}\msgb{\xi}{X_n'''}^{(0)}\mright)
\end{equation}
and 
\begin{equation}
	\msgb{W}{X_{n-1}} = \sigma^{-2} \mleft( A^{\mathsf{T}} \msgb{W}{X_n'''}^{(1)}A \mright).
\end{equation}

\itemhead{(\ref{eqn:PrimalMarginalEstimateInitialState})} If $\msgf{\xi}{X_0} = \sigma^{-2}\msgf{\xi}{X_0}^{(1)} + \msgf{\xi}{X_0}^{(0)}$,  
$\msgf{W}{X_0} = \sigma^{-2}\msgf{W}{X_0}^{(1)}$, $\msgb{\xi}{X_0} = \sigma^{-2}\msgb{\xi}{X_0}^{(1)} + \msgb{\xi}{X_0}^{(0)}$, and
$\msgb{W}{X_0} = \sigma^{-2}\msgb{W}{X_0}^{(1)}$, then
\begin{equation}
	\hat{x}_{0}=\sigma^{2} \bigg( \mleft(\msgf{W}{X_0}^{(1)} + \msgb{W}{X_0}^{(1)}  \mright) ^{-1}\mleft( \msgf{\xi}{X_0}^{(0)} + \msgb{\xi}{X_0}^{(0)} \mright) \bigg) + \bigg( \mleft(\msgf{W}{X_0}^{(1)} + \msgb{W}{X_0}^{(1)}  \mright) ^{-1}\mleft( \msgf{\xi}{X_0}^{(1)} + \msgb{\xi}{X_0}^{(1)} \mright) \bigg).
\end{equation}

\itemhead{(\ref{eqn:ForwardRecursionBFFD})} If $\hat{x}_{n-1}=\sigma^{2}\hat{x}_{n-1}^{(1)} + \hat{x}_{n-1}^{(0)} $, then
\begin{equation}
	\hat{x}_n''' = \sigma^{2}\mleft(A\hat{x}_{n-1}^{(1)}\mright) + \mleft( A \hat{x}_{n-1}^{(0)} \mright).
\end{equation}

\itemhead{(\ref{eqn:alg:BFFD:fwd:msgbVU}) and (\ref{eqn:alg:BFFD:fwd:msgbU})} If $\msgb{\xi}{X_n''} = \sigma^{-2}\msgb{\xi}{X_n''}^{(1)} + \msgb{\xi}{X_n''}^{(0)}$,  
$\msgb{W}{X_n''} = \sigma^{-2}\msgb{W}{X_n''}^{(1)}$ and $	\hat{x}_n''' = \sigma^{2}	 \hat{x}_n'''^{(1)} + \hat{x}_n'''^{(0)} $, then
\begin{equation}
	\msgb{V}{U_{n}} = \sigma^{2}\mleft(b_{n}^{\mathsf{T}}\msgb{W}{X_n''}^{(1)}b_{n}\mright)^{-1},
\end{equation}
and
\begin{equation}
	\msgb{m}{U_{n}} = \sigma^{2}\bigg(\mleft(b_{n}^{\mathsf{T}}\msgb{W}{X_n''}^{(1)}b_{n}\mright)^{-1}\mleft(b_{n}^{\mathsf{T}}\msgb{\xi}{X_n''}^{(0)} - b_{n}^{\mathsf{T}}\msgb{W}{X_n''}^{(1)} \hat{x}_n'''^{(1)} \mright)\bigg)
	+ \bigg(\mleft(b_{n}^{\mathsf{T}}\msgb{W}{X_n''}^{(1)}b_{n}\mright)^{-1}\mleft(b_{n}^{\mathsf{T}}\msgb{\xi}{X_n''}^{(1)} - b_{n}^{\mathsf{T}}\msgb{W}{X_n''}^{(1)} \hat{x}_n'''^{(0)} \mright)\bigg).
\end{equation}

\itemhead{(\ref{eqn:alg:BFFD:fwd:hatU})}  If $\msgb{V}{U_{n}}=\sigma^{2}\msgb{V}{U_{n}}^{(1)}$ and $\msgb{m}{U_{n}}=\sigma^{2}\msgb{m}{U_{n}}^{(1)} + \msgb{m}{U_{n}}^{(0)}$, then 
\begin{equation}
	\hat{u}_{n} = \begin{cases}
		\msgf{m}{U_{n}}, & \text{if $\msgf{V}{U_n}=0$}\\
		\sigma^{2}\mleft( \msgb{m}{U_{n}}^{(1)} + \msgb{V}{U_{n}}^{(1)}\msgf{\xi}{U_{n}} \mright) + \mleft(
		 \msgb{m}{U_{n}}^{(0)}\mright) & \text{if $\msgf{W}{U_n}=0$}\\
	\end{cases}
\end{equation}
where $\msgf{W}{U_{n}}=0$ and $\msgf{\xi}{U_{n}}$ does not depend on $\sigma$, or else $\msgf{V}{U_{n}}=0$ and $\msgf{m}{U_{n}}$ does not depend on $\sigma$.

\itemhead{(\ref{eqn:alg:BFFD:fwd:hatX}) and (\ref{eqn:alg:BFFD:fwd:hatY})} If 	$\hat{x}_n''' = \sigma^{2}	  \hat{x}_n'''^{(1)} +\hat{x}_n'''^{(0)} $ and $\hat{u}_{n}=\sigma^{2}\hat{u}_{n}^{(1)} + \hat{u}_{n}^{(0)}$, then
\begin{equation}
		\hat{x}_{n} = \sigma^{2}\mleft( \hat{x}_n'''^{(1)} + b_{n}\hat{u}_{n}^{(1)}  \mright) + \mleft( \hat{x}_n'''^{(0)} + b_{n}\hat{u}_{n}^{(0)}  \mright)
\end{equation}
and
\begin{equation}
	\hat{y}_{n} = \sigma^{2}\bigg( c_{n}^{\mathsf{T}}\mleft(  \hat{x}_n'''^{(1)} + b_{n}\hat{u}_{n}^{(1)} \mright)
	\bigg) + \bigg( c_{n}^{\mathsf{T}}\mleft(  \hat{x}_n'''^{(0)} + b_{n}\hat{u}_{n}^{(0)} \mright)
	\bigg) .
\end{equation}
\bigskip

\section{Details of Parametric FFBDD and Proof of Proposition~\ref{prop:FFBDD:affine}}
\label{appsec:ParametricFFBDD}

We give the details of, and thereby verify, the parameteric representation 
of the pertinent quantities 
for each step in Algorithms \ref{alg:FFBDD:SSM} and~\ref{alg:FFBDD:DecompOutput}.
The proof of Proposition~\ref{sec:FFBDD:Parametric}
then follows by induction, using these representations throughout the algorithm.

\itemhead{(\ref{eqn:alg:FFBDD:forw:mXppp}) and (\ref{eqn:alg:FFBDD:forw:VXppp})}
If $\msgf{m}{X_{n-1}} = \sigma^{2}\msgf{m}{X_{n-1}}^{(1)} +\msgf{m}{X_{n-1}}^{(0)}$ 
and $\msgf{V}{X_{n-1}} = \sigma^{2}\msgf{V}{X_{n-1}} ^{(1)}$,
then 
\begin{equation}
	\msgf{m}{X_{n}^{'''}} =  \sigma^{2}\mleft(A\msgf{m}{X_{n-1}}^{(1)}\mright) + \mleft(A\msgf{m}{X_{n-1}}^{(0)} \mright)
\end{equation}
and
\begin{equation}
	\msgf{V}{X_{n}^{'''}} =  \sigma^{2}\mleft(A\msgf{V}{X_{n-1}}^{(1)}A\mright). 
\end{equation}

\itemhead{(\ref{eqn:alg:FFBDD:forw:mXpp}) and (\ref{eqn:alg:FFBDD:forw:VXpp})}
If $	\msgf{m}{X_{n}^{'''}} = \sigma^{2}	\msgf{m}{X_{n}^{'''}}^{(1)} +	\msgf{m}{X_{n}^{'''}} ^{(0)}$ 
and $\msgf{m}{U_{n}} = \sigma^{2}\msgf{m}{U_{n}}^{(1)} + \msgf{m}{U_{n}}^{(0)}$, then
\begin{equation}
	\msgf{m}{X_{n}^{''}} = \sigma^{2} \mleft( \msgf{m}{X_{n}^{'''}}^{(1)} + b_{n}\msgf{m}{U_{n}}^{(1)}  \mright) 
	+\mleft( \msgf{m}{X_{n}^{'''}}^{(0)} + b_{n}\msgf{m}{U_{n}}^{(0)}  \mright).
\end{equation}
Likewise, if $\msgf{V}{X_{n}^{'''}} =\sigma^{2}\msgf{V}{X_{n}^{'''}}^{(1)}$ and $\msgf{V}{U_{n}}=\sigma^{2}\msgf{V}{U_{n}}^{(1)}$, then
\begin{equation}
	\msgf{V}{X_{n}^{''}} = \sigma^{2}\mleft(\msgf{V}{X_{n}^{'''}}^{(1)} + b_{n} \msgf{V}{U_{n}}^{(1)}b_{n}^{\mathsf{T}} \mright).
\end{equation}

\itemhead{(\ref{eqn:alg:FFBDD:forw:G}) and (\ref{eqn:alg:FFBDD:forw:g})}
If $\msgf{m}{X_{n}^{''}} = \sigma^{2}\msgf{m}{X_{n}^{''}}^{(1)} + \msgf{m}{X_{n}^{''}}^{(0)}$ and $	\msgf{V}{X_{n}^{''}} =\sigma^{2}	\msgf{V}{X_{n}^{''}} ^{(1)}$,then
\begin{equation}
	G_{n}  = \begin{cases}
		0,  & \text{if $\msgb{W}{Y_n}=0$}\\ 
		\sigma^{-2}\mleft(c_{n}^{\mathsf{T}}\msgf{V}{X_{n}^{''}}^{(1)}c_{n}\mright)^{-1},  & \text{if $\msgb{V}{Y_n}=0$}\\
	\end{cases}
\end{equation}
and
\begin{equation}
	g_{n} =  \begin{cases}
		c_{n}\msgb{\xi}{Y_{n}}, &  \text{if $\msgb{W}{Y_{n}}=0$},\\
		% b_{n}    \big(b_{n}^{\mathsf{T}}\msgb{W}{X_{n}^{''}}b_{n}\big)^{-1}\big(\msgf{\xi}{U_{n}}+b_{n}^{\mathsf{T}}\msgb{\xi}{X_{n}^{''}}\big), & \msgf{W}{U_{n}}=0,\\
		\sigma^{-2}\bigg(c_{n}\mleft(c_{n}^{\mathsf{T}}\msgf{V}{X_{n}^{''}}^{(1)}c_{n}\mright)^{-1} \mleft( \msgb{m}{Y_{n}}-c_{n}^{\mathsf{T}}\msgf{m}{X_{n}^{''}}^{(0)}\mright)\bigg)
		+ \bigg(c_{n}\mleft(c_{n}^{\mathsf{T}}\msgf{V}{X_{n}^{''}}^{(1)}c_{n}\mright)^{-1} \mleft( -c_{n}^{\mathsf{T}}\msgf{m}{X_{n}^{''}}^{(1)}\mright)\bigg)
		, & \text{if $\msgb{V}{Y_n}=0$},
	\end{cases}
\end{equation}
where $\msgb{W}{Y_{n}}=0$ and $\msgb{\xi}{Y_{n}}$ does not depend on $\sigma$, or else $\msgb{V}{Y_{n}}=0$ and $\msgb{m}{Y_{n}}$ does not depend on $\sigma$.

\itemhead{(\ref{eqn:alg:FFBDD:forw:mX}) and (\ref{eqn:alg:FFBDD:forw:VX})} If $G_{n}=\sigma^{-2}G_{n}^{(1)}$, $g_{n}=\sigma^{-2}g_{n}^{(1)} + g_{n}^{(0)}$, $\msgf{m}{X_{n}^{''}} = \sigma^{2}\msgf{m}{X_{n}^{''}}^{(1)} + \msgf{m}{X_{n}^{''}}^{(0)}$ and $	\msgf{V}{X_{n}^{''}} =\sigma^{2}	\msgf{V}{X_{n}^{''}} ^{(1)}$, then
\begin{equation}
	\msgf{m}{X_{n}} = \sigma^{2} \mleft(\msgf{m}{X_{n}^{''}}^{(1)} + \msgf{V}{X_{n}^{''}} ^{(1)} g_{n}^{(0)}\mright) + 
	\mleft(\msgf{m}{X_{n}^{''}}^{(0)} + \msgf{V}{X_{n}^{''}} ^{(1)} g_{n}^{(1)}\mright)
\end{equation}
and
\begin{equation}
	\msgf{V}{X_{n}} = \sigma^{2}\mleft( \msgf{V}{X_{n}^{''}}^{(1)} - \msgf{V}{X_{n}^{''}}^{(1)} c_{n}G_{n}^{(1)}c_{n}^{\mathsf{T}}\msgf{V}{X_{n}^{''}}^{(1)}\mright).
\end{equation}

\itemhead{(\ref{eqn:DualMarginalEstimateTerminalState})} If $\msgf{m}{X_{N}}=\sigma^{2}\msgf{m}{X_{N}}^{(1)} +\msgf{m}{X_{N}}^{(0)}$, $\msgf{V}{X_{N}}=\sigma^{2}\msgf{V}{X_{N}}^{(1)} $, $\msgb{m}{X_{N}}=\sigma^{2}\msgb{m}{X_{N}}^{(1)} +\msgb{m}{X_{N}}^{(0)}$ and $\msgb{V}{X_{N}}=\sigma^{2}\msgb{V}{X_{N}}^{(1)} $, then
\begin{equation}
	\tilde{\xi}_{X_{N}} = \sigma^{-2}\bigg( \mleft(\msgf{V}{X_{N}}^{(1)} + \msgb{V}{X_{N}}^{(1)}\mright)^{-1}\mleft(\msgf{m}{X_{N}}^{(0)} - \msgb{m}{X_{N}}^{(0)}\mright)\bigg)
	+ \bigg( \mleft(\msgf{V}{X_{N}}^{(1)} + \msgb{V}{X_{N}}^{(1)}\mright)^{-1}\mleft(\msgf{m}{X_{N}}^{(1)} - \msgb{m}{X_{N}}^{(1)}\mright)\bigg)
\end{equation}

\itemhead{(\ref{eqn:alg:FFBDD:backw:msgfWY}) and (\ref{eqn:alg:FFBDD:fwd:msgfXiY})} If $\msgf{m}{X_{n}^{''}} = \sigma^{2}\msgf{m}{X_{n}^{''}}^{(1)} + \msgf{m}{X_{n}^{''}}^{(0)}$, $\msgf{V}{X_{n}^{''}} =\sigma^{2}	\msgf{V}{X_{n}^{''}} ^{(1)}$ and $\tilde{\xi}_{X_{n}} =\sigma^{-2}\tilde{\xi}_{X_{n}}^{(1)} +\tilde{\xi}_{X_{n}}^{(0)}$, then
\begin{equation}
	\msgf{V}{Y_{n}} = \sigma^{2}\mleft(c_{n}^{\mathsf{T}}\msgf{V}{X_{n}^{''}}^{(1)}c_{n}\mright)
\end{equation}
and
\begin{equation}
	\msgf{m}{Y_{n}} = 
	\sigma^{2}\mleft(c_{n}^{\mathsf{T}}\msgf{m}{X_{n}^{''}}^{(1)}-c_{n}^{\mathsf{T}}\msgf{V}{X_{n}^{''}}^{(1)}	\tilde{\xi}_{X_{n}}^{(0)} \mright) +	\mleft(c_{n}^{\mathsf{T}}\msgf{m}{X_{n}^{''}}^{(0)}-c_{n}^{\mathsf{T}}\msgf{V}{X_{n}^{''}}^{(1)}	\tilde{\xi}_{X_{n}}^{(1)} \mright).
\end{equation}
\itemhead{(\ref{eqn:alg:FFBDD:backw:hatDualY})} If $\msgf{V}{Y_{n}} = \sigma^{2}\msgf{V}{Y_{n}} ^{(1)}$ and  $\msgf{m}{Y_{n}} = \sigma^{2}\msgf{m}{Y_{n}}^{(1)} +  \msgf{m}{Y_{n}}^{(0)} $, then
\begin{equation} 
	\tilde{\xi}_{Y_{n}} = \begin{cases}
		-\msgb{\xi}{Y_n}, & \text{if $\msgb{W}{Y_n}=0$} \\
		\sigma^{-2} \bigg( (\msgf{V}{Y_{n}}^{(1)})^{-1}\mleft( \msgf{m}{Y_n}^{(0)}  -  \msgb{m}{Y_n}\mright)\bigg) + \bigg((\msgf{V}{Y_{n}}^{(1)})^{-1}\mleft( \msgf{m}{Y_n}^{(1)}\mright)\bigg),  & \text{if $\msgb{V}{Y_n}=0$}
	\end{cases}
\end{equation}
where $\msgb{W}{Y_{n}}=0$ and $\msgb{\xi}{Y_{n}}$ does not depend on $\sigma$, or else $\msgb{V}{Y_{n}}=0$ and $\msgb{m}{Y_{n}}$ does not depend on $\sigma$.

\itemhead{(\ref{eqn:alg:FFBDD:backw:hatDualXpp})} If $\tilde{\xi}_{X_{n}}=\sigma^{-2}\tilde{\xi}_{X_{n}}^{(1)} + \tilde{\xi}_{X_{n}}^{(0)}$ and $\tilde{\xi}_{Y_{n}}=\sigma^{-2}\tilde{\xi}_{Y_{n}}^{(1)} + \tilde{\xi}_{Y_{n}}^{(0)}$, then
\begin{equation}
	\tilde{\xi}_{X_n''}=\sigma^{-2}\mleft(	\tilde{\xi}_{X_{n}}^{(1)} + c_{n} \tilde{\xi}_{Y_{n}}^{(1)} \mright)
	+ \mleft(	\tilde{\xi}_{X_{n}}^{(0)} + c_{n} \tilde{\xi}_{Y_{n}}^{(0)} \mright).
\end{equation}

\itemhead{(\ref{eqn:alg:FFBDD:backw:hatDualX}) and (\ref{eqn:alg:FFBDD:backw:hatU})} If $\tilde{\xi}_{X_n''}=\sigma^{-2}\tilde{\xi}_{X_n''}^{(1)} +\tilde{\xi}_{X_n''}^{(0)}$, $\msgf{m}{U_{n}} = \sigma^{2}\msgf{m}{U_{n}}^{(1)} +\msgf{m}{U_{n}}^{(0)}$ 
and $\msgf{V}{U_{n}} = \sigma^{2}\msgf{V}{U_{n}} ^{(1)}$, then
\begin{equation}
	\tilde{\xi}_{X_{n-1}}= \sigma^{-2}\mleft(A^{\mathsf{T}}\tilde{\xi}_{X_n''}^{(1)} \mright) + \mleft( A^{\mathsf{T}}\tilde{\xi}_{X_n''}^{(0)} \mright)
\end{equation}
and 
\begin{equation}
	\hat{u}_{n} = \sigma^{2}\mleft(\msgf{m}{U_{n}}^{(1)} - \msgf{V}{U_{n}} ^{(1)}b_{n}^{\mathsf{T}} \tilde{\xi}_{X_n''}^{(0)}\mright) + \mleft(\msgf{m}{U_{n}}^{(0)} - \msgf{V}{U_{n}} ^{(1)}b_{n}^{\mathsf{T}} \tilde{\xi}_{X_n''}^{(1)}\mright). 
\end{equation}

\itemhead{(\ref{eqn:alg:FFBDD:backw:hatY})} If $\msgf{V}{Y_{n}} = \sigma^{2}\msgf{V}{Y_{n}} ^{(1)}$ and $\msgf{m}{Y_{n}} = \sigma^{2}\msgf{m}{Y_{n}}^{(1)} +  \msgf{m}{Y_{n}}^{(0)} $, then
\begin{equation} 
	\hat{y}_{n} = \begin{cases}
		\msgb{m}{Y_{n}}, & \text{if $\msgb{V}{Y_n}=0$} \\
		\sigma^{-2} \mleft( \msgf{m}{Y_n}^{(1)}  + \msgf{V}{Y_n}^{(1)} \msgb{\xi}{Y_{n}}\mright) + \mleft( \msgf{m}{Y_n}^{(0)}\mright),  & \text{if $\msgb{W}{Y_n}=0$}
	\end{cases}
\end{equation}
where $\msgb{W}{Y_{n}}=0$ and $\msgb{\xi}{Y_{n}}$ does not depend on $\sigma$, or else $\msgb{V}{Y_{n}}=0$ and $\msgb{m}{Y_{n}}$ does not depend on $\sigma$.

\itemhead{(\ref{eqn:alg:FFBDD:hatx0})} If $\msgf{m}{X_{0}} = \sigma^{2}\msgf{m}{X_{0}} ^{(1)}+\msgf{m}{X_{0}} ^{(0)}$, $\msgf{V}{X_{0}} = \sigma^{2}\msgf{V}{X_{0}} ^{(1)}$ and $\tilde{\xi}_{X_0} = \sigma^{-2}\tilde{\xi}_{X_0}^{(1)} +  \tilde{\xi}_{X_0}^{(0)} $, then
\begin{equation}
	\hat{x}_{0} =  \sigma^{2}\mleft(\msgf{m}{X_{0}}^{(1)} - \msgf{V}{X_{0}} ^{(1)}\tilde{\xi}_{X_0}^{(0)}\mright) + 
	\mleft(\msgf{m}{X_{0}}^{(0)} - \msgf{V}{X_{0}} ^{(1)}\tilde{\xi}_{X_0}^{(1)}\mright).
\end{equation}

\clearpage
%\newpage

\bibliographystyle{IEEEtran}
\bibliography{IEEEabrv,paper}

\end{document}